\documentclass[lettersize,journal]{IEEEtran}
\usepackage{amsmath,amsfonts}
\usepackage{algorithmic}
\usepackage{algorithm}
\usepackage{array}
\usepackage{textcomp}
\usepackage{stfloats}
\usepackage{url}
\usepackage{verbatim}
\usepackage{graphicx}
\usepackage{cite}
\usepackage{multirow}
\usepackage{subcaption}
\usepackage{caption}
\usepackage{amsthm, amssymb}
\theoremstyle{definition}
\newtheorem{definition}{Definition}[section]

\newcommand\shnote[1]{\textcolor{blue}{SH: #1}}
\newcommand{\eanote}[1]{\textcolor{purple}{EA: #1}}
\newcommand{\rcnote}[1]{\textcolor{teal}{RC: #1}}

\newcommand{\eg}{\textit{e.g.,~}}
\newcommand{\ie}{\textit{i.e.,~}} 

\newcommand{\ignore}[1]{}

\begin{document}

\title{Anticipating Technical Expertise and Capability Evolution in Research Communities using Dynamic Graph Transformers}

\author{Sameera Horawalavithana,
Ellyn Ayton,
Anastasiya Usenko,
Robin Cosbey,
and
Svitlana Volkova
\thanks{S. Horawalavithana, E. Ayton, A. Usenko, and R. Cosbey are with Pacific Northwest National Laboratory, Richland, 
WA 99354 USA. S. Volkova was with Pacific Northwest National Laboratory during this research. Corresponding Author e-mail: yasanka.horawalavithana@pnnl.gov. (see https://www.pnnl.gov/projects/expert for more details)}
\thanks{Manuscript received ..; revised ...}}



\maketitle

\begin{abstract}
The ability to anticipate technical expertise and capability evolution trends globally is essential for national and global security, especially in safety-critical domains like nuclear nonproliferation (NN) and rapidly emerging fields like artificial intelligence (AI). In this work, we extend traditional statistical relational learning approaches (\eg link prediction in collaboration networks) and formulate a problem of anticipating technical expertise and capability evolution using dynamic heterogeneous graph representations. We develop novel capabilities to forecast collaboration patterns, authorship behavior, and technical capability evolution at different granularities (\eg scientist and institution levels) in two distinct research fields.
We implement a dynamic graph transformer (DGT) neural architecture, which pushes the state-of-the-art graph neural network models by (a) forecasting heterogeneous (rather than homogeneous) nodes and edges, and (b) relying on both discrete- and continuous-time inputs. We demonstrate that our DGT models predict collaboration, partnership, and expertise patterns with 0.26, 0.73, and 0.53 mean reciprocal rank values for AI and 0.48, 0.93, and 0.22 for NN domains. DGT model performance exceeds the best-performing static graph baseline models by 30-80\% across AI and NN domains.
Our findings demonstrate that DGT models boost inductive task performance, when previously unseen nodes appear in the test data, for the domains with emerging collaboration patterns (\eg AI). Specifically, models accurately predict which established scientists will collaborate with early career scientists and vice-versa in the AI domain. 
\end{abstract}


\section{Introduction}
Monitoring technical expertise and capability evolution globally is an extremely challenging but highly desired task. This is especially true for critical national security domains like artificial intelligence (AI) and nuclear nonproliferation (NN). It is also important to know when new technical capabilities emerge globally, when scientists start or stop publishing about specific technologies, when new collaborations (\eg international and multidisciplinary) are established,  when industry and academic partnerships emerge, and when publication behaviours for scientists of interests have a potential to transform the national security mission,
specifically by:
\begin{itemize}
    \item providing understanding of how publicly available data could be used to monitor, forecast, and reason about potential proliferation and adversarial AI technologies globally~\cite{national2021nuclear};
    \item assuring quality, scale, and timeliness required for operational monitoring capability;
    \item moving away from traditional reactive analyses and taking a proactive posture.
\end{itemize}
To create this operational capability, we
developed and  validated a dynamic graph transformer (DGT) network, a novel deep-learning architecture that leverages  attributes from both graph neural networks (GNNs) and Transformer models, for forecasting 
nodes and edges given discrete- or continuous-time historical inputs (\eg publication behavior, collaboration patterns).
We construct these inputs from digital scholarly publications that capture scientific knowledge development and collaboration patterns across disciplines, e.g., artificial intelligence, nuclear science.
They provide new insights on how careers evolve, how collaborations drive scientific discovery, and how scientific progress emerges 
 which enable researchers to gain a deeper understanding of the relationships and dynamics within the scientific community~\cite{wang2021science,horawalavithana2022expert}.
Unlike any other work, our DGT models learn from dynamic {\it heterogenous} structured representations of scientific collaborations, capability, and expertise evolution patterns at multiple levels of granularity (\eg at scientist and institution).
In contrast to predicting missing edges in a static graph, our models forecast the temporal edges in a dynamic graph.
For example, the model predicts which two scientists would collaborate in next month while it predicts which emerging technical capability a scientist will publish on next year.

To understand advantages and limitations of DGT models, we performed an in-depth analysis and comparison of performance across GNN models, types of relations (scientist-to-capability, scientist-to-scientist, etc.), and exogenous variables (country, venues, etc.). 
First, we test the ability of the models to generalize across Nuclear nonproliferation (NN) and Artificial Intelligence (AI) domains with diverse publication characteristics.
We noticed that the DGT models trained with discrete-time and continuous-time dynamic graphs outperform the best-performing static graph baseline models by 30-80\% in the Nuclear nonproliferation and Artificial Intelligence domains, respectively. 
Second, we demonstrate that our DGT models can predict both edges for nodes they have seen during training as well as "unseen" nodes they will encounter once deployed, which is critical for operational capability. 
For example, models generalize predictions to scientists who focus on emerging research topics in Artifical Intelligence, or to newcomer scientists who involve in collaborations with veteran scientists in Nuclear nonproliferation.
Third, our detailed performance analysis suggests that collaborations across scientists and institutions within
the same country (domestic) are easier to anticipate than cross-country collaborations (international); collaboration
patterns within the United States are easier to anticipate than those outside the United States, with collaborations from
China being the most difficult to forecast in the nuclear domain.
The models also make highly accurate predictions for
highly prolific, influential and interdisciplinary scientists.


\section{Related Work}

In this section, we summarize prior work on dynamic graph modeling approaches. Previous research primarily focused on three tasks: node classification, edge prediction, and graph classification~\cite{kazemi2020representation}.
Edge prediction problems are well studied under two settings: interpolation and extrapolation. While interpolation focuses on predicting missing links in the past, extrapolation focuses on predicting future links and is more challenging. Our work focuses on an extrapolation  in both discrete- and continuous-time dynamic graphs.
An example extrapolation task is to predict who a given scientist will collaborate with next.
We describe the recent dynamic graph models that perform these link prediction tasks in Section~\ref{sec:gnn_vs_transformers} and highlight the most relevant works closest to our problem domain in Section~\ref{sec:aca_graphs}.


\subsection{Graph Neural Networks vs. Transformers}
\label{sec:gnn_vs_transformers}
Most of existing dynamic graph models use graph neural networks (GNN) and recurrent neural networks (RNN) to predict links in dynamic graphs~\cite{pareja2020evolvegcn,goyal2020dyngraph2vec,kazemi2020representation}.
RNNs and GNNs are used jointly to learn the temporal graph sequence and graph structural information, respectively.
For example, \textsc{GCRN}~\cite{seo2018structured} and \textsc{EvolveGCN}~\cite{pareja2020evolvegcn} use RNN and graph convolution neural network (GCN)~\cite{defferrard2016convolutional} to learn from discrete-time graph snapshots.
JODIE~\cite{kumar2019predicting} extended the RNN models to learn dynamic embeddings from a sequence of temporal interactions.
However, most of these methods are limited to \textit{homogeneous} (single-relational) dynamic graphs and do not handle multiple types of nodes and edges or other node features prevalent in the dynamic \textit{heterogeneous} graphs.
More recently, some studies focused on predicting the future links in dynamic heterogeneous graphs~\cite{jin2019recurrent,rossi2020temporal}.
Jin et al. proposed the Recurrent Event Network (RE-Net) to predict future links in a discrete-time dynamic graphs ~\cite{jin2019recurrent}.
Rossi et al. proposed the Temporal Graph Network (TGN) to predict future edges in a continuous-time dynamic graph ~\cite{rossi2020temporal}.
In contrast to RE-Net, TGN accepts a sequence of timed edges as input to learn time-aware node embeddings. 
Both RE-Net and TGN models use RNN to handle node and edge updates.
However, RNNs does not perform well when the number of timesteps increases in the temporal link prediction tasks~\cite{cong2021dynamic}.
We benchmark novel DGT models against the RE-Net and TGN approaches.

Transformers achieved a great success in a broad class of machine-learning (ML) problems across multiple data modalities such as language~\cite{horawalavithana-etal-2022-foundation} and vision~\cite{khan2022transformers}, and recently on graphs~\cite{bommasani2021opportunities}.
For example, \textsc{Graphormer}~\cite{ying2021transformers} and \textsc{GraphTransformer}~\cite{dwivedi2020generalization} use Transformer architecture~\cite{vaswani2017attention} to implement message aggregation and positional encoding in graphs.
However, \textsc{Graphormer} is evaluated on small molecule graphs and \textsc{GraphTransformer} only extracts features from one-hop neighborbood.
\textsc{GraphBERT}~\cite{zhang2020graph} and \textsc{TokenGT}~\cite{kim2022pure} transform a graph into the node and edge sequences and feed into Transformers.
\textsc{TokenGT} has shown to be more expressive than all message-passing GNNs and outperforms \textsc{GraphTransformer} in a standard large-scale graph regression benchmark.
However, when graph topology is important to the downtream prediction task, both \textsc{GraphBERT} and \textsc{TokenGT} can perform poorly since they do not take advantage of it.
While Transformer-based methods applied on graphs show clear performance advantage over other GNN methods, most of these methods are limited to static graphs~\cite{ying2021transformers,kreuzer2021rethinking} 

Several works combined a self-attention mechanism of the Transformer architecture with GNN and demonstrated performance advantage over message passing GNNs~\cite{min2022transformer}.
For instance, DYSAT~\cite{sankar2018dynamic} suggests using the self-attention mechanism for the aggregation of temporal and structural information. 
TGAT~\cite{xu2020inductive} first applies self-attention to the temporal augmented node features after encoding the temporal information into the node feature.
However, most of these methods make graph-specific architectural assumptions~\cite{kim2022pure}.
Cong et al.~\cite{cong2021dynamic} used a Transformer-based method in their approach to learn from temporal-union graphs extracted from dynamic graph snapshots.
However, this method has not been evaluated on dynamic heterogeneous graphs.
In this work, we advance GNN and Transformer architectures to operate on both discrete- and continuous-time dynamic heterogeneous graphs.
Specifically, we use a self-attention mechanism to learn  dynamic graph- and node-level changes and GNN to learn structural information in both global and local neighborhoods.
We do not make any domain specific architectural assumptions.
DGT models jointly learn from both temporal edge features and heterogeneous graph neighborhoods.

\subsection{Academic Graph Modeling}
\label{sec:aca_graphs}
Previous work in the science of science domain primarily focused on co-citation or co-authorship networks (\eg predicting  missing edges in a co-citation network~\cite{yang2015defining} or a co-authorship network~\cite{hu2020open}).
DBLP~\cite{yang2015defining} and ogbl-citation2~\cite{hu2020open} are two commonly used benchmarks for link prediction in static academic networks.  
Similarly, HEP-PH~\cite{gehrke2003overview} is a benchmark for link prediction in a co-citation network, but takes into account a dynamic graph setting.
However, these datasets and benchmarks are not appropriate to use for evaluation on edge forecasting tasks on \textit{dynamic heterogeneous} graphs.

Most recently, Hu et al. introduced a new {OGB-LSC} challenge benchmark for graph ML problems~\cite{hu2021ogb}.
OGB datasets are extracted from the Microsoft Academic Graph~\cite{wang2020microsoft}.
Several recent works show the usefulness of the OGB datasets for large-scale graph learning~\cite{hu2021ogb}.
One of the tasks in the challenge is to predict the missing subject categories of scientific articles in a heterogeneous academic graph.
The top performing solutions in the challenge used different variants of message-passing based GNNs (e.g., R\_UniMP~\cite{shi2021r}, MDGNN~\cite{zhaomdgnn} and MPNN\&BGRL~\cite{addanki2021large}).
They also highlighted the importance of relation-aware node sampling in the heterogeneous graph learning.
While these solutions provide more insights to academic graph modeling, they are limited to static graphs, and the corresponding node classification tasks.
Apart from these ad-hoc prediction problems, there have been very few attempts to model global expertise and capability evolution in large-scale dynamic heterogeneous academic graph data.
These graphs contain multiple types of nodes (e.g., scientists, institutions, capabilities) and edges (e.g., collaboration, partnership) that evolve over time~\cite{horawalavithana2022expert}.
For example, these graphs capture evolving interaction patterns across scientists that may exhibit the research trends and traits of academic communities.
In addition, the temporal link prediction in dynamic heterogeneous academic graphs provides an ideal benchmark to test how well machine learning models generalize to the unseen test distributions, often called as \textit{spatio-temporal distribution shifts}~\cite{zhang2022dynamic}.

\section{Methodology}
\label{sec:methodology}

This work leverages millions of research articles between 2015 and 2022 in two research domains (NN and AI) to understand and reason about the evolution of technical expertise and capabilities globally. For that we propose DGT, a new deep-learning method that integrates GNN with a Transformer architecture to forecast how technical expertise and capability development emerge through a combination of multiple interconnected factors. For example, our model learns features of human behavior extracted from historical collaborations, partnerships, and capability evolution to answer operationally relevant questions about proliferation risk assessment globally and competition in developing AI technologies. 

Specifically, we seek to answer research questions about scientific collaborations, partnerships, and capability development when studying dynamic graph model performance.
%
\begin{itemize}
    \item \emph{Can we model varied patterns of behavior underlying the way scientists collaborate?}
Previous work~\cite{wang2021science} has shown that team-authored publications are more popular in terms of citations than single-authored publications. In our model we study scientists who are engaging more or less in collaborations. We model collaborations that occur within tightly connected groups of scientists with some engaging within the same or across multiple institutions.

    \item \emph{Can we model individuals and institutions making new partnerships?}
In our proposed model we focus on studying researchers at top universities who are more likely to collaborate with scientists at other top universities.

    \item \emph{Can we model the differences in technical capabilities that scientists research?}
We model scientists who adopt the most recent and emerging research trends and disrupt science by developing novel technologies and other scientists generate more theoretical innovations in contrast to applied technologies.
\end{itemize}

\subsection{Problem Formulation}
\label{sec:problem_formulation}
\begin{figure}[htbp]
    \centering
    \includegraphics[width=8cm]{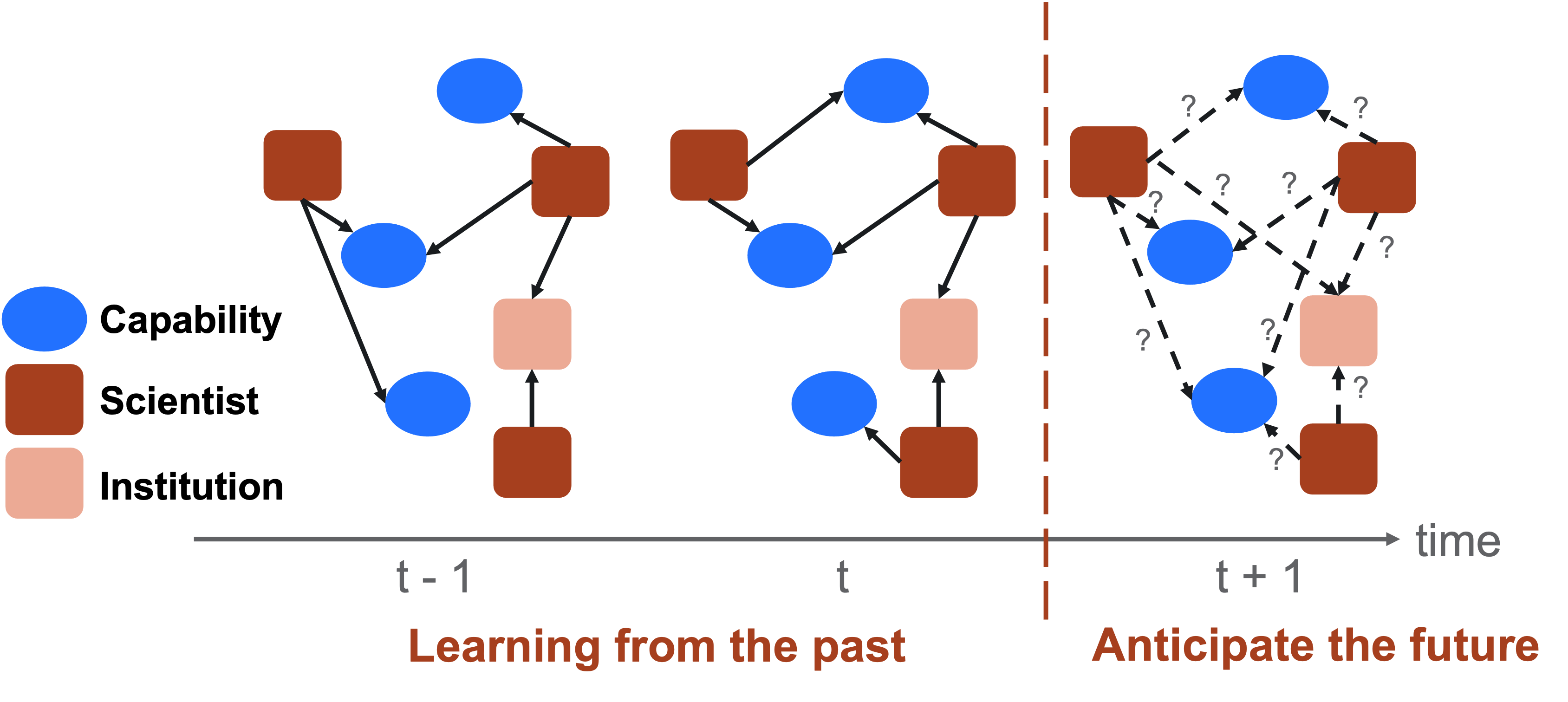}
    \caption{Problem definition - forecasting technical expertise and capability evolution using structured dynamic representations.}
    \label{fig:task_definition}
\end{figure}

We consider dynamic heterogeneous graphs $G$ consisting of scientists, institutions, and capabilities as nodes $N$.
A pair of nodes is connected at a timestamp $t$, by a directed edge that is denoted by a quadruplet (\textit{$N_i$}, $E$, \textit{$N_j$}, $t$). 
Edges are of multiple types $E$ such as collaboration $E_c$ (\textit{scientist-to-scientist}), partnership $E_p$ (\textit{scientist-to-institution}), and research focus $E_c$
 (\textit{scientist-to-capability}).
An ordered sequence of quadruplets represents the dynamic heterogeneous graph.
In contrast to predicting missing edges in a static graph (\textit{interpolation}), we need to predict the future edges in a dynamic graph (\textit{extrapolation}).
As these edges occur over multiple timestamps in the future, we treat the prediction task as a multistep inference problem
(see Figure~\ref{fig:task_definition} and Definition~\ref{def:problem}).
Thus, we need to develop methods that can extrapolate the heterogeneous graph structure over future timestamps~\cite{horawalavithana2022expert,jin2019recurrent}.
Such predictions are extremely useful to forecast emerging science trends in terms of global expertise and capability development in domains like NN and AI.


\begin{definition}
\label{def:problem}
Given a graph $G_{t}$ that represents the ordered sequence of quadruplets until time $t$, the task is to forecast the graph ($G_{t:t+m}$) over multiple future timesteps $m$.
$G_{t}$ can be represented as \textit{discrete-time dynamic graphs} (\eg sequences of static graph snapshots) and~\textit{continuous-time dynamic graphs} (\eg timed lists of heterogeneous edges).
\end{definition}

\subsection{Dynamic Graph Transformers}
\label{sec:dgt}
\begin{figure*}[htbp]
    \centering
    \includegraphics[scale=0.41]{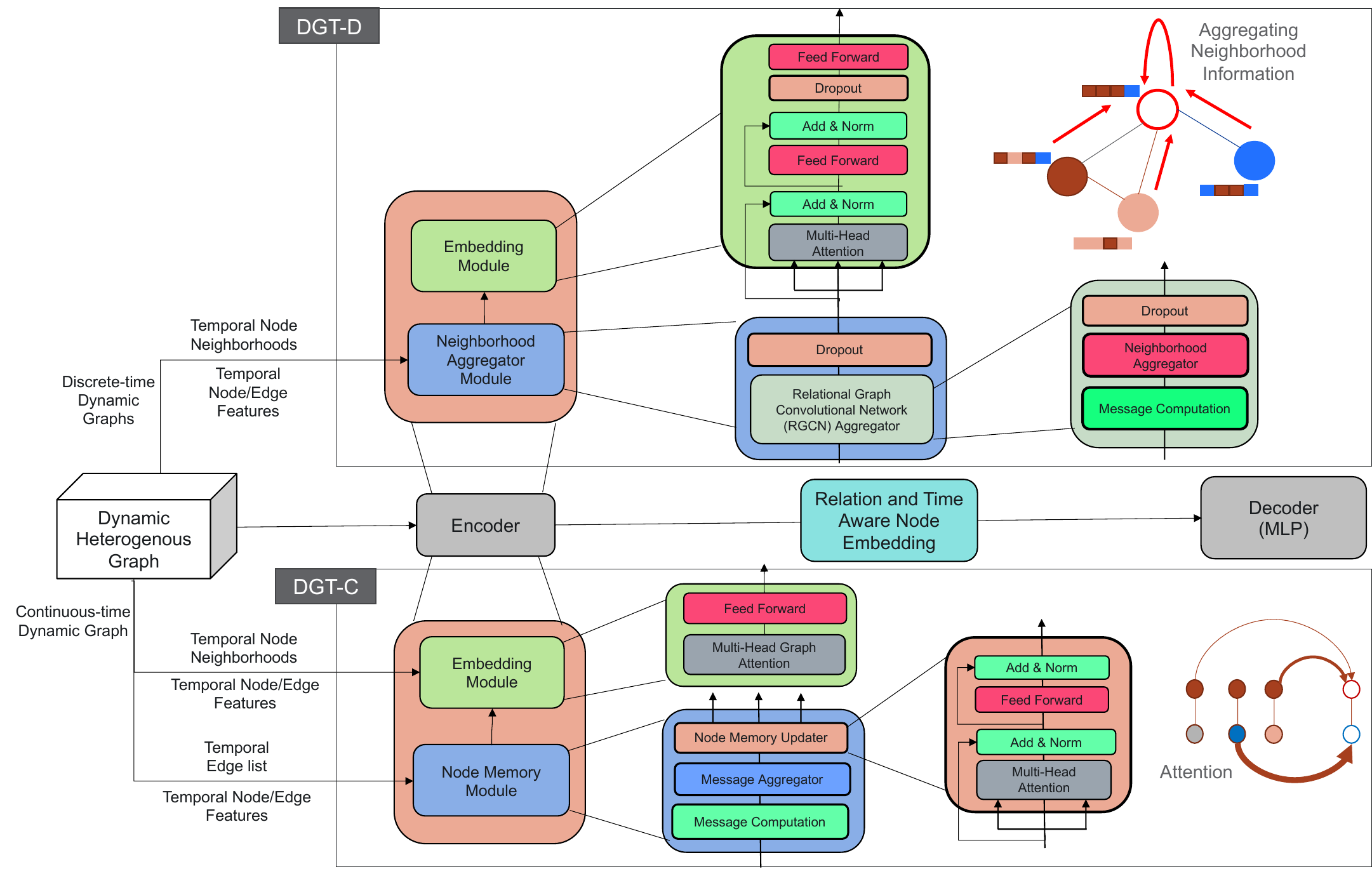}
    \caption{DGT architecture with DGT-D and DGT-C variations to handle discrete-time and continuous-time inputs.}
    \label{fig:dgt_architecture}
\end{figure*}

In this work, we introduce DGT to operate on dynamic heterogeneous graph inputs.
DGT learns latent node representations from both discrete-time and continuous-time dynamic graphs that consist of heterogeneous node and edge types.
We combine GNN and Transformers to learn time-aware and structure-aware node representations (Figure~\ref{fig:dgt_architecture}).
Our objective is to map dynamic graphs to node embedding that can be useful for the temporal link prediction task~\cite{kazemi2020representation}.
These node embeddings should contain multiple types of information (\eg heterogeneous node and edges, node and edge attributes, and graph temporal dynamics) captured by node and their structural neighborhood changes.
We describe the discrete-time and continuous-time model architecture in Sections~\ref{sec:discrete_model} and~\ref{sec:continous_model}, respectively.


\subsubsection{Discrete Model Architecture}
\label{sec:discrete_model}
Given a sequence of discrete-time graph snapshots, $G = G_{(1)}, G_{(2)},..., G_{(t-1)}$, the pretraining objective is to generate the next graph snapshot $G_{(t)}$ as shown in Equation~\ref{eq:discrete_pretraining}.

\begin{equation}
\label{eq:discrete_pretraining}
    p(G)=\prod_{t} \prod_{(N_i,E,N_j)\in G} p(N_{i},E,N_{j}|G(t-N:t-1))
\end{equation}

Our assumption is that all edges in $G_t$ depend on the edges at the previous $N$ timesteps.
We improve the RE-Net~\cite{jin2019recurrent} architecture's ability to learn the temporal and structural information from the discrete graph snapshots.
The proposed DGT Discrete (DGT-D) model consists of a neighborhood aggregator module and an embedding module as shown in Figure~\ref{fig:dgt_architecture}.
The neighborhood aggregator module includes the Relational Graph Convolutional Network (RGCN) layers to learn across multiple relations within a graph snapshot. This neighborhood aggregator outputs latent node representations that capture the $k$-hop neighborhood information. This allows the model to capture long-range context dependency present across nodes.
DGT-D advances the RE-Net  decoder to learn graph- and node-level temporal dependencies.
We use the Transformer architecture~\cite{vaswani2017attention} with a self-attention module and a position-wise feed-forward network to learn across latent representations from each graph snapshot.
This helps to alleviate over-smoothing and over-squashing problems present in the recurrent event decoder~\cite{kazemi2020representation}.

\ignore{
\begin{equation}
\label{eq:transformer}
\begin{aligned}
    Attn (H) = softmax (\frac{Q.K^T}{\sqrt{d_K}}) V, \\
    Q=H \mathbb{R}^{d\times d_K}, K=H \mathbb{R}^{d\times d_K}, V=H \mathbb{R}^{d\times d_V}, H \in \mathbb{R}^{n\times d} \\
\end{aligned}
\end{equation}
}

The input to the self-attention module is $H \in \mathbb{R}^{n\times d}$, where $\mathbb{R}^{1\times d}$ represent $d$-dimensional graph ($H_G$) and node ($H_N$) level representations.
For the global representation $H_G$, we use an element-wise max-pooling operation over all node representations within a graph snapshot.
We construct the local representations $H_N$ for each node by aggregating
neighborhood information from the node $N$. 
While global representations summarize graph-level information from the discrete graphs, the local representations capture the events related to the node.




\subsubsection{Continuous Model Architecture}
\label{sec:continous_model}
Given a sequence of time-stamped edges $G=\{\{N_i(1),E(1), N_j(1)\},...,\{N_{i}(t-1),E(t-1),N_{j}(t-1)\}\}$, the pretraining objective is to predict the probability of an edge in the next timestep $t$ as shown in the Equation~\ref{eq:continuous_pretraining}.

\begin{equation}
\label{eq:continuous_pretraining}
\begin{aligned}
    p((N_i,N_j)|t)&=p((z_{N_i}||z_{N_j})|
    \\
    &\{\{N_i(1),E(1),N_j(1)\},..., \\
    &\{N_i(t-1),E(t-1),N_j(t-1)\}\}) \\
\end{aligned}
\end{equation}

$z_{N_i}$ represents the hidden node representations for node $N_i$ that aggregates neighborhood information from the continuous-time dynamic graph. We assume that the presence of edge $(N_i,E,N_j)$ in the next timestep depends on the edge updates in the $k$-hop neighborhood of node endpoints at the previous timesteps.
We represent the continuous-time dynamic heterogeneous graph  as a list of timestamped edges.
We rely on the TGN architecture~\cite{rossi2020temporal} to inform the implementation of our DGT Continuous (DGT-C) model. Note, TGN generalizes to a majority of existing graph message passing-type architectures proposed for learning on both static and dynamic graphs.
This model consists of a node memory module and an embedding module. The node memory module learns to compress and memorize the historical events as node states. Node states are updated upon each event associated with the respective node. For example, when there is an interaction between nodes \textit{$N_i$} and \textit{$N_j$} at timestamp $t$, node states $\mathbb{M}_{N_i}$ and $\mathbb{M}_{N_j}$ are updated.
DGT-C advances the TGN node memory module by introducing a self-attention module and a position-wise feed-forward network to update the node states upon new events.
The input to the self-attention module
is a sequence of messages (\textit{msg}) computed from the temporal events ($N_i(t)$, $E(t)$, $N_j(t)$) $\in B(t)$) as shown in Equation~\ref{eq:tgn_memory}.

\begin{equation}
\label{eq:tgn_memory}
\begin{aligned}
        msg (t)_{Ni} &= \mathbb{M}(t-1)_{N_i} \oplus \mathbb{M}(t-1)_{N_j} \oplus \Delta t \oplus (N_i(t), N_j(t)) \\
\end{aligned}
\end{equation}

$\Delta t$ is the time difference between the current and previous event associated with the node $N_i$.
The output of the self-attention module is the updated node state $\mathbb{M}(t)_{N_i}$.
The embedding module takes the node states as the input and produces the time-aware node representations $z_i (t)$ as shown in Equation~\ref{eq:tgn_embedding}.
\begin{equation}
\label{eq:tgn_embedding}
\begin{aligned}
        z_i (t)=\sum_{j\in \eta_{i}^{k}([0,t])} TGAT (\mathbb{M}(t)_i,\mathbb{M}(t)_j,i,j)
\end{aligned}
\end{equation}

We use temporal graph attention (TGAT) to implement the embedding function~\cite{rossi2020temporal}.
TGAT uses self-attention to aggregate information from the most important $k$-hop node neighborhoods and Time2Vec to encode temporal information~\cite{kazemi2019time2vec}.
TGAT avoids the memory staleness problem on handling sparse dynamic graph signals~\cite{rossi2020temporal}.

\section{Data Collection and Processing}
\label{Data}
We evaluate our approaches to the edge prediction problem using data from the AI and NN domains. In this section we describe four datasets, two for each domain, and the methods of data collection and preprocessing.


\ignore{
\eanote{Get Soni's input on node feature curation}
Institution node features are aggregated \textcolor{red}{sums/average?} \textcolor{pink}{They're either "first" for non-temporal features like country and "mean" for others -Soni} measures of all scientists that are affiliated with that particular institution.
Using HuggingFace\footnote{\url{https://huggingface.co/}}, we create capability node features by extracting token level embeddings from SciBERT \cite{beltagy-etal-2019-scibert}. 
For every year in our dataset, we average the embedding vectors for every instance of each capability to build a single vector representation for each keyword. This preserves the dynamic, semantic representation of each capability keyword.}

\subsection{Artificial Intelligence Data}
The first domain we focus on is AI. For this use case, we constructed two separate graph networks using public, open-source conference and journal publication data. The result is the computational linguistics dataset from the Association for Computational Linguistics (ACL) and the ML dataset from ML conferences. 

\subsubsection{Computational Linguistics}
 The first of the two datasets relating to the AI domain is a collection of publications from the ACL. We collected 51K papers published between 1965 and 2021 from the ACL Anthology.\footnote{\url{https://aclanthology.org}}
This total set of publications was then filtered to include only papers containing one AI keyword.\footnote{AI keywords: adversarial, causal, clustering, dialog, ethic, explanation, fair, genetic algorithm, interpretability, interpretable, language model, machine translation, nlg, question-answer, reinforcement learning, sentiment, summarization, transfer learning, translation model, transparent.} The keyword list was curated by subject matter experts based on the frequency of keywords and coverage in the documents.
 Additionally, papers published before 2010 were removed to reduce sparsity and noise. 
 These two filtering steps resulted in $33K$ papers collected. 
 Using the GROBID~\cite{GROBID} approach, we extracted titles, abstracts, author names, author institutions, and location details. If location data were provided, we parsed the city/state and country names.
 After extracting the collaboration interactions, the resulting dataset contained $35.6K$ unique authors from $7.5K$ unique institutions. The final dataset included $478K$ edges across training, validation, and test splits. 
 
 
\subsubsection{Machine Learning}
In addition to the ACL dataset, we collected ML-related publications from the International Conference on Machine Learning (ICML), the International Conference on Learning Representations (ICLR), and the Conference on Neural Information Processing Systems (NeurIPS), merging them into a general ML dataset to complement the AI domain use case. We selected $6K$ papers from ICML during the years of 2009 to 2021, $2.5K$ papers from ICLR during the years of 2016 to 2021, and $10.5K$ papers from NeurIPS. The GROBID extraction process, as with ACL, was used to identify all necessary metadata.
We manually performed entity resolution to identify duplicate scientist and institution nodes and merged them across publication venues. The final ML dataset included $48.5K$ unique authors and $1.8K$ unique institutions, with a total of $210K$ edges across training, validation, and test sets.
In comparison to the ACL dataset, the ML dataset is less dynamic, with changes happening to the graph once a year. This dataset is also characterized as having more emerging interactions, meaning the links between nodes in the graph are overwhelmingly between previously unseen scientists, capabilities, and institutions. 

 \begin{figure*}[!t]
    \centering
    \includegraphics[width=.98\linewidth]{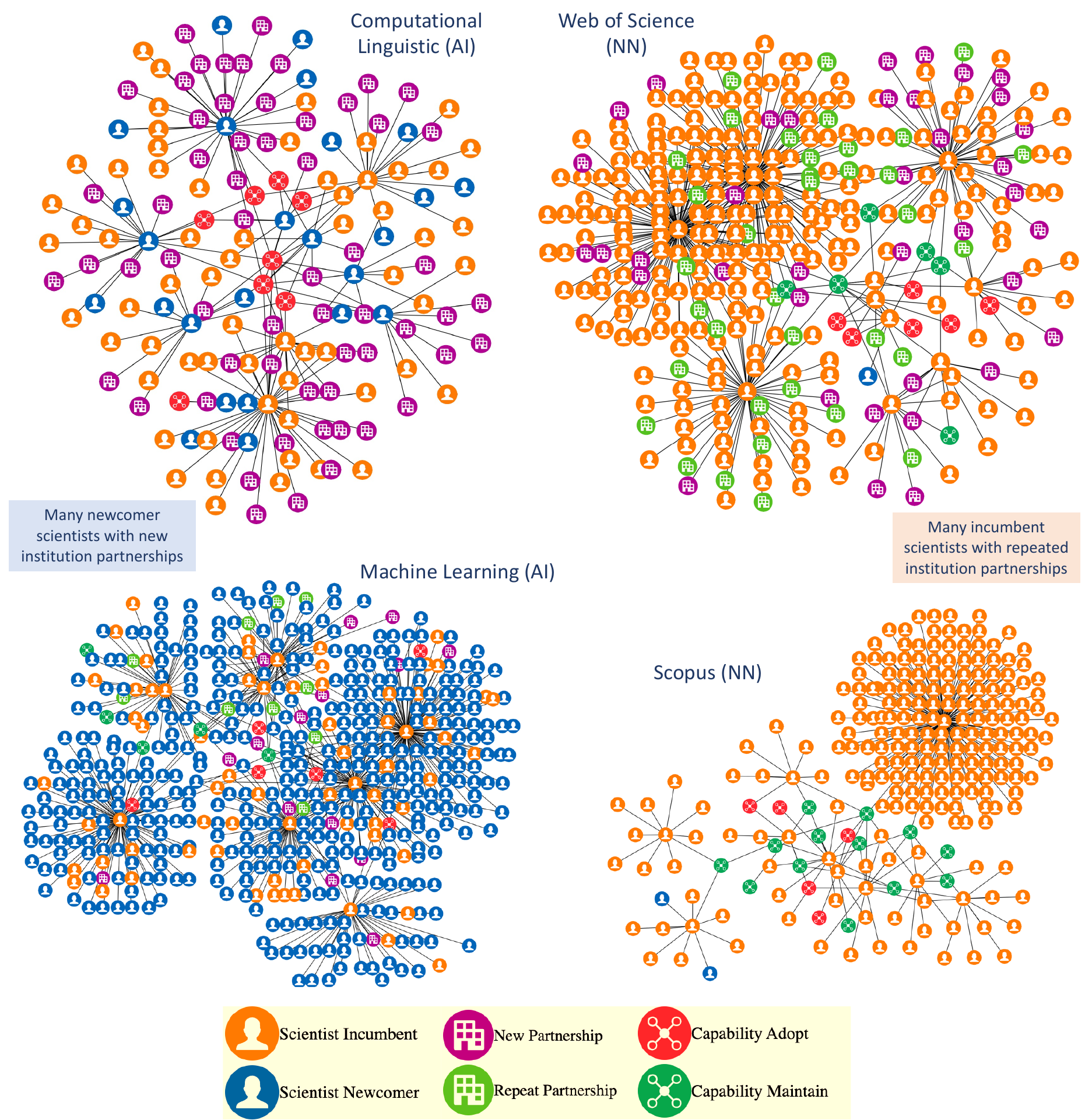}
    \caption{Visualization of the dynamic heterogeneous graphs across four datasets with diverse scientist, institution, and capability nodes with a focus on the subgraphs extracted from the top 10 most active scientists. Note, we removed all edge timestamps to reduce visual clutter.}
    \label{fig:network_viz}
\end{figure*}

\subsection{Nuclear Nonproliferation Data} 
The second domain we are interested in is NN. We constructed two datasets for this domain: from Web of Science (WoS) and Scopus. WoS and Scopus are multidisciplinary databases containing reference and citation records from over 250 academic fields. To create NN-specific datasets from multidisciplinary databases, we used a set of 51 keywords or phrases related to the topic of nuclear science and nonproliferation, which have been curated by subject matter experts.\footnote{NN keywords: centrifuge, chemical conversion, chemical engineer, chemical extraction, civil engineer, closed nuclear fuel cycle, computational physics, criticality test, depleted uranium, dose coefficient, electrical engineer, enriched uranium, fissile material, fission fragment, fission product, fissionable material, fuel cycle, fuel rod, hydrodynamic, international community, ionizing radiation, low enriched uranium, low linear energy transfer, military use, natural uranium, nuclear, nuclear accident, nuclear emergency, nuclear facility, nuclear fuel, nuclear incident, nuclear installation, nuclear material, nuclear security, plutonium, potential alpha energy, radiation emergency, radiation risks, radiation safety, radiation source, radioactive, radioactive equilibrium, radioactive half-life, radioactive material, radioactive source, radioactive waste, radiological emergency, radiological protection, research reactor, spent fuel, uranium.}
Our subsampled NN datasets contained at least one domain-specific keyword in the title or abstract.

 \subsubsection{Web of Science} 
The first dataset in the NN domain is a collection of publications from WoS.\footnote{\url{https://www.webofscience.com/wos/woscc/basic-search}} We sampled $531K$ scientific publications published between 2015 and 2021. We filtered this collection to include only publications that contained one or more NN keywords in the title or abstract. 
 Next, we choose to constrain the dataset to include publications authored by scientists with 20 or more papers published in WoS during this time. 
 This filtering step greatly reduced the number of total papers from $531K$ to $48K$ and reduced the sparsity of the dataset, increasing the likelihood for scientist nodes to have more than one link with another scientist, institution, and capability. The final collection of data from the WoS contained $6.7K$ authors and $7.6K$ institutions.
 
 \subsubsection{Scopus} 
 In addition to the WoS papers, we collected $105K$ publications from the Scopus database published between 2015 and 2021. We performed the same keyword filtering step to select publications containing at least one of the NN keywords in the title or abstract. Additionally, we filtered the resulting dataset to include only publications authored by scientists that had 15 or more papers published in Scopus during the this time. The final Scopus dataset contained $3.3K$ unique scientists and $7.2K$ unique institutions.
 This dataset was quickly dynamic with timesteps in a monthly granularity. The majority of the interactions were repeated, so the links in the graph were primarily across a groups of scientists and the same capabilities.

\subsection{Data Augmentation}
We performed additional preprocessing steps to support our analysis. First, we labeled scientists as incumbent or newcomer nodes based on their frequency of appearances in the dataset. For example, when a scientist published her first article in the computational linguistic (CL) community, we labeled her as a newcomer. Incumbent scientists published multiple times in the same community.
Similar to the above classification, we labeled the institutions into new and repeated partnerships.
For example, a scientist may establish a new partnership with an institution in the next publication, or might continue with the previously established partnership with an institution.
Finally, we characterized the publication trends on the adoption and persistence of certain research foci in the respective research communities~\cite{glenski2021identifying}.
When a scientist adopted a new research focus to work on for the first time, we labeled them as new capabilities.
Otherwise, a scientist maintained the same research focus across publications.
Figure~\ref{fig:network_viz} shows the visualization of the academic graphs with different types of nodes 
in the AI and NN domains.
There are more new scientists in the AI community who often collaborate with the incumbent scientists.
In contrast, the NN community is more densely structured with many incumbent scientists who participate in the new and repeated collaborations. 
For example, more than 95\% of collaborations occurred between two incumbent scientists in the WoS community in the NN domain.

As a part of our metadata extraction process, we extracted location information directly from the PDF (available for the ACL and ML datasets) or citation records (available for the WoS and Scopus datasets). 
The quality of extracted location data varied greatly across the datasets. The manual inspection was done to verify valid country names. City, state, or province names were mapped to country names whenever possible (\eg Beijing to China). Those publications missing location information, or those containing incorrect location information, or location names that could not be mapped to a specific country were labeled as Other.

\section{Experiments}
In this section, we describe the experimental setup and discuss the results from our experiments.
We evaluate the performance of the proposed  neural architecture developed to predict edges on dynamic heterogeneous graphs.
These graphs encode collaboration (\textit{scientist-to-scientist}), partnership (\textit{scientist-to-institution}), and expertise (\textit{scientist-to-capability}) edges. 
Similar to previous work~\cite{visin2015renet,kazemi2020representation}, we formulate this temporal link prediction task as a ranking problem. The goal is to rank potential nodes that would be present in a future graph.

Section~\ref{sec:exp_setup} describes our evaluation metrics and  baselines. First, we report model performance breakdown for the list of tasks with an increasing order of complexity 
 in Section~\ref{sec:tran_induct_tasks}.
We present inductive task performance when
previously unseen nodes could appear in the test data~\cite{horawalavithana2022expert}, and we summarize performance across different edge types in Section~\ref{sec:perf_break_etypes}.
Finally, we provide an in-depth analysis of model performance across important data factors, such as international vs. domestic collaborations, international capability development, collaboration and partnership behavior of scientific
elites, cross-disciplinary collaborations, and
industry vs. academic partnerships, to better understand how and
why DGT models behave in a certain way in the supplementary materials.

\subsection{Experimental Setup}
\label{sec:exp_setup}
Training, validation, and test data consist of temporal edge lists in the format of quadruplets.
These quadruplets contain the head node, relation type, tail node, and  timestamp.
Given a head node and relation type, the model predicts the tail node at a given timestep.
For example, the model ranks all other scientists that would collaborate with a given scientist in a future timestep.
Note that for comprehensive evaluation, we  report model predictions for the head node given the tail node and the relation type at a given timestep.
We split the dataset by timestamp to have nonoverlapping records between training, validation, and testing splits as shown in Table~\ref{tab:data_char}.

\begin{table}[htbp]
    \centering
    \begin{tabular}{|c|l|r|r|r|}
    \hline
         \textit{Dataset} & \textit{Split} & \textit{Time} & \textit{\# Nodes, K} & \textit{\# Edges, K} \\ \hline \hline
         \multirow{3}{*}{\textbf{ACL}} & Training & 2010-2018 & 33 & 335 \\ \cline{2-5}
          & Validation & 2019 & 12  & 79 \\ \cline{2-5}
          & Testing & 2020-2021 & 10 & 64 \\ \hline
         \multirow{3}{*}{\textbf{ML}} & Training & 2010-2019 & 31 & 119 \\ \cline{2-5}
          & Validation & 2020 & 15 & 57 \\\cline{2-5}
          & Testing & 2021 & 8 & 34 \\ \hline
         \multirow{3}{*}{\textbf{WoS}} & Training & 2015-2018 & 17 & 456 \\ \cline{2-5}
          & Validation & 2019 & 10 & 121\\\cline{2-5}
          & Testing & 2020-2021 & 8 & 71 \\ \hline
         \multirow{3}{*}{\textbf{Scopus}} & Training & 2015-2018 & 9 & 300 \\ \cline{2-5}
          & Validation & 2019 & 5 & 85 \\\cline{2-5}
          & Testing & 2020-2021 & 4 & 46  \\ \hline
    \end{tabular}
    \caption{Characteristics of AI and NN datasets used in our experiments for training, validation, and testing.}
    \label{tab:data_char}
\end{table}

\subsubsection{Evaluation Metrics}
Forecasting temporal edges presents a much harder challenge due to the number of all possible candidate nodes.
For example, the model needs to evaluate all tail nodes given the head node in a quadruplet and repeat the process across all timesteps in the testing period.
Similar to previous work~\cite{horawalavithana2022expert}, we follow a standard protocol of evaluating the link prediction performance by limiting the evaluation to a set of candidate nodes~\cite{hu2021ogb}.
For each validation or test quadruplet, we perturb the tail node with 200 randomly sampled entities that do not appear in any of the training, validation, or test sets. 
Thus, models rank 201 candidates (consisting of 1 positive and 200 negative candidates) for each quadruplet.
We use these ranks to calculate  top $k$  positive candidates among the corresponding negative candidates (Hits@K) as well as the mean reciprocal rank (MRR) metrics.

\subsubsection{Baselines}
\label{sec:model_variants}
We implement the eight most representative state-of-the-art GNN baseline models, focusing on two main approaches: shallow and compositional encoding models. Shallow encoding models map each entity to a unique embedding vector. These methods rely on embedding lookup during  inference, and can only make predictions for the nodes observed during training.
For example, TransE~\cite{bordes2013translating} and RotatE use head-to-tail node relations to compute the plausibility of triples based on a distance function (\eg Euclidean distance between entities).
ConvE and ComplEx~\cite{trouillon2016complex} exploit similarity of latent features. RGCN~\cite{schlichtkrull2018modeling} uses a Graph Convolutional Network-based entity and relation encoder to learn entity representations.
NodePiece~\cite{galkin2021nodepiece} represents each node as a set of top-k nearest anchor nodes and $m$ unique relation types around the node. Anchor nodes and relation types are encoded in a node representation that can be used in any downstream prediction task for any entity, including those unobserved during training. 


        
        

    

For baseline experiments we use the Pykeen library~\cite{ali2021pykeen} and construct a static graph from the training data. We follow the best practices introduced in the Pykeen library (\eg training approach, loss function, and the explicit modeling of inverse relations)~\cite{ali2020bringing}. For example, we follow \textit{stochastic local closed world assumption}, where a random candidate set of triplets that are not part of the graph are considered as negative candidates.

In addition, we compare our proposed model performance with the original RE-Net and TGN models.
Compared to our DGT model, these models rely on  gated recurrent unit sequence layers instead of Transformer layers. For the RE-Net model and its variants, we supply a batch size of 1,024, a hidden dimension of 200, and maintain default parameters listed in the original paper~\cite{jin2019recurrent}.
For the TGN model and its variants, we supply a batch size of 200 and a memory dimension of 172.
All other parameters are unchanged from the original TGN paper~\cite{rossi2020temporal}.

\begin{figure*}[htbp]
\begin{subfigure}{.48\textwidth}
  \centering
  \includegraphics[width=.98\linewidth]{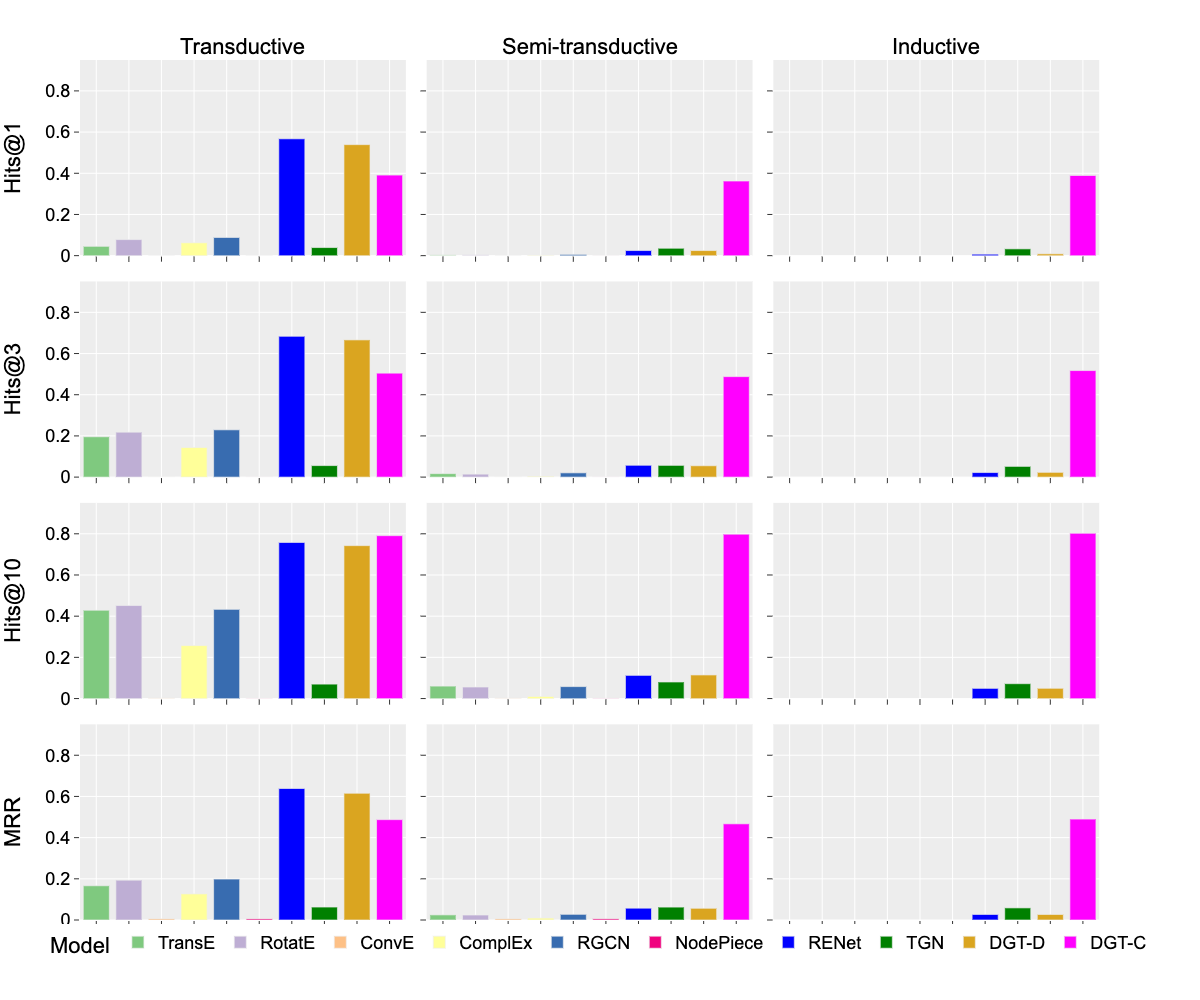}
  \caption{ACL (29\% transductive, 35\% semi-transductive, 36\% inductive)}
  \label{fig:performance_transductive_inductive_ACL}
\end{subfigure}
\begin{subfigure}{.48\textwidth}
  \centering
  \includegraphics[width=.98\linewidth]{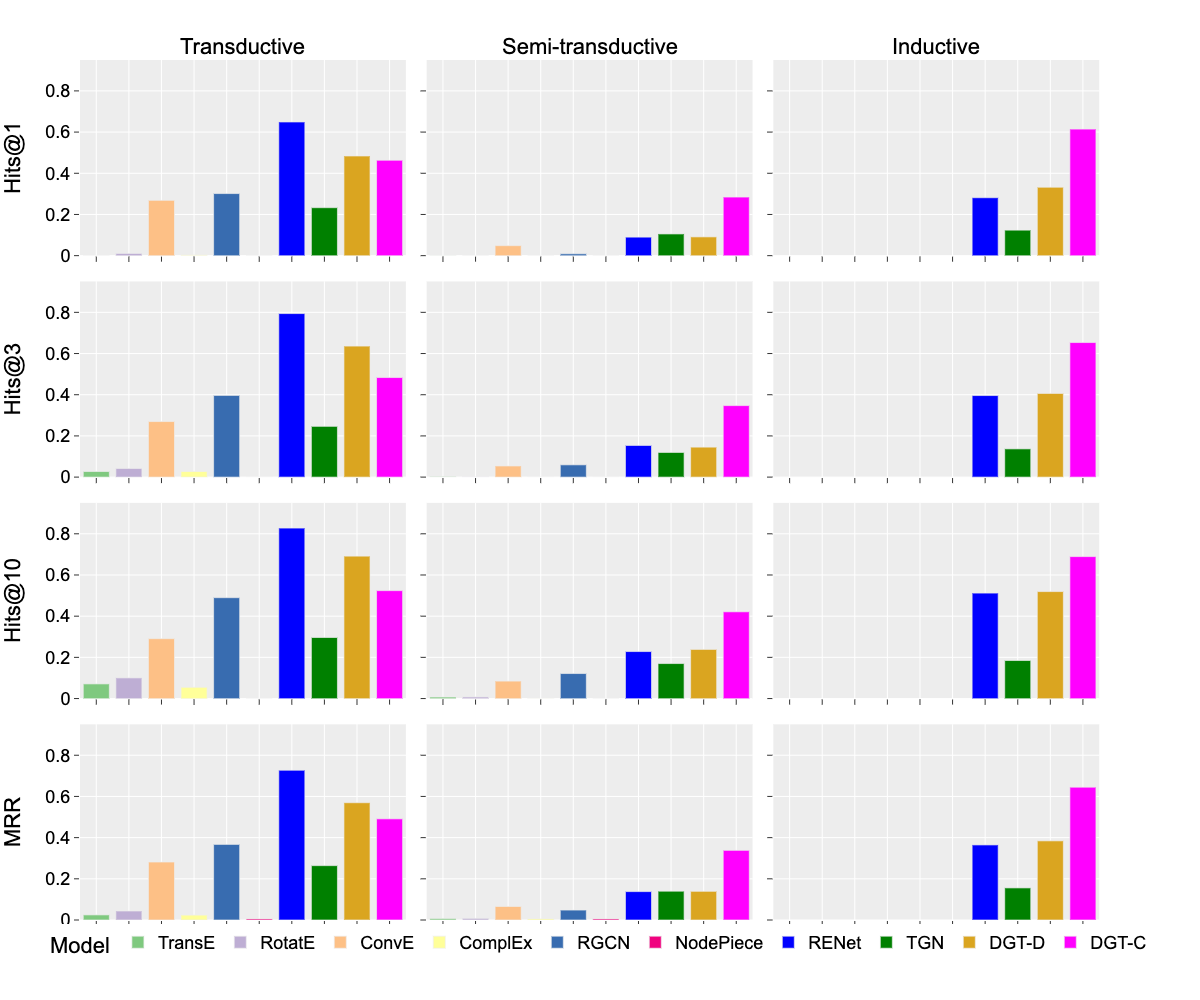}
  \caption{ML (4\% transductive, 6\% semi-transductive, 90\% inductive)}
  \label{fig:performance_transductive_inductive_Combined}
\end{subfigure}
\begin{subfigure}{.48\textwidth}
  \centering
  \includegraphics[width=.98\linewidth]{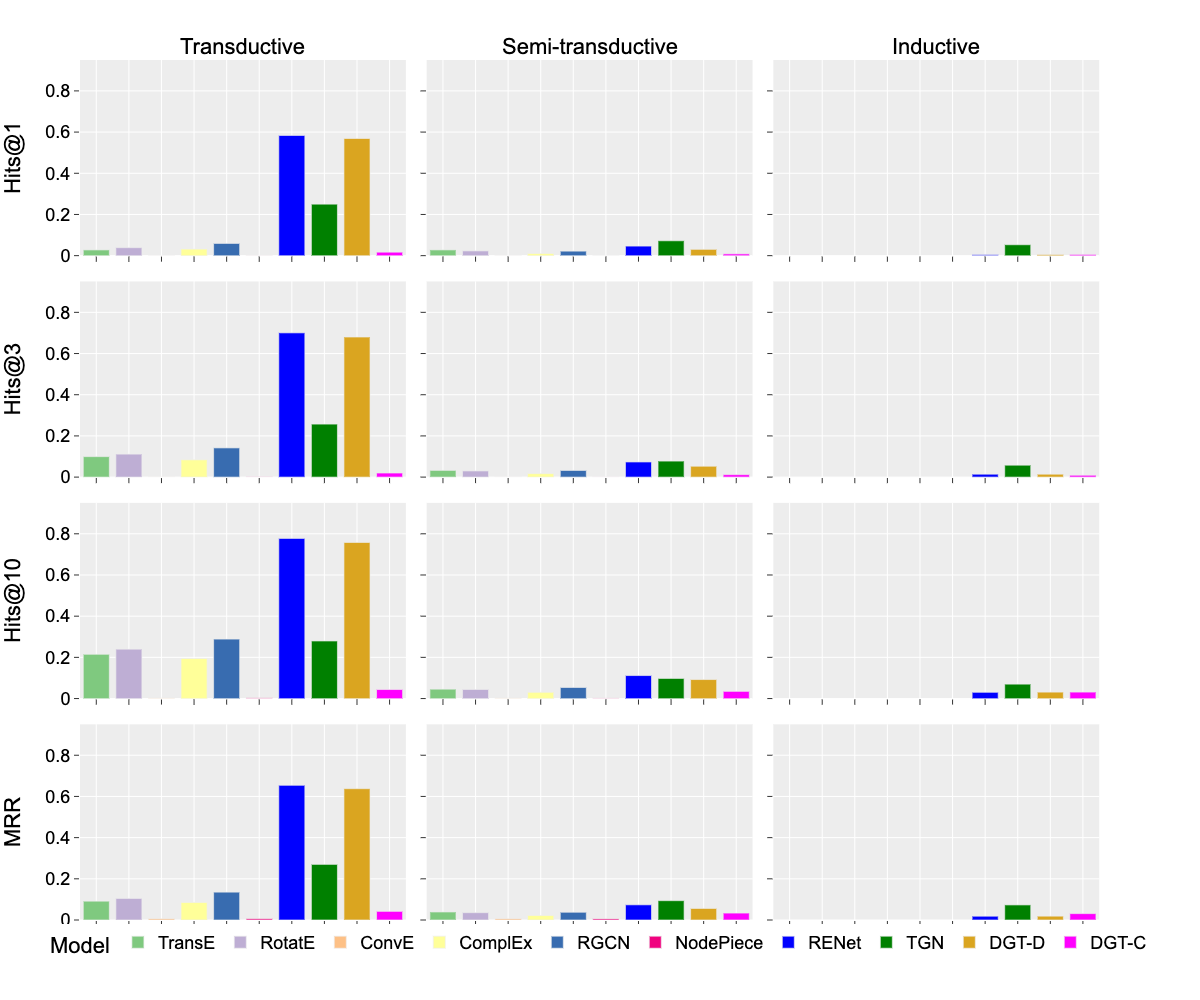}
  \caption{WoS (45\% transductive, 48\% semi-transductive, 7\% inductive)}
  \label{fig:performance_transductive_inductive_WOS}
\end{subfigure}
\begin{subfigure}{.48\textwidth}
  \centering
  \includegraphics[width=.98\linewidth]{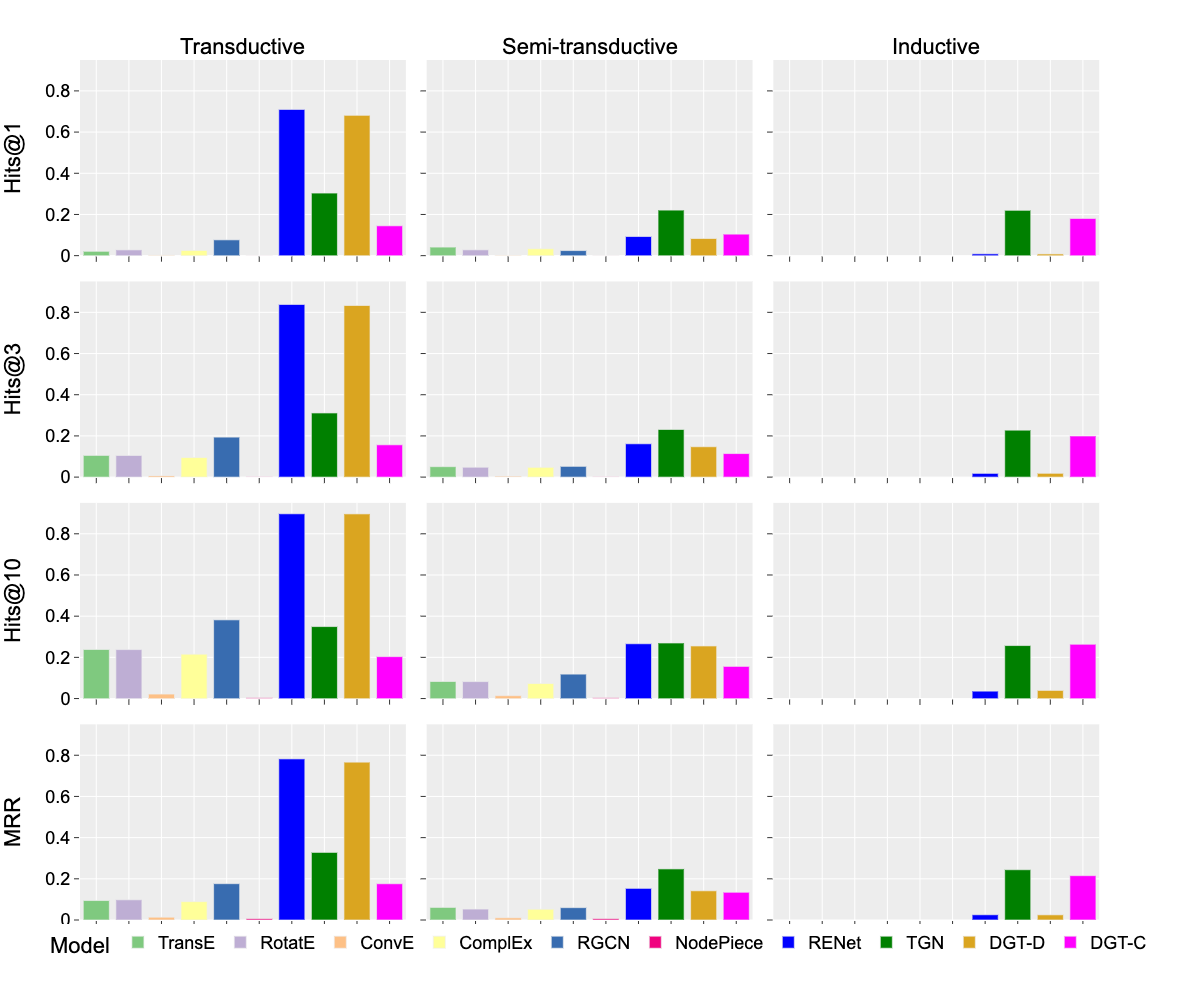}
  \caption{Scopus (63\% transductive, 31\% semi-transductive, 5\% inductive)}
  \label{fig:performance_transductive_inductive_Scopus}
\end{subfigure}
\caption{Forecasting performance in transductive and inductive settings. We measure the performance on predicting edges in three groups: 1) full-transductive (seen nodes with repeated interactions); 2) semi-transductive (seen nodes with first-time interactions); and 3) inductive (interactions with an unseen node). The task complexity increases in the same order. Note that the missing bars in the inductive task are due to the static graph baseline methods that are not able to perform the inductive tasks. Percentages of inductive and transductive edges are shown on the figure titles for each dataset.
}
\label{fig:performance_transductive_inductive}
\end{figure*}

\subsection{Transductive vs. Inductive Tasks}
\label{sec:tran_induct_tasks}
We evaluate model performance separately on edges between nodes that are \textit{observed} during training (transductive setting), and edges between at least one \textit{unobserved} node (inductive setting).
We group edges in the test data into these categories.
In  a transductive setting, edges connect two nodes observed in both training and testing data. We group  transductive edges into the ``first-time" and ``repeated" categories based on edge frequency.
We evaluate transductive performance with repeated edges, and  semi-transductive performance with first-time edges~\cite{horawalavithana2022expert}.
In an inductive setting, there is at least one "unseen" node for every edge.
For example, a graduate student ("unseen") can publish her first paper in the CL community with her mentor ("seen").
We use these edge groups to report the forecasting performance in both
settings as shown in Figure~\ref{fig:performance_transductive_inductive}.
Dynamic graph models (RE-Net, TGN, and DGT variants) perform significantly better than the static graph baseline models on forecasting temporal links.
For the example, DGT models achieve 30\%-80\% performance benefit across AI and NN domains over the static graph baseline models.
This confirms the previous findings~\cite{jin2019recurrent,kazemi2020representation} that dynamic graph information is important for forecasting links.
However, the performance varies across different dynamic graph models across transductive and inductive tasks.
For example, DGT-D model variants consistently perform well on full-transductive tasks than on inductive tasks.
Full-transductive tasks may be easier to predict than inductive tasks since the model can repeat the regular patterns observed in the training data.
On the other hand, DGT-C models perform significantly well on the inductive tasks, especially in AI datasets where inductive edges are the majority (Figures~\ref{fig:performance_transductive_inductive_ACL} and~\ref{fig:performance_transductive_inductive_Combined}).
Note that inductive tasks require predictions for unseen nodes.
The model achieves 0.64 MRR in the ML dataset, which outperforms the best-performing baseline model with 78\% performance advantage.
This shows that the Transformer models trained with continuous-time graph models can generalize into unseen nodes and edges.




\subsection{Performance Analysis}
\label{sec:perf_break_etypes}
In this section, we report the performance of models across collaboration, partnership, and capability edge types.
We focus on the transductive edges for this analysis to make a fair performance comparison across datasets.

\subsubsection{Forecasting Collaboration Patterns}
\begin{figure*}[htbp]
\begin{subfigure}{.48\textwidth}
  \centering
  \includegraphics[width=.98\linewidth]{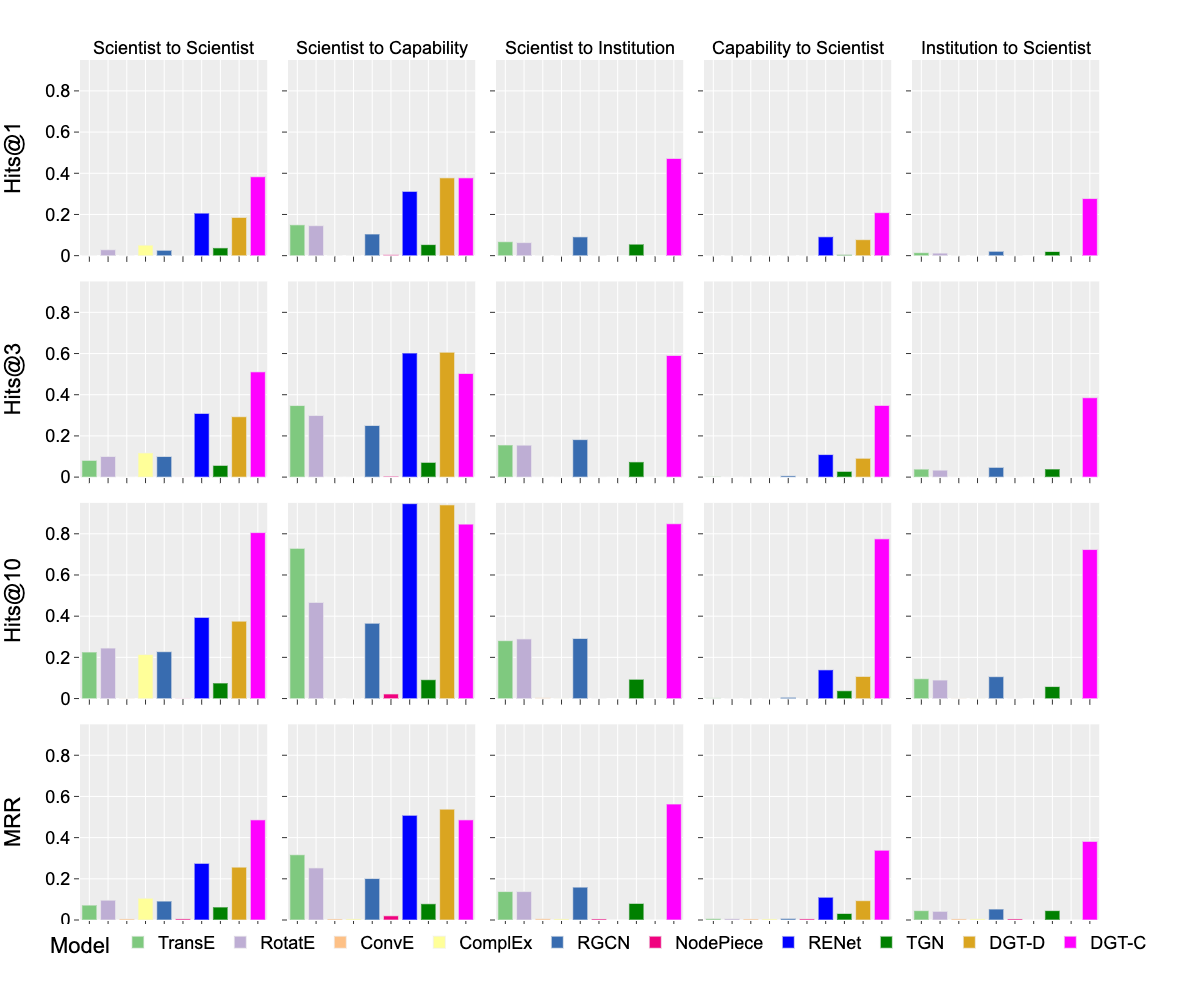}
  \caption{ACL}
  \label{fig:performance_main_ACL}
\end{subfigure}
\begin{subfigure}{.48\textwidth}
  \centering
  \includegraphics[width=.98\linewidth]{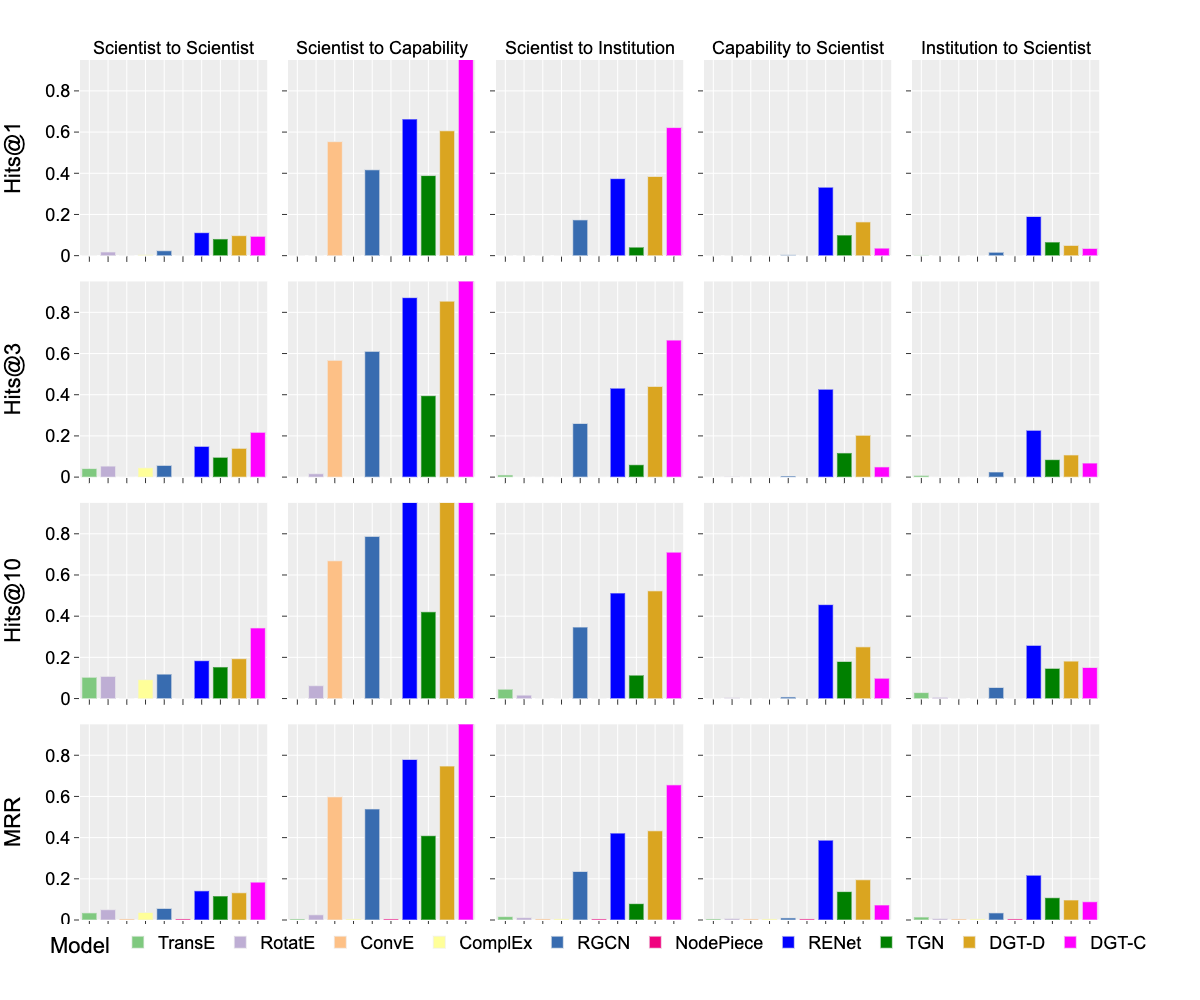}
  \caption{ML}
  \label{fig:performance_main_Combined}
\end{subfigure}
\begin{subfigure}{.48\textwidth}
  \centering
  \includegraphics[width=.98\linewidth]{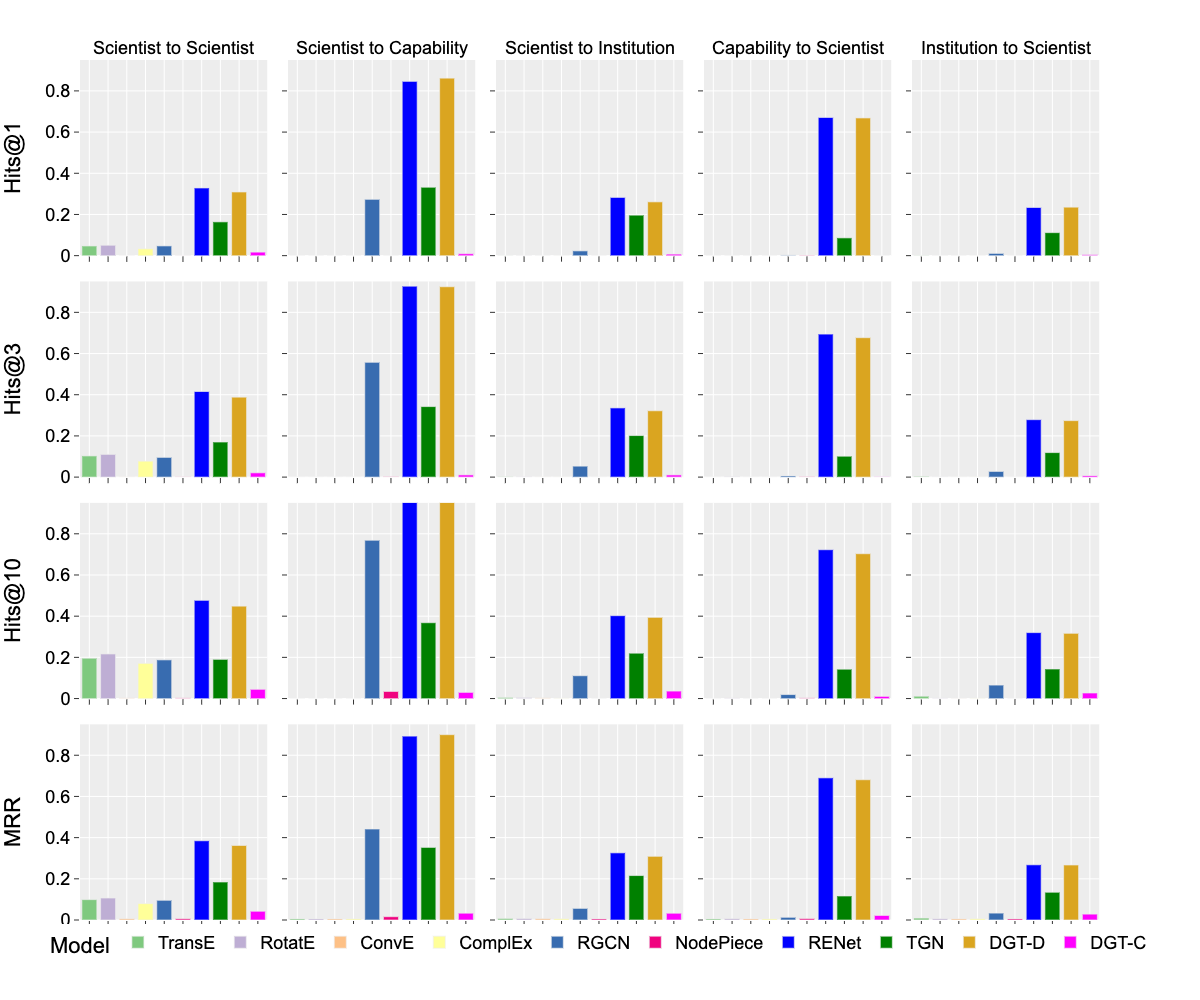}
  \caption{WoS}
  \label{fig:performance_main_WOS}
\end{subfigure}
\begin{subfigure}{.48\textwidth}
  \centering
  \includegraphics[width=.98\linewidth]{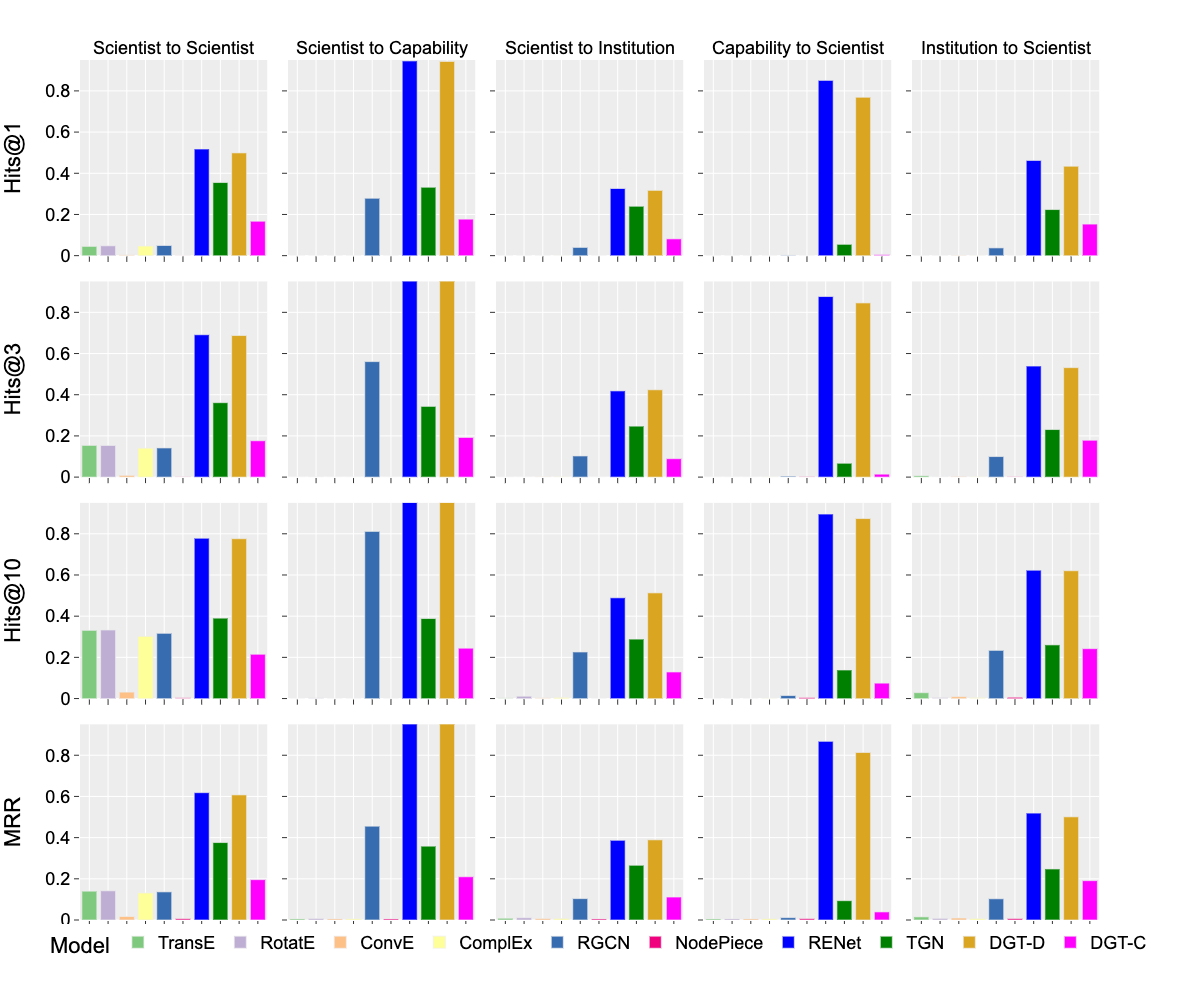}
  \caption{Scopus}
  \label{fig:performance_main_Scopus}
\end{subfigure}
\caption{Transductive forecasting performance breakdown by different directed edge types.
}
\label{fig:performance_main}
\end{figure*}

We report the performance of forecasting collaboration links across a group of scientists.
Our objective is to answer the following research questions: \textit{What scientist will a given scientist collaborate with next?} 
\textit{Which veteran scientist collaborates with an early career scientist?}
\textit{Which groups of scientists collaborate repeatedly?}
\textit{Which collaborations occur within tightly connected groups of scientists?}
Figures~\ref{fig:performance_main_ACL}-~\ref{fig:performance_main_Scopus}  report the performance of DGT model variants across the AI and NN domains.

First, the DGT-C model outperforms the baseline models on the AI datasets.
The model achieves 0.18 MRR on forecasting ML collaborations that outperforms the best-performing baseline with nearly 50\% performance advantage (Figure~\ref{fig:performance_main_Combined}).
More importantly, the DGT-C model is able to predict the incumbent scientists who collaborate with newcomer scientists more accurately than the baseline models.
We see 96\% of the newcomer-to-newcomer collaborations across ICLR, ICML, and NeurIPs communities in the testing period, while 20\% of collaborations between incumbent and newcomer scientists appear in the testing period.
For example, Sergey Levine, Percy Liang, and Zhuoran Yang are top-three scientists that publish in ML venues with newcomer scientists.
The DGT-C model achieves 0.11 MRR and 0.3 Hits@10 for predicting such ML collaborations.
The DGT-C performs similarly in the ACL dataset, with 78\% performance advantage over the best-performing baseline (Figure~\ref{fig:performance_main_ACL}).
DGT-C achieves 0.34 MRR and 0.72 Hits@10 for predicting incumbent-newcomer collaborations in the ACL dataset.

Second, the DGT-D and RE-Net models achieve the best performance in the WoS and Scopus datasets.
For example, the DGT-D model achieves 0.36 MRR and 0.45 Hits@10 in the WoS dataset.
The collaborations in the AI domain are more frequently emerging than in the NN domain (see the supplementary materials for more details).
Of NN collaborations, 99\% occurred across a group of incumbent scientists and the majority of such collaborations are repeated.
For example, the model is able to predict a group of Japanese scientists who repeatedly publish in the WoS dataset.
Hiroyoshi Sakurai is the most active scientist appearing in the testing period who repeatedly published with other scientists.

We believe the performance differences across the DGT-C and DGT-D variants are mainly due to the discrete- and continuous-time granularity of the graph inputs in the respective model variants.
Models may capture different temporal graph characteristics when trained with such different graph inputs.
This performance difference also explains the generalization ability of the DGT-C model, which is implemented with the Transformer architecture to handle continuous-time graph inputs.



\begin{figure*}[!t]
    \begin{subfigure}{.48\textwidth}
        \centering
        \includegraphics[width=0.95\textwidth]{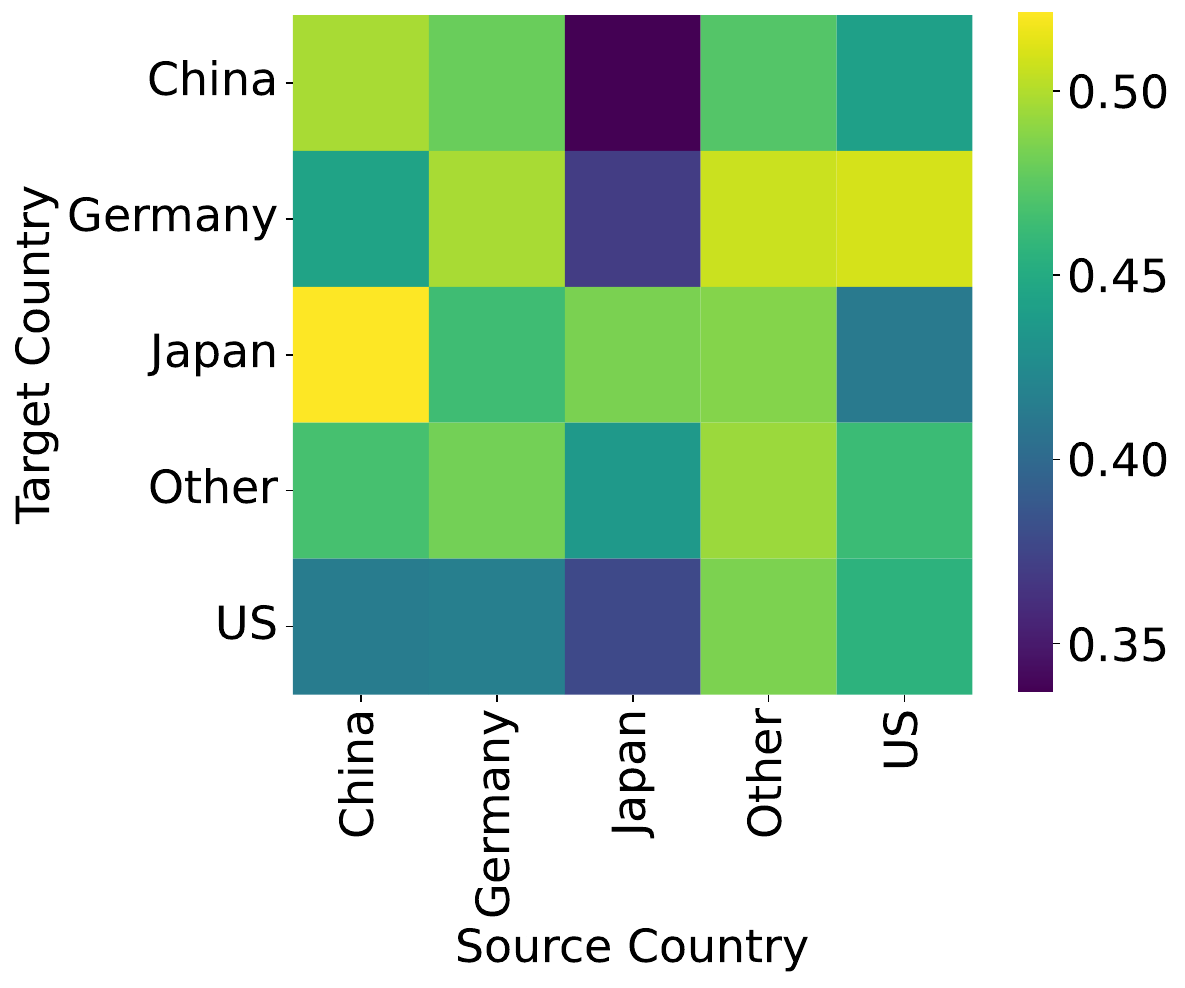}
        \caption{Collaborations (ACL)}
        \label{fig:acl_collab}
    \end{subfigure}
    \begin{subfigure}{.48\textwidth}
        \centering
        \includegraphics[width=0.95\textwidth]{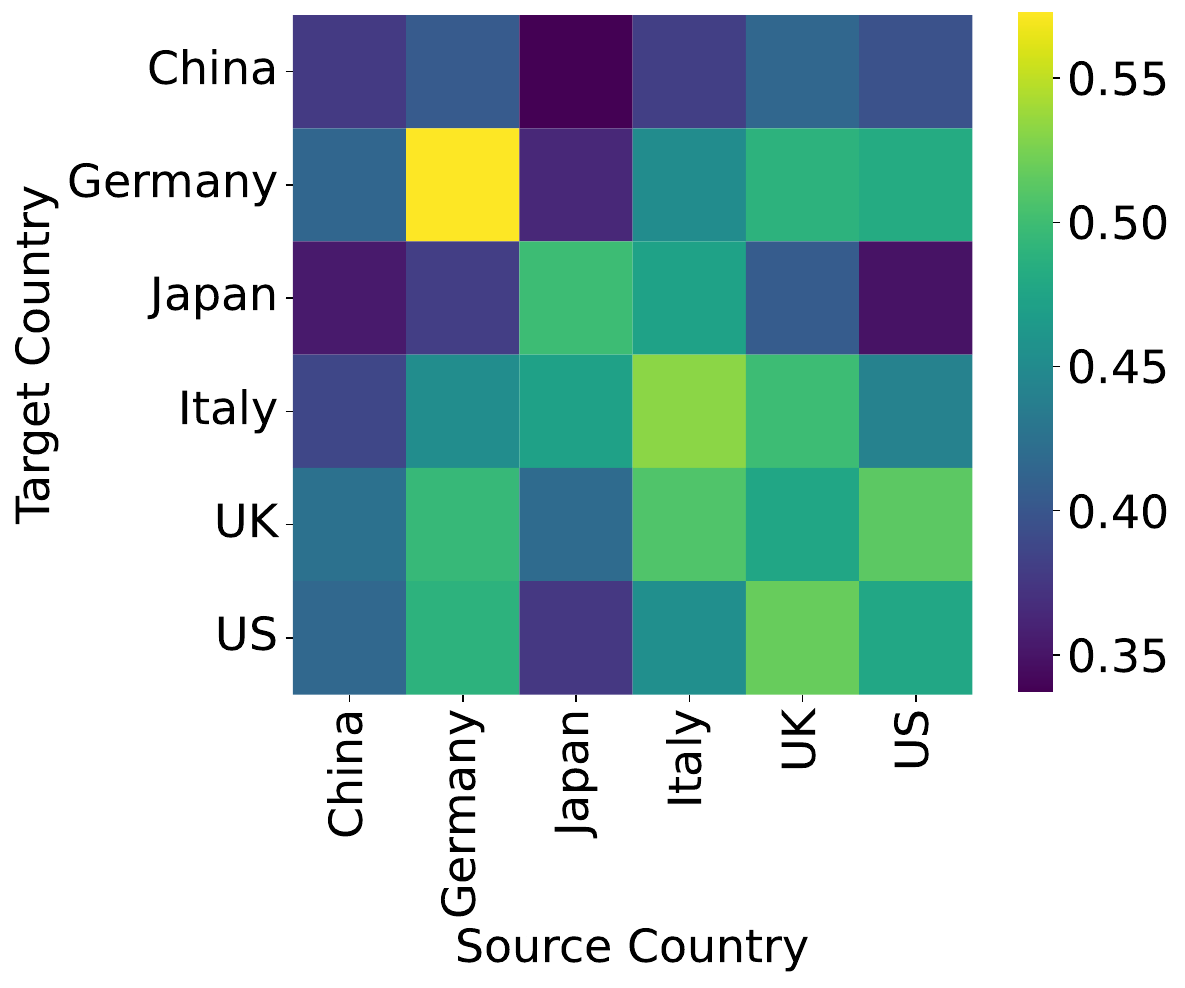}
        \caption{Collaborations (WoS)}
        \label{fig:wos_collab}
    \end{subfigure}\\
    \begin{subfigure}{.48\textwidth}
        \centering
        \includegraphics[width=0.95\textwidth]{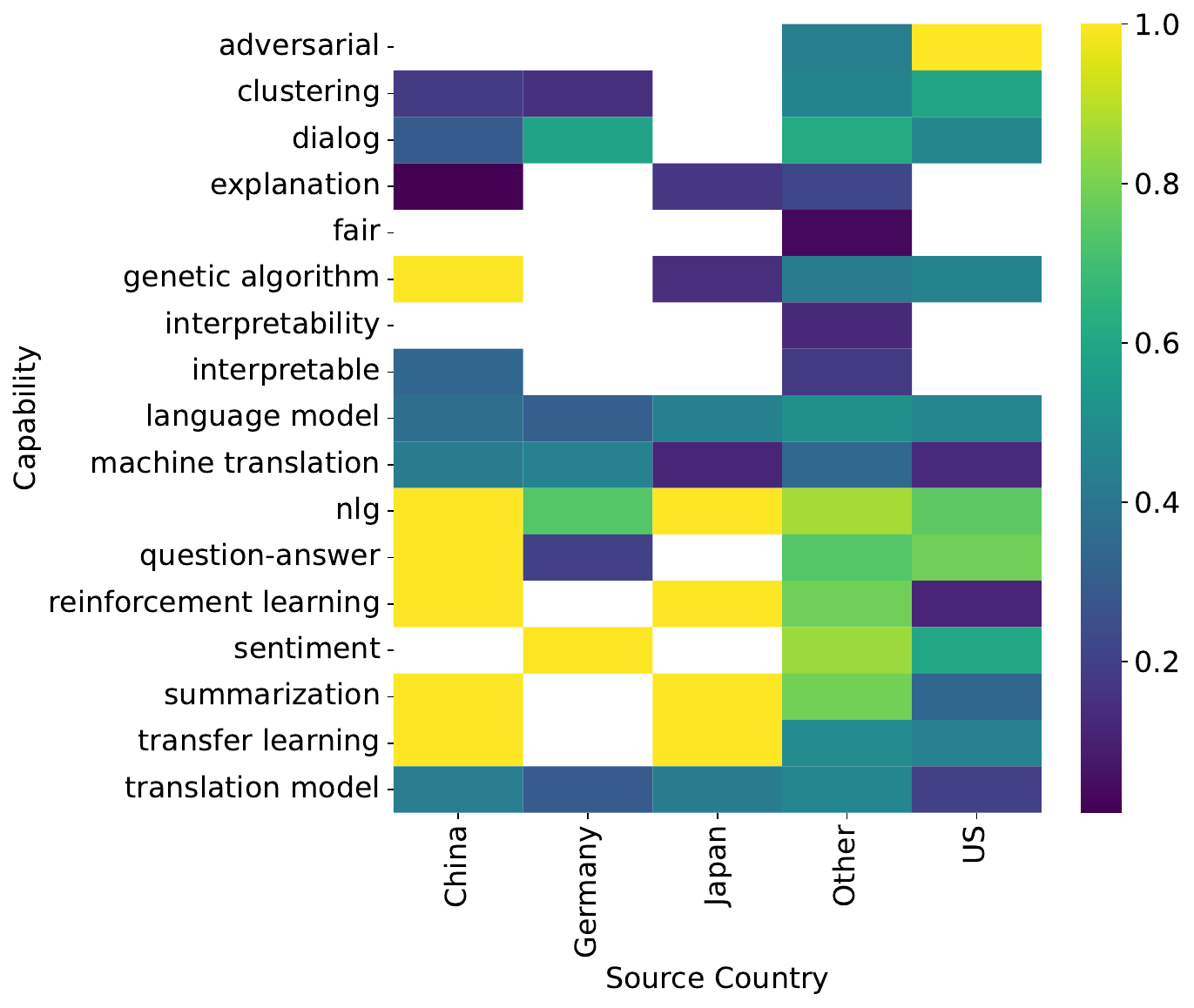}
        \caption{Technical Expertise (ACL)}
        \label{fig:acl_cap}
    \end{subfigure}
    \begin{subfigure}{.48\textwidth}
        \centering
        \includegraphics[width=0.95\textwidth]{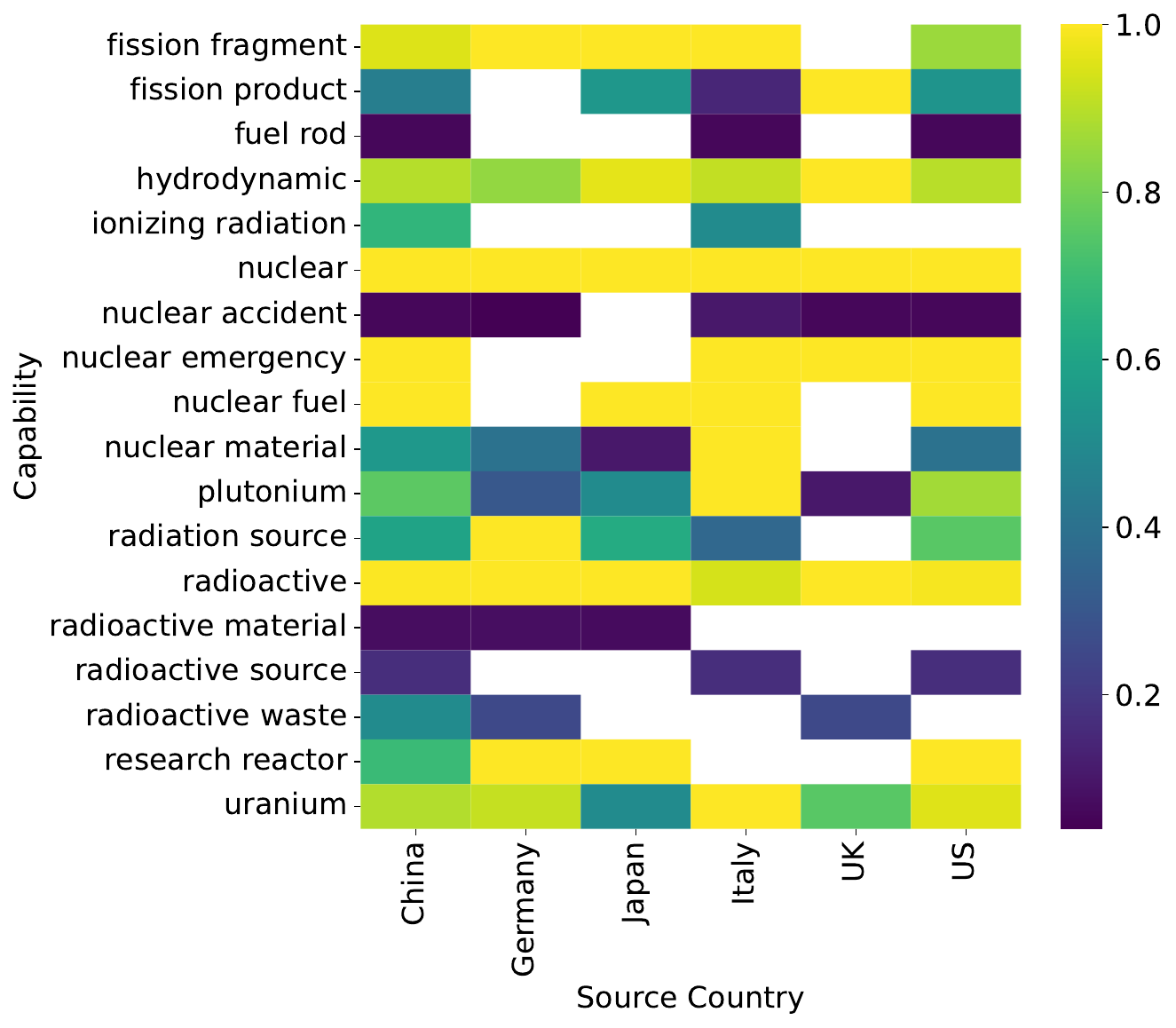}
        \caption{Technical Expertise (WoS)}
        \label{fig:wos_cap}
    \end{subfigure}
    \caption{Model performance (MRR) of collaboration edges between the five most frequent countries and capabilities. Left: best model performance for collaborations between countries of scientist-scientist and scientist-institution edges. Right: best model performance for scientist-capability edges. DGT-C and DGT-D are the best-performing models for ACL and WoS datasets, respectively. Blank cells indicate no edge between entities.}
\end{figure*}

\subsubsection{Forecasting Partnership Patterns}
In this section, we investigate the performance with respect to partnership edges, between scientists and institutions. Figure~\ref{fig:performance_main} reports the breakdown in performance across all datasets. 
Thus, we answer the following research questions: 
\textit{What institution will a given scientist partner with in a research collaboration?}
\textit{Who are the scientists partnered with an institution in the next publication?}
\textit{Are authors partnered with multiple institutions harder to forecast?}
\textit{How do the models perform on forecasting partnerships across large and small organizations?}

We find that in the case of both AI datasets, the best-performing model for partnership edges is DGT-C. 
All models also suffer a loss in performance when predicting on mirrored edges, in this case institution to scientist. When breaking down the edge-specific performance further into full-transductive, semi-transductive, and inductive edges, we find that the DGT-C outperforms all other models in all cases for the ACL dataset. 
We see the same pattern in the ML dataset, except for the case of full-transductive edges, where RE-Net outperforms DGT-C. 
The small percentage of full-transductive edges in the ML dataset may be the reason for this difference between overall and edge-type specific performance. 
Overall, the DGT-C is the only model to consistently perform well across all types of partnership edges in ACL and across the majority of partnership edges in the ML dataset, despite struggling to predict mirrored edges in the ML dataset.

In contrast, across both NN datasets, the best-performing models are the RE-Net and DGT-D. 
When breaking down the performance further into types of scientists and institutions, we notice  this performance increase only holds in the case of full-transductive partnerships. For semi-transductive and inductive edges, the TGN is the best-performing model. 
However, all models struggle to predict new partnership edges compared to the full-transductive task. 
Because there are so few inductive edges in the NN datasets, only the semi-transductive performance has a significant effect on the overall performance. 
Due to there being a majority of full-transductive edges in both NN datasets, this performance advantage by the TGN on semi-transductive edges does not hold for the overall performance. 
We also note that in the Scopus dataset, institution-to-scientist edges are easier to predict than scientist-to-institution edges. 

In addition, we conducted a fine-grained analysis on the best-performing model, DGT-C, on the ACL dataset. 
In order to answer the question of whether institution size has an effect on model performance across partnership edges, we ranked each institution by number of individual collaborators within the dataset. 
One thing to note is that the majority of the institutions in the ACL dataset have fewer than 10 collaborators. 
We found that the DGT-C achieved high performance (0.9 MRR or higher) among a set of smaller institutions with under 100 collaborators, namely INRIA, University of California, and Xi'an Jiaotong University. 
However, the hardest institutions to predict (0.04 MRR or lower) were also smaller institutions with under 100 collaborators, such as Institute for Human and Machine Cognition, University of Bucharest, and Universitat Politècnica de Catalunya. 
The largest institutions in the ACL dataset are Carnegie Mellon University, University of Edinburgh, and Peking University. 
The DGT-C model on average achieves 0.5 MRR for each of these institutions, which is the same as the overall average MRR for partnership edges on ACL. 
The DGT-C model does reliably perform well on larger institutions; however, because several smaller institutions have a high-performance advantage and several larger institutions have a performance disadvantage, it seems as though in many cases the size of the institution does not have a direct correlation on performance.  

\subsubsection{Forecasting Authorship Behavior}
In this section, we report the performance of forecasting links between scientists and capabilities.
We answer the following research questions: i) \textit{What is the next capability a scientist will publish on?} ii) \textit{Which scientists will publish on a given capability?} iii) \textit{Which capabilities are harder to forecast?}
To answer these questions, we predict either the head or tail node given a test quadruplet that consists of a scientist and a capability.
For example, we rank all candidate capabilities given a scientist, or rank all scientists given a capability.
Figures~\ref{fig:performance_main_ACL}-~\ref{fig:performance_main_Scopus} report the forecasting accuracy in the AI and NN datasets.
We have four observations from these figures.


First, the DGT-D model consistently outperforms the rest of the baselines on predicting authorship behavior in all datasets except ML.
DGT-D achieves 0.54 and 0.89 MRR in the ACL and WoS datasets, respectively.
We also noticed that DGT-D and RE-Net models have comparable performance across multiple metrics, but the DGT-D predictions are much closer to the ground truth.
For example, the model predicts the \textit{language model} and \textit{machine translation} as the most popular topics in the ACL dataset, while predicting the \textit{radioactive} and \textit{hydrodynamic} as the most popular topics in the WoS dataset.

Second, the DGT-C model has a performance advantage in the ML dataset (Figure~\ref{fig:performance_main_Combined}).
For example, the model achieves 0.98 MRR on predicting which ML capabilities scientists will publish on.
The model predicts many newcomer scientists publishing under \textit{reinforcement learning}, \textit{adversarial}, and \textit{language model} topics in the ML dataset.
The ML capability evolution has unique characteristics from the rest of the dataset as 84\% scientist-capability edges in the testing period are newly formed in contrast to
$<1\%$ similar edges 
in the WoS and Scopus datasets.
As we keep capabilities the same across both training and testing periods, the model is able to generalize the predictions for scientists unseen in the training period.

Third, we analyze how the models predict which scientists will publish on a given capability instead of the next capability a scientist will publish on.
We noticed that the DGT-D is the best-performing model in most of the cases (Figure~\ref{fig:performance_main_Combined},~\ref{fig:performance_main_WOS} and~\ref{fig:performance_main_Scopus}), but the DGT-C model predicts more accurately in the ACL dataset (Figure~\ref{fig:performance_main_ACL}).
This model achieves 0.34 MRR and 0.78 Hits@10 with a two to four times performance advantage over the best-performing baseline in the ACL dataset.
We see that the DGT-D model performs significantly well in the WoS and Scopus datasets.
For example, the model accurately predicts the most active scientists ({T. Hayat, A. Alsaedi, etc.}) who publish on the \textit{hydrodynamic} topic in the testing period.

Finally, we investigate which capabilities are more difficult to forecast than others.
We rank the capabilities from the best to worst forecasting accuracy and filter out the infrequent ones.
We noticed that \textit{nuclear, radioactive, fission fragment}, and \textit{hydrodynamic} are the best-performing capabilities, while \textit{nuclear accident,radioactive source, radioactive material}, and \textit{nuclear material} are the worst-performing capabilities in the WoS dataset.
Similarly, \textit{causal, adversarial}, and  \textit{transfer learning} and \textit{summarization, ethic}, and  \textit{natural language generation} are the sets of best- and worst-performing capabilities in the ML dataset.

\ignore{
We report an in-depth analysis of
model performance across important data factors, such as
international vs. domestic collaborations, international capability development, collaboration and partnership behavior of
scientific elites, cross-disciplinary collaborations, and industry
vs. academic partnerships, to better understand how and why
DGT models behave in a certain way in the supplementary
materials.
}

\subsection{Knowledge-Informed Performance Reasoning}


We examined international and domestic collaborations in two datasets: ACL and WoS as shown in Figures \ref{fig:acl_collab} and \ref{fig:wos_collab}.
Interactions between the same countries have different performances across the two domains, \eg China to Japan has a higher MRR value of 0.52 compared to 0.35 MRR from WoS. 
Interactions involving China are the hardest for the WoS DGT-D model to predict despite being the most frequent country.
In WoS, domestic collaborations (\ie interactions within the same country) achieve greater performance over international collaborations (see Figure \ref{fig:wos_collab}).
In contrast, model performance from the ACL community varies in both domestic and international collaborations.

Next, we looked at model performances across capabilities in the AI and NN domains. Figures \ref{fig:acl_cap} and \ref{fig:wos_cap} show the performance (MRR) on edges between the top five countries and capabilities (subsampled due to space) for ACL and WoS. In Figure \ref{fig:acl_cap}, near perfect MRR scores for China and Japan highlight the model's ability to forecast in those expertise areas, \ie natural language generation (nlg), reinforcement learning, summarization, and transfer learning. The 'Other' country category has interactions with all capabilities, but performance varies significantly. One reason for this is the uncertainty in country labeling. Potentially dozens of different countries are being represented by this one category.

\section{Summary and Discussions}

Forecasting technical expertise and capability evolution on national security domains would provide important clues for analysts to make informed decisions.
In this paper, we developed the DGT network, a novel deep-learning architecture that leverages attributes from both GNNs and transformer models, to anticipate technical expertise and capability evolution in critical national security domains like AI and NN.
To this end, we trained and evaluated eight DGT models from the publicly available digital scholarly data with complex relationships across scientists, institutions, and capabilities in both discrete and continuous time settings.
We made our graph datasets and code publicly available~\cite{horawalavithana2022expert}.

We show that DGT models perform well on inductive link forecasting tasks for the nodes unseen during the training.
While this is useful for analysts to detect the emerging scientists who work on operationally relevant disciplines, it is more challenging due to the lack of signals to track the involvement of new scientists.
DGT generalizes the patterns seen on the training data to detect which veteran scientists attract which new scientists and vice-versa.
We noticed models capturing different collaboration patterns across AI and NN domains.
For example, the models learn more from the tightly connected cliques of scientists in the NN domain than the hierarchical structure present in the AI domain.
We predict the collaboration patterns for highly influential scientists are more accurate than other scientists in AI datasets.
Our detailed performance analysis suggests that collaborations across scientists and institutions within the same country (domestic) are easier to anticipate than cross-country collaborations (international); collaboration patterns within the United States are easier to anticipate than those outside the United States, with collaborations from China being the most difficult to forecast in the NN domain.
Analysts can narrow down important scientists from a specific country who may start collaborating on important topics.

We also predict the research topics that scientists will tackle in the future.
For example, the models predict highly interdisciplinary scientists who work on multiple research topics such as nuclear, radioactive, fission fragment, etc. in the NN domain.
At the same time, models generalize predictions to AI scientists who focus on emerging research topics such as reinforcement learning, adversarial, and language model.
%
This provides important clues for analysts to detect impactful scientists who collaborate with people on new topics~\cite{pnas_impactsci}, or other scientists who continue publishing on the same topic.
This forecasting information would be useful to determine the direction in scientific discovery, especially for funding agencies to promote high-risk/high-reward projects testing unexplored hypothesis in national security domains~\cite{scisci_review}.
For example, the models predict that language modeling was one of the popular AI topics in 2020 through the patterns of collaboration and expertise in the training data before 2019.
Since 2020, language models revolutionized  AI research and advanced scientific breakthroughs in multiple domains such as chemistry, biology and security~\cite{bommasani2021opportunities,horawalavithana-etal-2022-foundation}.

\section*{Acknowledgments}
This work was supported by the NNSA Office of Defense Nuclear Nonproliferation Research and Development, U.S. Department of Energy, and Pacific Northwest National Laboratory, which is operated by Battelle Memorial Institute for the U.S. Department of Energy under Contract DE-AC05–76RLO1830.
This article has been cleared by PNNL for public release as PNNL-SA-181649.
Authors thank Sridevi Wagle, Shivam Sharma and Sannisth Soni for their help with preparing the datasets.

 
%

\bibliography{main}

\begin{thebibliography}{10}
\providecommand{\url}[1]{#1}
\csname url@samestyle\endcsname
\providecommand{\newblock}{\relax}
\providecommand{\bibinfo}[2]{#2}
\providecommand{\BIBentrySTDinterwordspacing}{\spaceskip=0pt\relax}
\providecommand{\BIBentryALTinterwordstretchfactor}{4}
\providecommand{\BIBentryALTinterwordspacing}{\spaceskip=\fontdimen2\font plus
\BIBentryALTinterwordstretchfactor\fontdimen3\font minus
  \fontdimen4\font\relax}
\providecommand{\BIBforeignlanguage}[2]{{%
\expandafter\ifx\csname l@#1\endcsname\relax
\typeout{** WARNING: IEEEtran.bst: No hyphenation pattern has been}%
\typeout{** loaded for the language `#1'. Using the pattern for}%
\typeout{** the default language instead.}%
\else
\language=\csname l@#1\endcsname
\fi
#2}}
\providecommand{\BIBdecl}{\relax}
\BIBdecl

\bibitem{national2021nuclear}
E.~National Academies~of Sciences, Medicine \emph{et~al.}, \emph{Nuclear
  Proliferation and Arms Control Monitoring, Detection, and Verification: A
  National Security Priority: Interim Report}, 2021.

\bibitem{wang2021science}
D.~Wang and A.-L. Barab{\'a}si, \emph{The science of science}.\hskip 1em plus
  0.5em minus 0.4em\relax Cambridge University Press, 2021.

\bibitem{horawalavithana2022expert}
S.~Horawalavithana, E.~Ayton, A.~Usenko, S.~Sharma, J.~Eshun, R.~Cosbey,
  M.~Glenski, and S.~Volkova, ``Expert: Public benchmarks for dynamic
  heterogeneous academic graphs,'' \emph{arXiv preprint arXiv:2204.07203},
  2022.

\bibitem{kazemi2020representation}
S.~M. Kazemi, R.~Goel, K.~Jain, I.~Kobyzev, A.~Sethi, P.~Forsyth, and
  P.~Poupart, ``Representation learning for dynamic graphs: A survey.''
  \emph{J. Mach. Learn. Res.}, vol.~21, no.~70, pp. 1--73, 2020.

\bibitem{pareja2020evolvegcn}
A.~Pareja, G.~Domeniconi, J.~Chen, T.~Ma, T.~Suzumura, H.~Kanezashi, T.~Kaler,
  T.~Schardl, and C.~Leiserson, ``Evolvegcn: Evolving graph convolutional
  networks for dynamic graphs,'' in \emph{Proceedings of the AAAI Conference on
  Artificial Intelligence}, vol.~34, no.~04, 2020, pp. 5363--5370.

\bibitem{goyal2020dyngraph2vec}
P.~Goyal, S.~R. Chhetri, and A.~Canedo, ``dyngraph2vec: Capturing network
  dynamics using dynamic graph representation learning,'' \emph{Knowledge-Based
  Systems}, vol. 187, p. 104816, 2020.

\bibitem{seo2018structured}
Y.~Seo, M.~Defferrard, P.~Vandergheynst, and X.~Bresson, ``Structured sequence
  modeling with graph convolutional recurrent networks,'' in
  \emph{International conference on neural information processing}.\hskip 1em
  plus 0.5em minus 0.4em\relax Springer, 2018, pp. 362--373.

\bibitem{defferrard2016convolutional}
M.~Defferrard, X.~Bresson, and P.~Vandergheynst, ``Convolutional neural
  networks on graphs with fast localized spectral filtering,'' \emph{Advances
  in neural information processing systems}, vol.~29, 2016.

\bibitem{kumar2019predicting}
S.~Kumar, X.~Zhang, and J.~Leskovec, ``Predicting dynamic embedding trajectory
  in temporal interaction networks,'' in \emph{Proceedings of the 25th ACM
  SIGKDD international conference on knowledge discovery \& data mining}, 2019,
  pp. 1269--1278.

\bibitem{jin2019recurrent}
W.~Jin, M.~Qu, X.~Jin, and X.~Ren, ``Recurrent event network: Autoregressive
  structure inference over temporal knowledge graphs,'' \emph{arXiv preprint
  arXiv:1904.05530}, 2019.

\bibitem{rossi2020temporal}
E.~Rossi, B.~Chamberlain, F.~Frasca, D.~Eynard, F.~Monti, and M.~Bronstein,
  ``Temporal graph networks for deep learning on dynamic graphs,'' \emph{arXiv
  preprint arXiv:2006.10637}, 2020.

\bibitem{cong2021dynamic}
W.~Cong, Y.~Wu, Y.~Tian, M.~Gu, Y.~Xia, M.~Mahdavi, and C.-c.~J. Chen,
  ``Dynamic graph representation learning via graph transformer networks,''
  \emph{arXiv preprint arXiv:2111.10447}, 2021.

\bibitem{horawalavithana-etal-2022-foundation}
\BIBentryALTinterwordspacing
S.~Horawalavithana, E.~Ayton, S.~Sharma, S.~Howland, M.~Subramanian,
  S.~Vasquez, R.~Cosbey, M.~Glenski, and S.~Volkova, ``Foundation models of
  scientific knowledge for chemistry: Opportunities, challenges and lessons
  learned,'' in \emph{Proceedings of BigScience Episode {\#}5 -- Workshop on
  Challenges {\&} Perspectives in Creating Large Language Models}.\hskip 1em
  plus 0.5em minus 0.4em\relax virtual+Dublin: Association for Computational
  Linguistics, May 2022, pp. 160--172. [Online]. Available:
  \url{https://aclanthology.org/2022.bigscience-1.12}
\BIBentrySTDinterwordspacing

\bibitem{khan2022transformers}
S.~Khan, M.~Naseer, M.~Hayat, S.~W. Zamir, F.~S. Khan, and M.~Shah,
  ``Transformers in vision: A survey,'' \emph{ACM computing surveys (CSUR)},
  vol.~54, no. 10s, pp. 1--41, 2022.

\bibitem{bommasani2021opportunities}
R.~Bommasani, D.~A. Hudson, E.~Adeli, R.~Altman, S.~Arora, S.~von Arx, M.~S.
  Bernstein, J.~Bohg, A.~Bosselut, E.~Brunskill \emph{et~al.}, ``On the
  opportunities and risks of foundation models,'' \emph{arXiv preprint
  arXiv:2108.07258}, 2021.

\bibitem{ying2021transformers}
C.~Ying, T.~Cai, S.~Luo, S.~Zheng, G.~Ke, D.~He, Y.~Shen, and T.-Y. Liu, ``Do
  transformers really perform badly for graph representation?'' \emph{Advances
  in Neural Information Processing Systems}, vol.~34, pp. 28\,877--28\,888,
  2021.

\bibitem{dwivedi2020generalization}
V.~P. Dwivedi and X.~Bresson, ``A generalization of transformer networks to
  graphs,'' \emph{arXiv preprint arXiv:2012.09699}, 2020.

\bibitem{vaswani2017attention}
A.~Vaswani, N.~Shazeer, N.~Parmar, J.~Uszkoreit, L.~Jones, A.~N. Gomez,
  {\L}.~Kaiser, and I.~Polosukhin, ``Attention is all you need,''
  \emph{Advances in neural information processing systems}, vol.~30, 2017.

\bibitem{zhang2020graph}
J.~Zhang, H.~Zhang, C.~Xia, and L.~Sun, ``Graph-bert: Only attention is needed
  for learning graph representations,'' \emph{arXiv preprint arXiv:2001.05140},
  2020.

\bibitem{kim2022pure}
J.~Kim, T.~D. Nguyen, S.~Min, S.~Cho, M.~Lee, H.~Lee, and S.~Hong, ``Pure
  transformers are powerful graph learners,'' \emph{arXiv preprint
  arXiv:2207.02505}, 2022.

\bibitem{kreuzer2021rethinking}
D.~Kreuzer, D.~Beaini, W.~Hamilton, V.~L{\'e}tourneau, and P.~Tossou,
  ``Rethinking graph transformers with spectral attention,'' \emph{Advances in
  Neural Information Processing Systems}, vol.~34, pp. 21\,618--21\,629, 2021.

\bibitem{min2022transformer}
E.~Min, R.~Chen, Y.~Bian, T.~Xu, K.~Zhao, W.~Huang, P.~Zhao, J.~Huang,
  S.~Ananiadou, and Y.~Rong, ``Transformer for graphs: An overview from
  architecture perspective,'' \emph{arXiv preprint arXiv:2202.08455}, 2022.

\bibitem{sankar2018dynamic}
A.~Sankar, Y.~Wu, L.~Gou, W.~Zhang, and H.~Yang, ``Dynamic graph representation
  learning via self-attention networks,'' \emph{arXiv preprint
  arXiv:1812.09430}, 2018.

\bibitem{xu2020inductive}
D.~Xu, C.~Ruan, E.~Korpeoglu, S.~Kumar, and K.~Achan, ``Inductive
  representation learning on temporal graphs,'' \emph{arXiv preprint
  arXiv:2002.07962}, 2020.

\bibitem{yang2015defining}
J.~Yang and J.~Leskovec, ``Defining and evaluating network communities based on
  ground-truth,'' \emph{Knowledge and Information Systems}, vol.~42, no.~1, pp.
  181--213, 2015.

\bibitem{hu2020open}
W.~Hu, M.~Fey, M.~Zitnik, Y.~Dong, H.~Ren, B.~Liu, M.~Catasta, and J.~Leskovec,
  ``Open graph benchmark: Datasets for machine learning on graphs,''
  \emph{arXiv preprint arXiv:2005.00687}, 2020.

\bibitem{gehrke2003overview}
J.~Gehrke, P.~Ginsparg, and J.~Kleinberg, ``Overview of the 2003 kdd cup,''
  \emph{Acm Sigkdd Explorations Newsletter}, vol.~5, no.~2, pp. 149--151, 2003.

\bibitem{hu2021ogb}
W.~Hu, M.~Fey, H.~Ren, M.~Nakata, Y.~Dong, and J.~Leskovec, ``Ogb-lsc: A
  large-scale challenge for machine learning on graphs,'' \emph{arXiv preprint
  arXiv:2103.09430}, 2021.

\bibitem{wang2020microsoft}
K.~Wang, Z.~Shen, C.~Huang, C.-H. Wu, Y.~Dong, and A.~Kanakia, ``Microsoft
  academic graph: When experts are not enough,'' \emph{Quantitative Science
  Studies}, vol.~1, no.~1, pp. 396--413, 2020.

\bibitem{shi2021r}
Y.~Shi, P.~Team, Z.~Huang, W.~Li, W.~Su, S.~Feng \emph{et~al.}, ``R-unimp:
  Solution for kdd cup 2021 mag240m-lsc,'' \emph{Open Graph
  Benchmark-Large-Scale Challenge@ KDD Cup 2021}, 2021.

\bibitem{zhaomdgnn}
H.~Zhao, Y.~Cen, Y.~He, Z.~Hou, X.~Liu, and X.~Cheng, ``Mdgnn: Metapath-based
  decoupled graph neural network for mag240m-lsc.''

\bibitem{addanki2021large}
R.~Addanki, P.~W. Battaglia, D.~Budden, A.~Deac, J.~Godwin, T.~Keck, W.~L.~S.
  Li, A.~Sanchez-Gonzalez, J.~Stott, S.~Thakoor \emph{et~al.}, ``Large-scale
  graph representation learning with very deep gnns and self-supervision,''
  \emph{arXiv preprint arXiv:2107.09422}, 2021.

\bibitem{zhang2022dynamic}
Z.~Zhang, X.~Wang, Z.~Zhang, H.~Li, Z.~Qin, and W.~Zhu, ``Dynamic graph neural
  networks under spatio-temporal distribution shift,'' in \emph{Advances in
  Neural Information Processing Systems}, 2022.

\bibitem{kazemi2019time2vec}
S.~M. Kazemi, R.~Goel, S.~Eghbali, J.~Ramanan, J.~Sahota, S.~Thakur, S.~Wu,
  C.~Smyth, P.~Poupart, and M.~Brubaker, ``Time2vec: Learning a vector
  representation of time,'' \emph{arXiv preprint arXiv:1907.05321}, 2019.

\bibitem{GROBID}
``Grobid,'' \url{https://github.com/kermitt2/grobid}, 2008--2022.

\bibitem{glenski2021identifying}
M.~Glenski and S.~Volkova, ``Identifying causal influences on publication
  trends and behavior: A case study of the computational linguistics
  community,'' \emph{arXiv preprint arXiv:2110.07938}, 2021.

\bibitem{visin2015renet}
F.~Visin, K.~Kastner, K.~Cho, M.~Matteucci, A.~Courville, and Y.~Bengio,
  ``Renet: A recurrent neural network based alternative to convolutional
  networks,'' \emph{arXiv preprint arXiv:1505.00393}, 2015.

\bibitem{bordes2013translating}
A.~Bordes, N.~Usunier, A.~Garcia-Duran, J.~Weston, and O.~Yakhnenko,
  ``Translating embeddings for modeling multi-relational data,'' \emph{Advances
  in neural information processing systems}, vol.~26, 2013.

\bibitem{trouillon2016complex}
T.~Trouillon, J.~Welbl, S.~Riedel, {\'E}.~Gaussier, and G.~Bouchard, ``Complex
  embeddings for simple link prediction,'' in \emph{International conference on
  machine learning}.\hskip 1em plus 0.5em minus 0.4em\relax PMLR, 2016, pp.
  2071--2080.

\bibitem{schlichtkrull2018modeling}
M.~Schlichtkrull, T.~N. Kipf, P.~Bloem, R.~v.~d. Berg, I.~Titov, and
  M.~Welling, ``Modeling relational data with graph convolutional networks,''
  in \emph{European semantic web conference}.\hskip 1em plus 0.5em minus
  0.4em\relax Springer, 2018, pp. 593--607.

\bibitem{galkin2021nodepiece}
M.~Galkin, J.~Wu, E.~Denis, and W.~L. Hamilton, ``Nodepiece: Compositional and
  parameter-efficient representations of large knowledge graphs,'' \emph{arXiv
  preprint arXiv:2106.12144}, 2021.

\bibitem{ali2021pykeen}
M.~Ali, M.~Berrendorf, C.~T. Hoyt, L.~Vermue, S.~Sharifzadeh, V.~Tresp, and
  J.~Lehmann, ``Pykeen 1.0: A python library for training and evaluating
  knowledge graph embeddings,'' \emph{Journal of Machine Learning Research},
  vol.~22, no.~82, pp. 1--6, 2021.

\bibitem{ali2020bringing}
M.~Ali, M.~Berrendorf, C.~T. Hoyt, L.~Vermue, M.~Galkin, S.~Sharifzadeh,
  A.~Fischer, V.~Tresp, and J.~Lehmann, ``Bringing light into the dark: A
  large-scale evaluation of knowledge graph embedding models under a unified
  framework,'' \emph{arXiv preprint arXiv:2006.13365}, 2020.

\bibitem{pnas_impactsci}
\BIBentryALTinterwordspacing
A.~Zeng, Y.~Fan, Z.~Di, Y.~Wang, and S.~Havlin, ``Impactful scientists have
  higher tendency to involve collaborators in new topics,'' \emph{Proceedings
  of the National Academy of Sciences}, vol. 119, no.~33, p. e2207436119, 2022.
  [Online]. Available:
  \url{https://www.pnas.org/doi/abs/10.1073/pnas.2207436119}
\BIBentrySTDinterwordspacing

\bibitem{scisci_review}
\BIBentryALTinterwordspacing
S.~Fortunato, C.~T. Bergstrom, K.~Börner, J.~A. Evans, D.~Helbing,
  S.~Milojević, A.~M. Petersen, F.~Radicchi, R.~Sinatra, B.~Uzzi,
  A.~Vespignani, L.~Waltman, D.~Wang, and A.-L. Barabási, ``Science of
  science,'' \emph{Science}, vol. 359, no. 6379, p. eaao0185, 2018. [Online].
  Available: \url{https://www.science.org/doi/abs/10.1126/science.aao0185}
\BIBentrySTDinterwordspacing

\bibitem{vlasceanu2022interdisciplinarity}
M.~Vlasceanu, M.~Dud{\'\i}k, and I.~Momennejad, ``Interdisciplinarity, gender
  diversity, and network structure predict the centrality of ai
  organizations,'' in \emph{2022 ACM Conference on Fairness, Accountability,
  and Transparency}, 2022, pp. 1--10.

\bibitem{industry_research_2018}
\BIBentryALTinterwordspacing
``The importance of industrial publications,'' \emph{Nature Catalysis}, vol.~1,
  no.~7, pp. 479--480, Jul. 2018. [Online]. Available:
  \url{https://doi.org/10.1038/s41929-018-0119-0}
\BIBentrySTDinterwordspacing

\end{thebibliography}
\bibliographystyle{IEEEtran}

{\appendices

\section{Data Preprocessing and Characteristics}
\label{sec:data_char}
 In this section we provide additional data characteristics about the types of scientists, institutions, and capabilities. Our objective is to demonstrate the differences across these two very distinct domains. This allows us to evaluate the generalization of the proposed models and measure the robustness of the models across the domains.
 
\subsection{Data Augmentation}
We performed additional preprocessing steps to support our analysis. First, we labeled scientists as incumbent or newcomer nodes based on their frequency of appearances in the dataset. For example, when a scientist published her first article in the computational linguistic (CL) community, we labeled her as a newcomer. Incumbent scientists published multiple times in the same community.
Similar to the above classification, we labeled the institutions into new and repeated partnerships.
For example, a scientist may establish a new partnership with an institution in the next publication, or might continue with the previously established partnership with an institution.
Finally, we characterized the publication trends on the adoption and persistence of certain research foci in the respective research communities~\cite{glenski2021identifying}.
When a scientist adopted a new research focus to work on for the first time, we labeled them as new capabilities.
Otherwise, a scientist maintained the same research focus across publications.

As a part of our metadata extraction process, we extracted location information directly from the PDF (available for the ACL and ML datasets) or citation records (available for the WoS and Scopus datasets). 
The quality of extracted location data varied greatly across the datasets. The manual inspection was done to verify valid country names. City, state, or province names were mapped to country names whenever possible (\eg Beijing to China). Those publications missing location information, or those containing incorrect location information, or location names that could not be mapped to a specific country were labeled as Other. Table \ref{tab:countries} lists the total number of countries extracted across the datasets.

\ignore{
 \begin{figure*}[htbp]
    \centering
    \includegraphics[width=.98\linewidth]{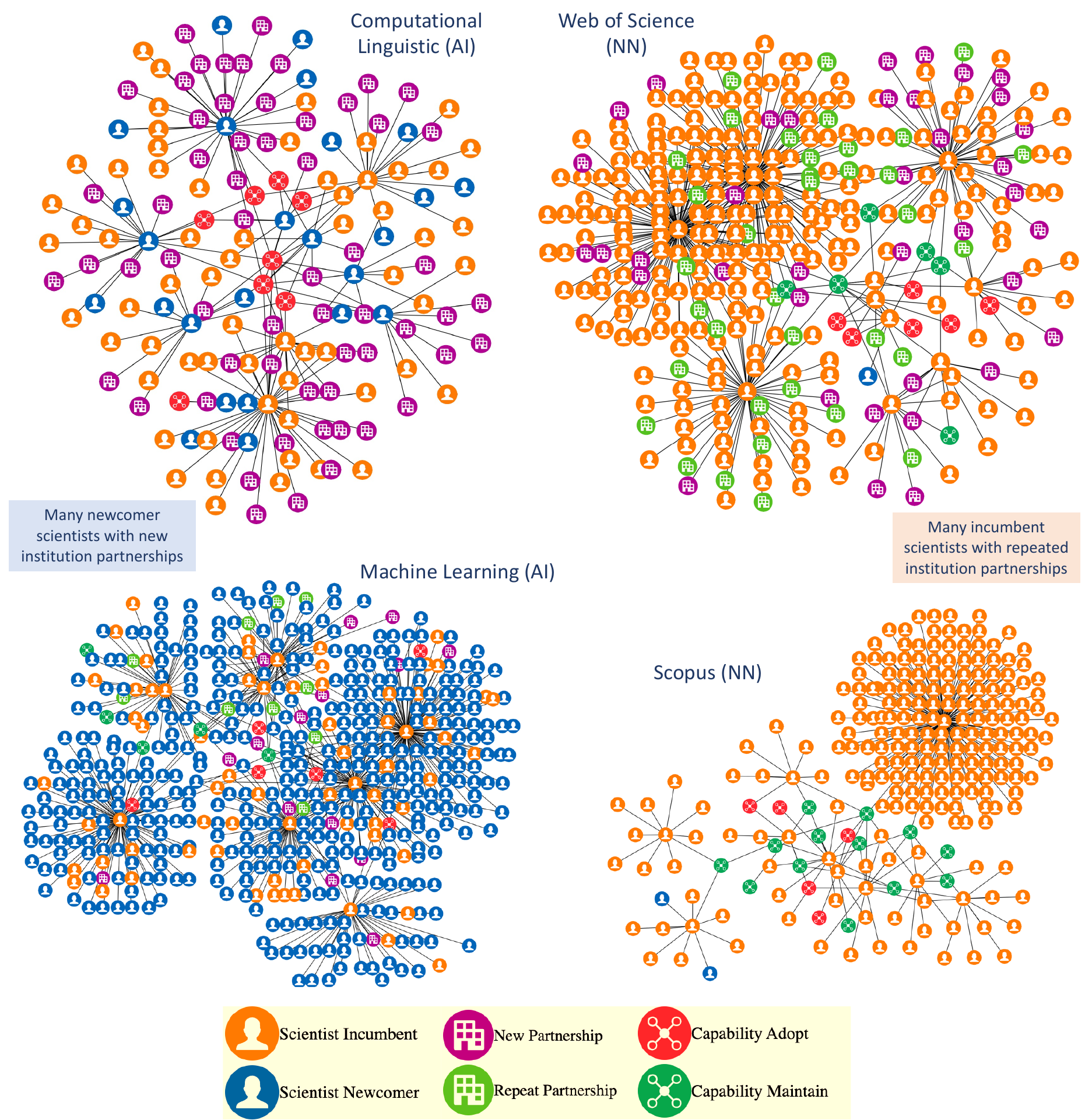}
    \caption{Visualization of the dynamic heterogeneous graphs across four datasets with diverse scientist, institution, and capability nodes with a focus on the subgraphs extracted from the top 10 most active scientists. Note, we removed all edge timestamps to reduce visual clutter.}
    \label{fig:network_viz}
\end{figure*}
}

\begin{figure*}[htbp]
\begin{subfigure}{.32\textwidth}
  \centering
  \includegraphics[width=.98\linewidth]{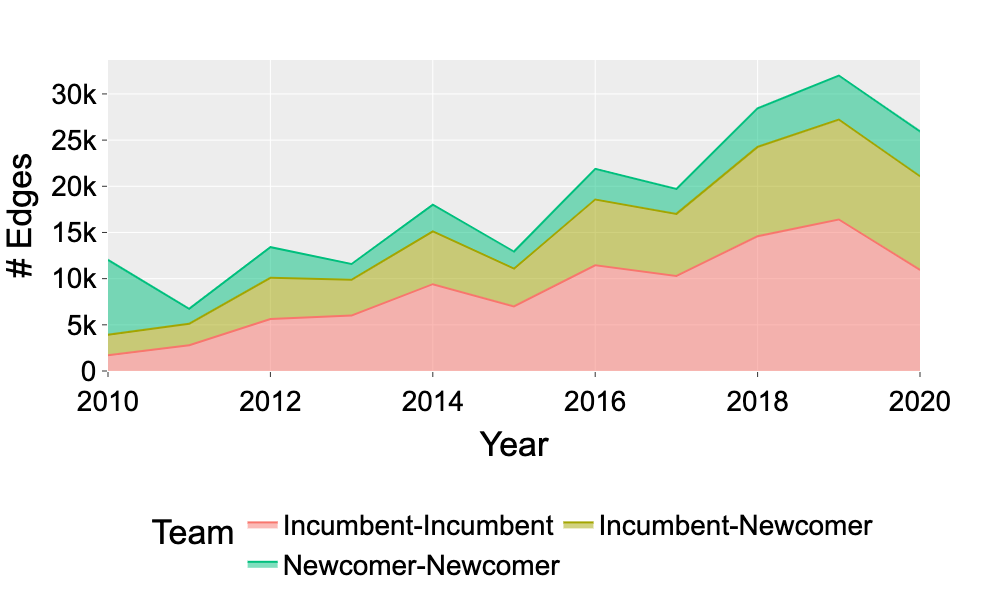}
  \caption{Collaborations (ACL)}
  \label{fig:sci_to_sci_acl}
\end{subfigure}
\begin{subfigure}{.32\textwidth}
  \centering
  \includegraphics[width=.98\linewidth]{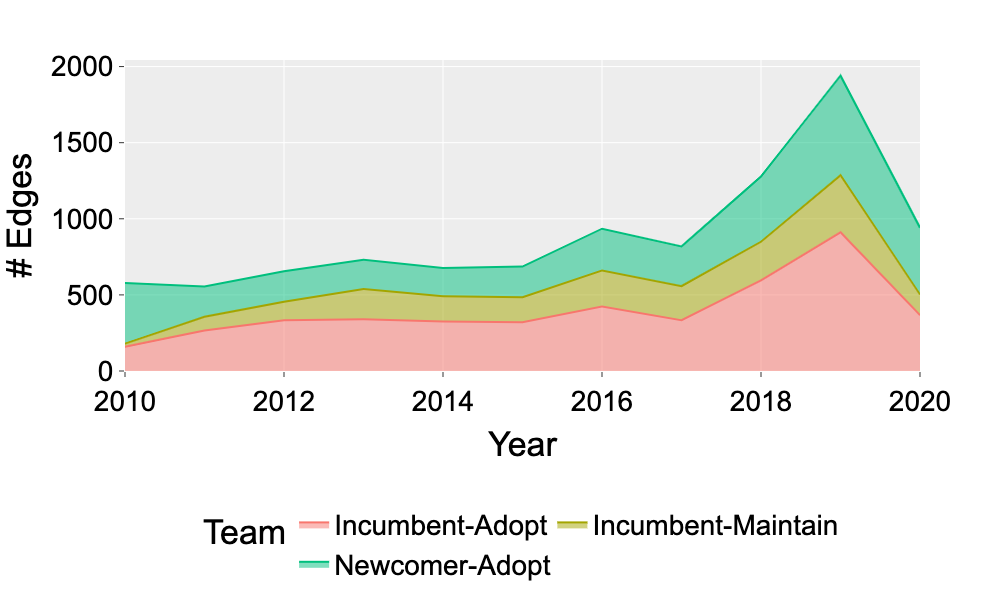}
  \caption{Capabilities (ACL)}
  \label{fig:sci_to_cap_acl}
\end{subfigure}
\begin{subfigure}{.32\textwidth}
  \centering
  \includegraphics[width=.98\linewidth]{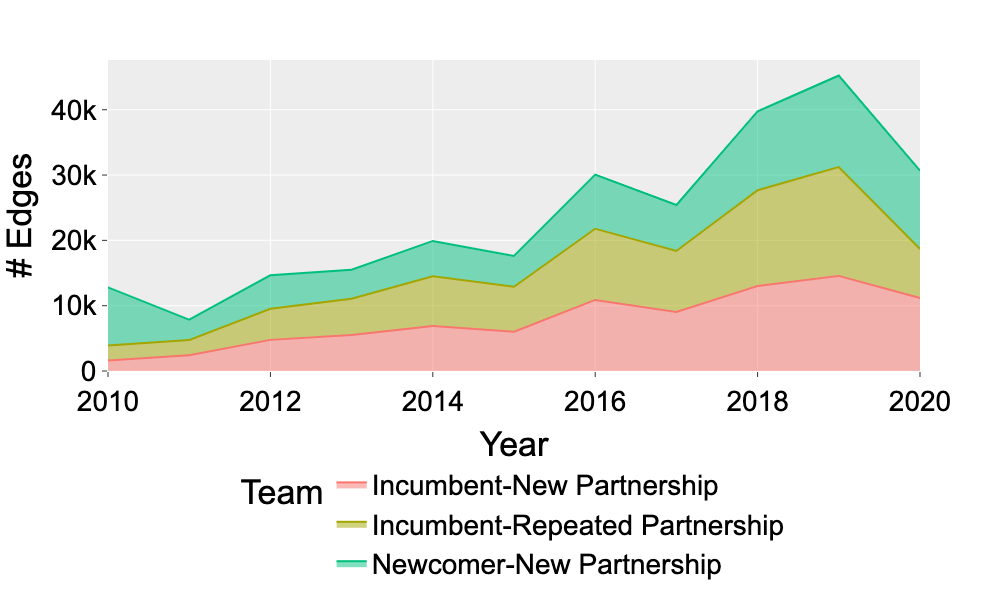}
  \caption{Partnerships (ACL)}
  \label{fig:sci_to_int_acl}
\end{subfigure}
\newline
\begin{subfigure}{.32\textwidth}
  \centering
  \includegraphics[width=.98\linewidth]{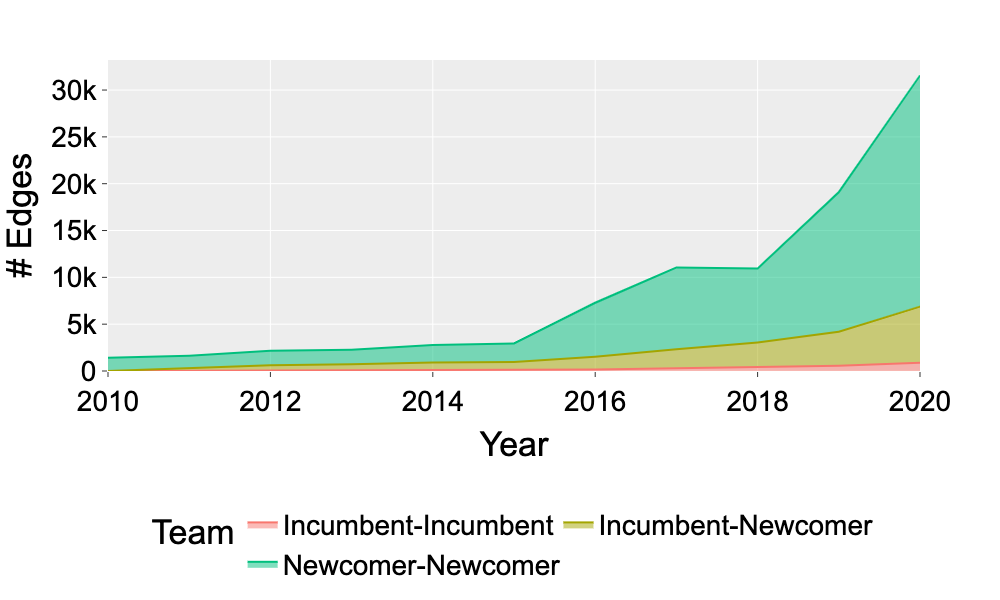}
  \caption{Collaborations (ML)}
  \label{fig:sci_to_sci_ml}
\end{subfigure}
\begin{subfigure}{.32\textwidth}
  \centering
  \includegraphics[width=.98\linewidth]{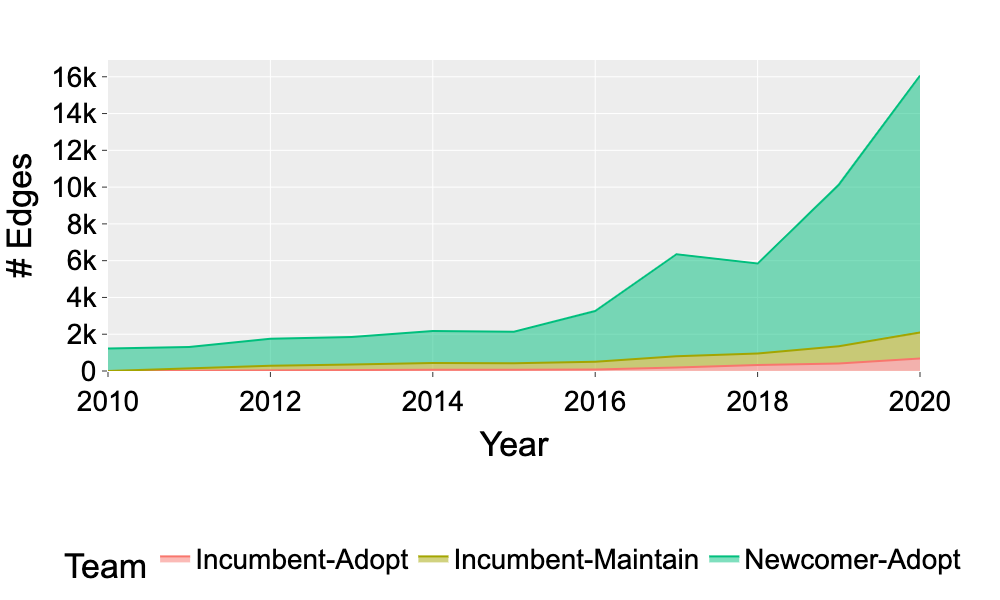}
  \caption{Capabilities (ML)}
  \label{fig:sci_to_cap_ml}
\end{subfigure}
\begin{subfigure}{.32\textwidth}
  \centering
  \includegraphics[width=.98\linewidth]{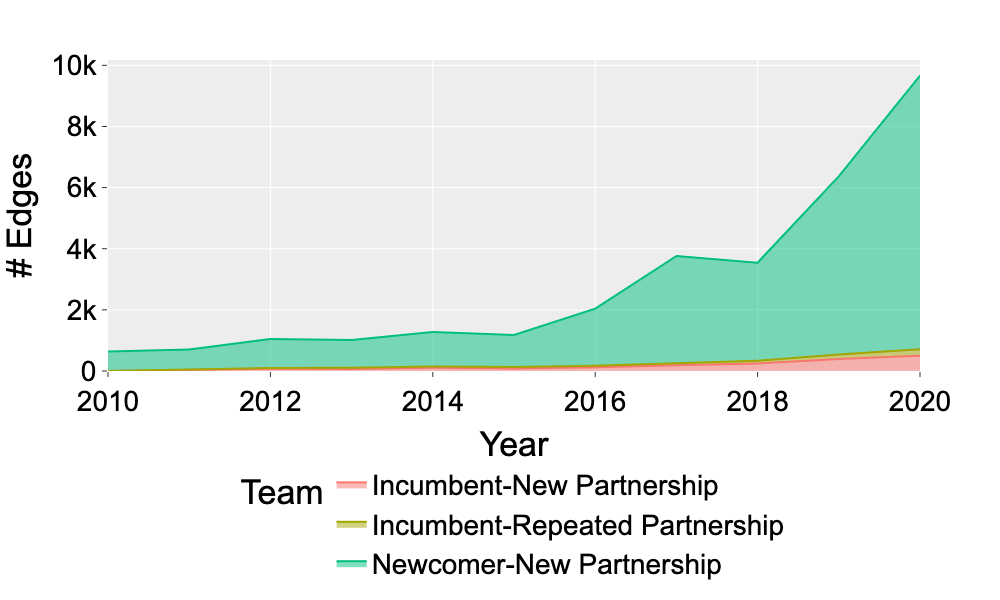}
  \caption{Partnerships (ML)}
  \label{fig:sci_to_int_ml}
\end{subfigure}
\newline
\begin{subfigure}{.32\textwidth}
  \centering
  \includegraphics[width=.98\linewidth]{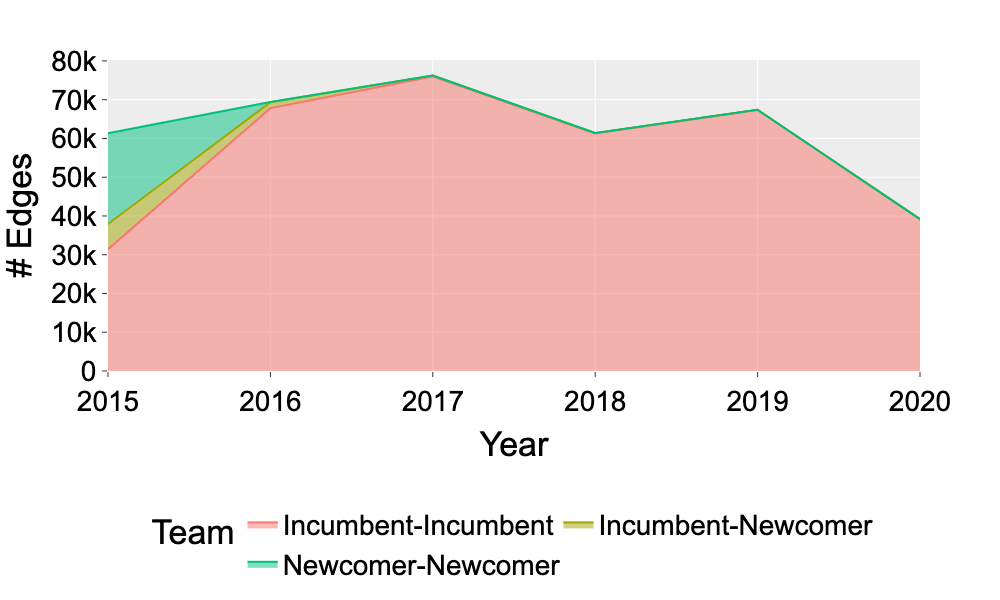}
  \caption{Collaborations (WoS)}
  \label{fig:sci_to_sci_wos}
\end{subfigure}
\begin{subfigure}{.32\textwidth}
  \centering
  \includegraphics[width=.98\linewidth]{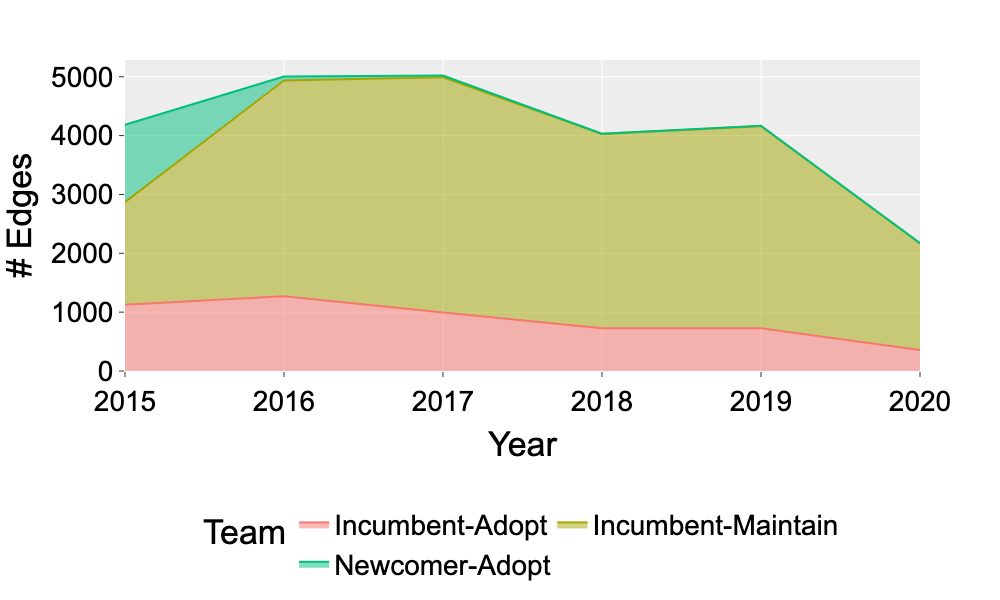}
  \caption{Capabilities (WoS)}
  \label{fig:sci_to_cap_wos}
\end{subfigure}
\begin{subfigure}{.32\textwidth}
  \centering
  \includegraphics[width=.98\linewidth]{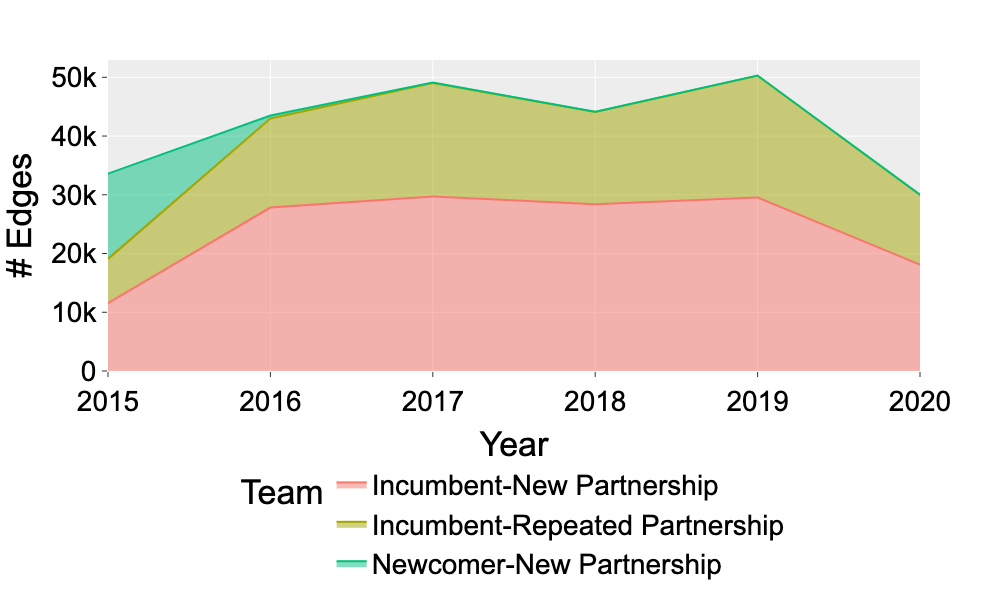}
  \caption{Partnerships (WoS)}
  \label{fig:sci_to_int_wos}
\end{subfigure}
\newline
\begin{subfigure}{.32\textwidth}
  \centering
  \includegraphics[width=.98\linewidth]{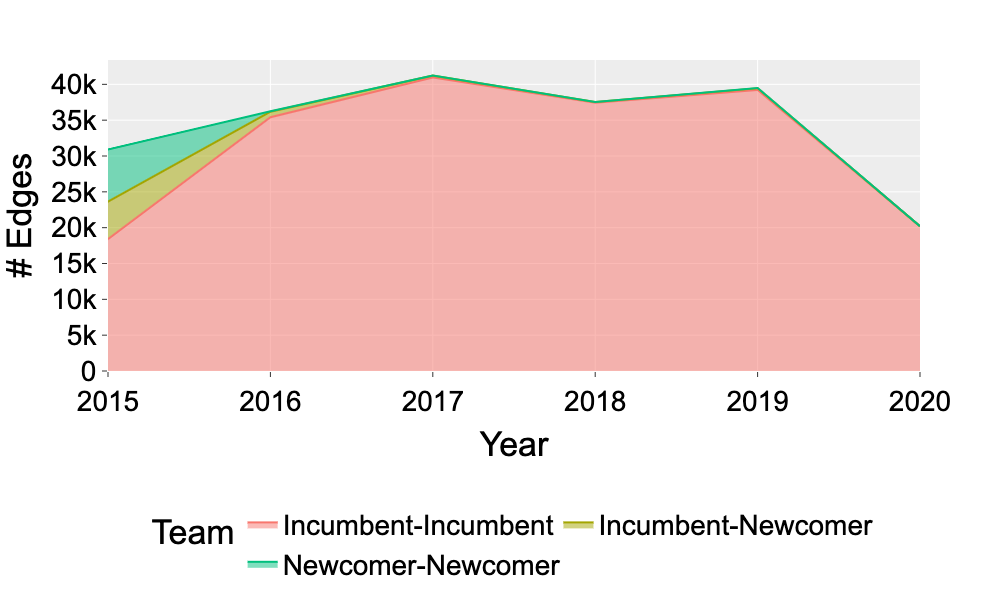}
  \caption{Collaborations (Scopus)}
  \label{fig:sci_to_sci_scopus}
\end{subfigure}
\begin{subfigure}{.32\textwidth}
  \centering
  \includegraphics[width=.98\linewidth]{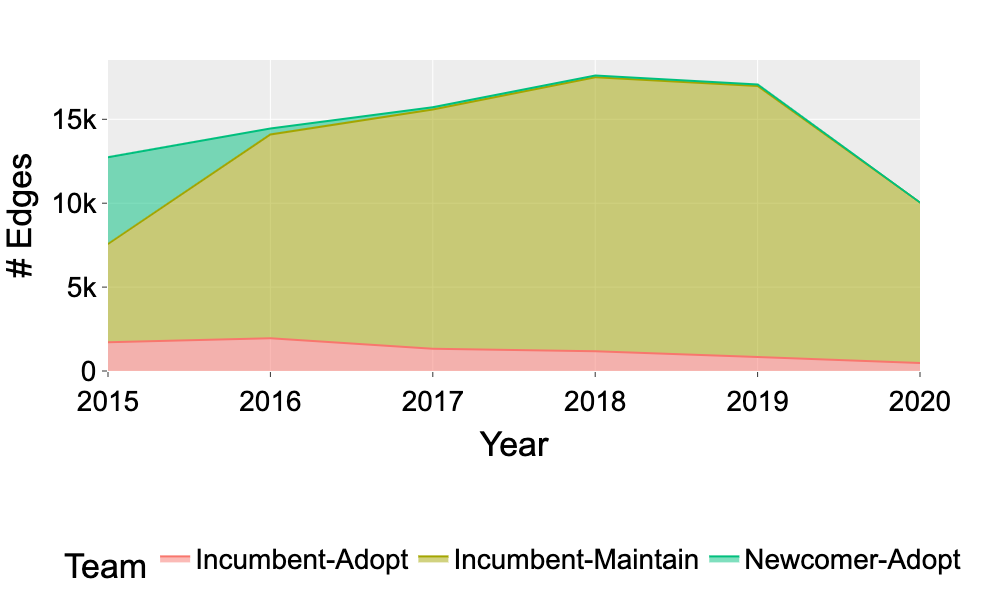}
  \caption{Capabilities (Scopus)}
  \label{fig:sci_to_cap_scopus}
\end{subfigure}
\begin{subfigure}{.32\textwidth}
  \centering
  \includegraphics[width=.98\linewidth]{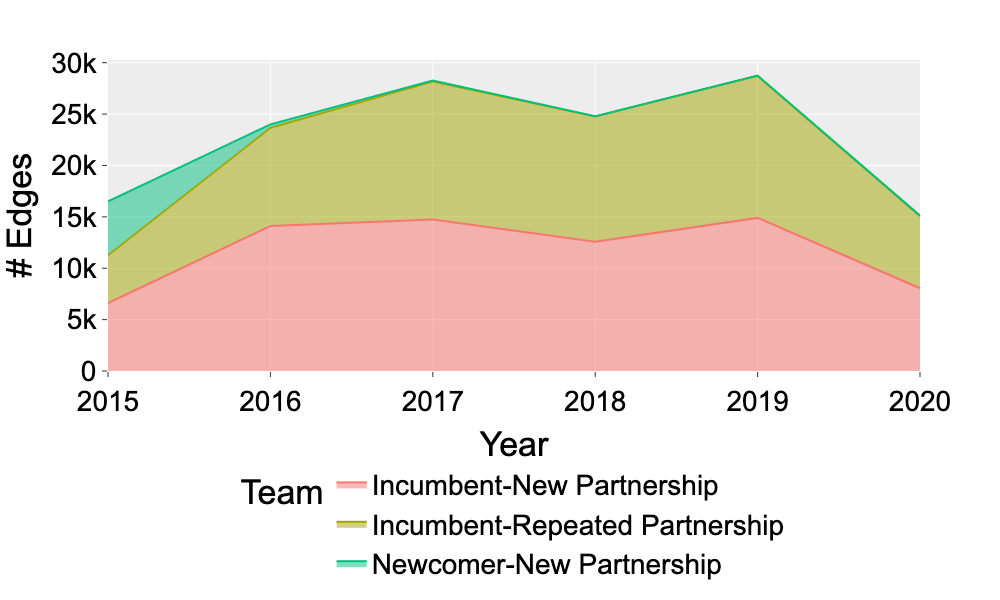}
  \caption{Partnerships (Scopus)}
  \label{fig:sci_to_int_scopus}
\end{subfigure}
\caption{The number of edges over time in academic graphs in the AI and NN domains. First, edges are grouped by the types of nodes (scientist, capability, institution). Second, edges are divided into subgroups based on the types of scientist (incumbent or newcomer) and their actions. For example, i) scientists who collaborate with incumbent vs. newcomer scientists, ii) scientists who  maintain vs. adopt technical capabilities, iii) scientists who establish new vs. repeat partnerships. 
}
\label{fig:timeseries_team_assembly}
\end{figure*}

\begin{figure*}[htbp]
    \begin{subfigure}{.32\textwidth}
  \centering
  \includegraphics[width=.98\linewidth]{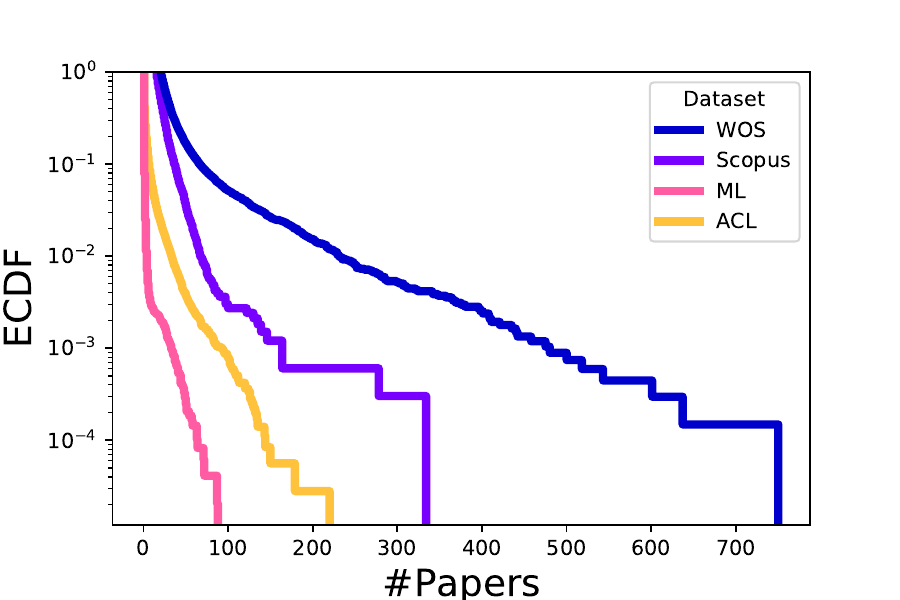}
  \caption{Prolificity}
  \label{fig:activity}
\end{subfigure}
\begin{subfigure}{.32\textwidth}
  \centering
  \includegraphics[width=.98\linewidth]{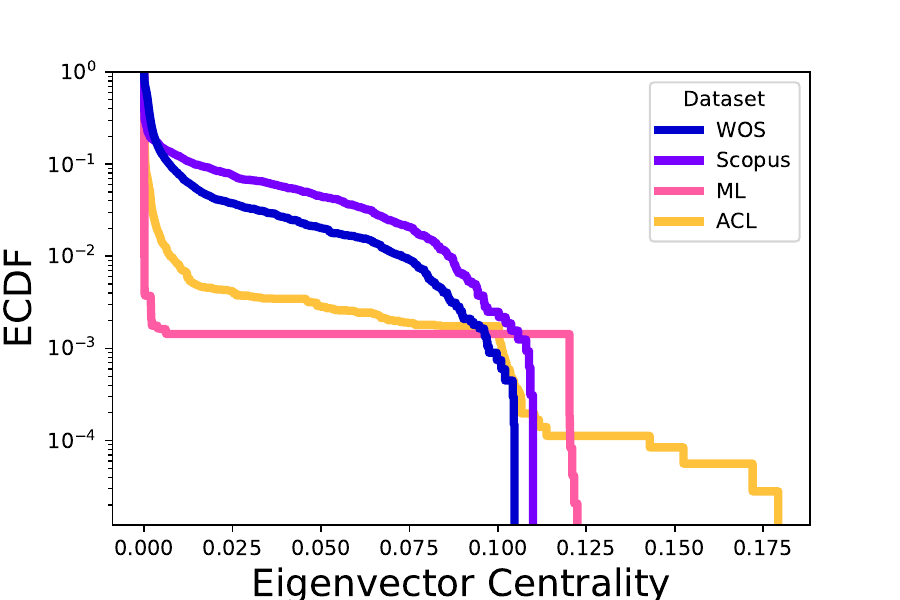}
  \caption{Influence}
  \label{fig:eigen_datasets}
\end{subfigure}
\begin{subfigure}{.32\textwidth}
  \centering
  \includegraphics[width=.98\linewidth]{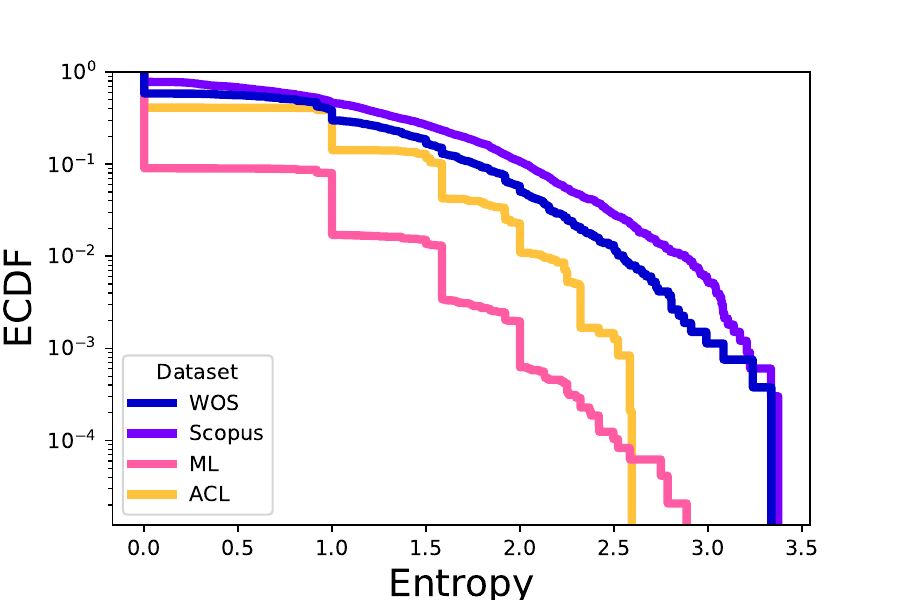}
  \caption{Interdisciplinarity}
  \label{fig:entropy}
\end{subfigure}
\caption{Empirical cumulative distribution of the number of papers (prolificity), centrality (influence), and topic diversity (interdisciplinarity) across datasets. CL and ML in the AI domain, and WoS and Scopus in the NN domain.}
\label{fig:stat}
\end{figure*}

\begin{table*}[htbp]
\centering
\begin{tabular}{|l|l|l|l|}
\hline
 & Top Five Academic Institutions & Top Five Nonacademic Institutions  & \#Institutions \\ \hline \hline
\multirow{5}{*}{ACL} & LTI-Carnegie Melon University, & IBM Research & \multirow{5}{*}{9510} \\
& Stanford University & Academia Sinica No & \\ 
      & Peking University & ACL SRI International & \\ 
      & University of Tsinghua & Baidu Inc & \\ 
      & University of Washington & Microsoft Research & \\ \hline
\multirow{5}{*}{ML} & University of California &  Google Research & \multirow{5}{*}{1753} \\
& Carnegie Mellon University & Microsoft Research & \\
      & Stanford University
      &  IBM Research  & \\ 
      & Princeton University & Microsoft Research Cambridge & \\ 
      & Columbia University & Google Brain - Mountain View & \\ \hline
\multirow{5}{*}{WOS} & Shanghai Jiao Tong Univ,  & Chinese Academy of Sciences,   & \multirow{5}{*}{7589} \\
& Zhejiang University & Chinese Academy of Medical Sciences & \\ 
      & Fudan University & Istituto Nazionale di Fisica Nucleare & \\ 
      & Sun Yat Sen University & Russian Academy of Science  & \\ 
      & Peking University & Harbin Institute of Technology & \\ \hline
\multirow{5}{*}{Scopus} & Michigan State University & Lawrence Livermore National Laboratory & \multirow{5}{*}{7151} \\
& University of Rochester & Oak Ridge National Laboratory & \\ 
      & COMSATS University & Los Alamos National Laboratory & \\ 
      & Quaid-i-Azam University & Idaho National Laboratory & \\ 
      & Harbin Engineering University & Argonne National Laboratory  &  \\ \hline
\end{tabular}
\caption{Total number of institutions present in each dataset and the five most frequently occurring industry and academic institutions.}
\label{tab:institutions}
\end{table*}

\begin{table*}[htbp]
\centering
\begin{tabular}{|l|l|l|}
\hline
 & Top Five Countries (\#Interactions) & \#Countries \\ \hline \hline
\multirow{2}{*}{ACL} & Other (27,661), US (6,931), China (3,059), & 103 \\
 & Germany (2,197), Japan (1,802) &  \\
 \hline
\multirow{2}{*}{ML} & US (1,130), Canada (391), China (390),  & 246 \\
 & UK (376), France (356) &  \\
 \hline
\multirow{2}{*}{WoS} & China (5,632), US (1,080), Germany (412), & 387 \\
 & Japan (405), Italy (344) &  \\
 \hline
\multirow{2}{*}{Scopus} & US (1,354), China (1,207), Italy (995),  & 77 \\
 & Japan (855), Russia (828) &  \\ \hline
\end{tabular}
\caption{Total number of countries present in each dataset and the five most frequently occurring countries.}
\label{tab:countries}
\end{table*}

\subsection{Collaboration Patterns}
As shown in Figure~\ref{fig:activity}, a scientist in the NN domain (\eg from the WoS data) published 41 papers on average in comparison to the scientist in the AI domain (\eg  from the CL community) who published 3 papers on average. This is despite the fact that the CL dataset covers 10 years of publications in comparison to 5 years of publications covered by our WoS dataset. 
Thus, there are more new scientists in the CL community who often collaborate with the incumbent scientists as shown in Figure~\ref{fig:sci_to_sci_acl}. In contrast, the NN community is more densely structured with many incumbent scientists who participate in the new and repeated collaborations. For example, more than 95\% of collaborations occurred between two incumbent scientists in the WoS community in the NN domain (Figure~\ref{fig:sci_to_sci_wos}).

Given these drastically different collaboration patterns across the AI and NN domains, we assessed how influential a scientist is in each collaboration graph. To this end, we computed the Eigenvector centrality on the collaboration networks extracted from all datasets. We found significant differences in the Eigenvector centrality values across AI and NN domains as shown in Figure~\ref{fig:eigen_datasets}.
For example, the average Eigenvector centrality was higher for scientists in the NN datasets in contrast to the AI datasets. One possible explanation could be the presence of highly prolific scientists in the NN domain who are more geographically distributed (Table~\ref{tab:countries}). For example, China has more than five times the scientists who work on nonproliferation compared to the United States in the WoS dataset. In contrast, the United States is dominant in the AI domain.

\subsection{Partnership Dynamics}
In the NN domain, scientists repeatedly form partnerships with institutions (as shown in Figures~\ref{fig:sci_to_int_wos} and~\ref{fig:sci_to_int_scopus}) in contrast to forming new partnerships in the AI domain (Figures~\ref{fig:sci_to_int_acl} and~\ref{fig:sci_to_int_ml}).
Table~\ref{tab:institutions} shows the most active institutions under such partnerships.
We demonstrate an uprising trend in forming new partnerships among scientists in CL and ML communities (Figures~\ref{fig:sci_to_int_acl} and~\ref{fig:sci_to_int_ml}).
Such partnership behavior correlates with the emerging collaboration patterns in the AI domain (Figure~\ref{fig:sci_to_sci_acl} and~\ref{fig:sci_to_sci_ml}).

\subsection{Capability Evolution}
We analyzed dynamic research foci that scientists work  on in the NN and AI domains.
We calculated the tendency to perform multidisciplinary research of each author as the entropy of the author’s capability distribution similar to earlier work~\cite{vlasceanu2022interdisciplinarity}. 
Scientists with higher entropy values published on multiple topics, and thus were more interdisciplinary compared to other scientists.
We found that scientists in the NN domain tended to work on multiple research foci (Figures~\ref{fig:sci_to_cap_wos} and~\ref{fig:sci_to_cap_scopus}) compared to scientists in the AI domain who published on emerging research foci (Figures~\ref{fig:sci_to_cap_acl} and~\ref{fig:sci_to_cap_ml}).

\section{Knowledge-Informed Performance Reasoning}
\label{sec:perf_reasoning}
We analyzed the performance systematically in the previous section based on the edge types.
In this section, we analyze the results across countries, prolificity, scientific elites, interdisciplinarity, and industry/academic partnerships
to better understand 
how and why our models behave the way they do.
This allows us to identify any data or model issues that could impact overall and individual forecasting performance.


\subsubsection{International Diversity}

\begin{figure*}[htbp]
    \begin{subfigure}{.48\textwidth}
        \centering
        \includegraphics[width=0.95\textwidth]{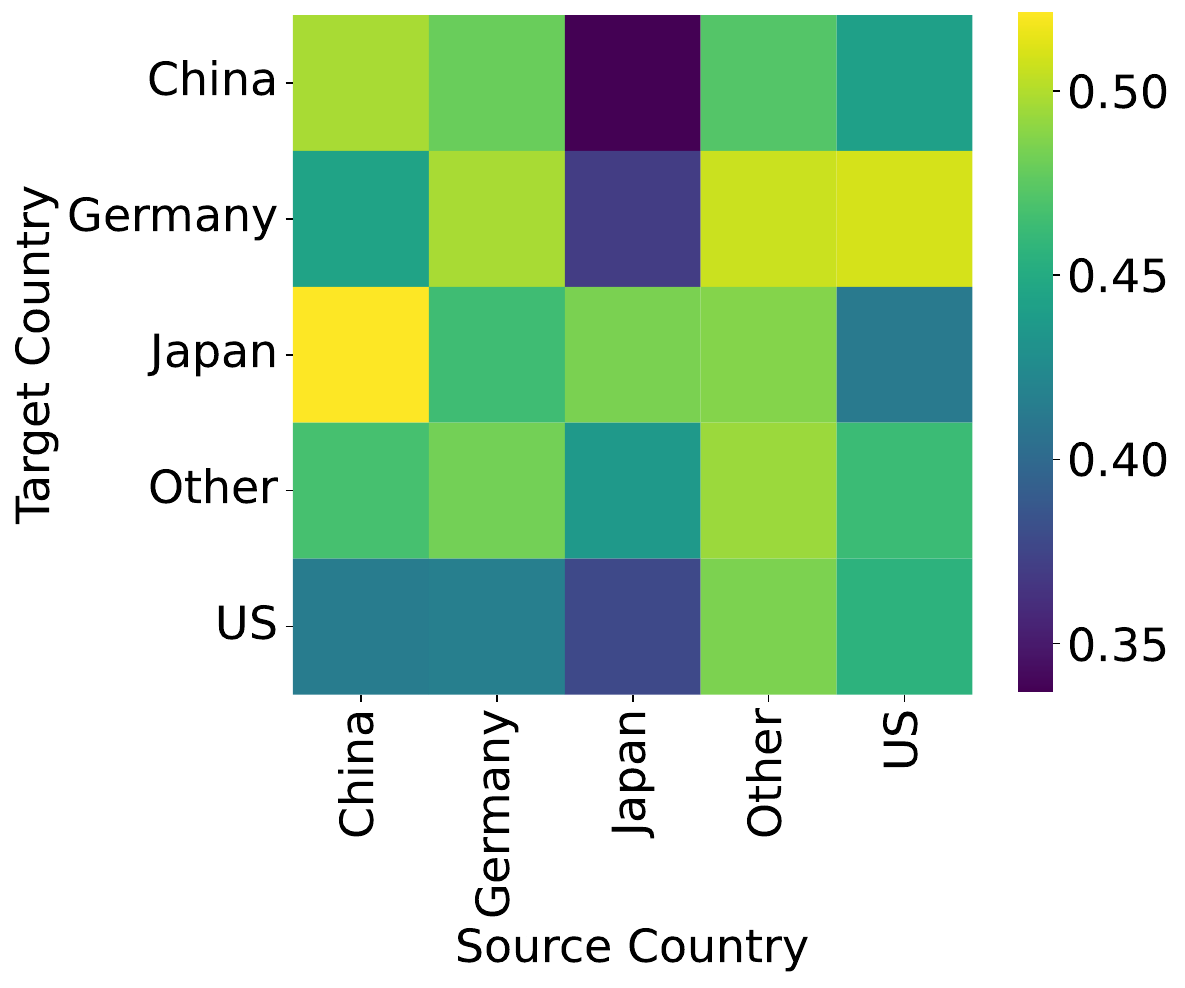}
        \caption{Collaborations (ACL)}
        \label{fig:acl_collab}
    \end{subfigure}
    \begin{subfigure}{.48\textwidth}
        \centering
        \includegraphics[width=0.95\textwidth]{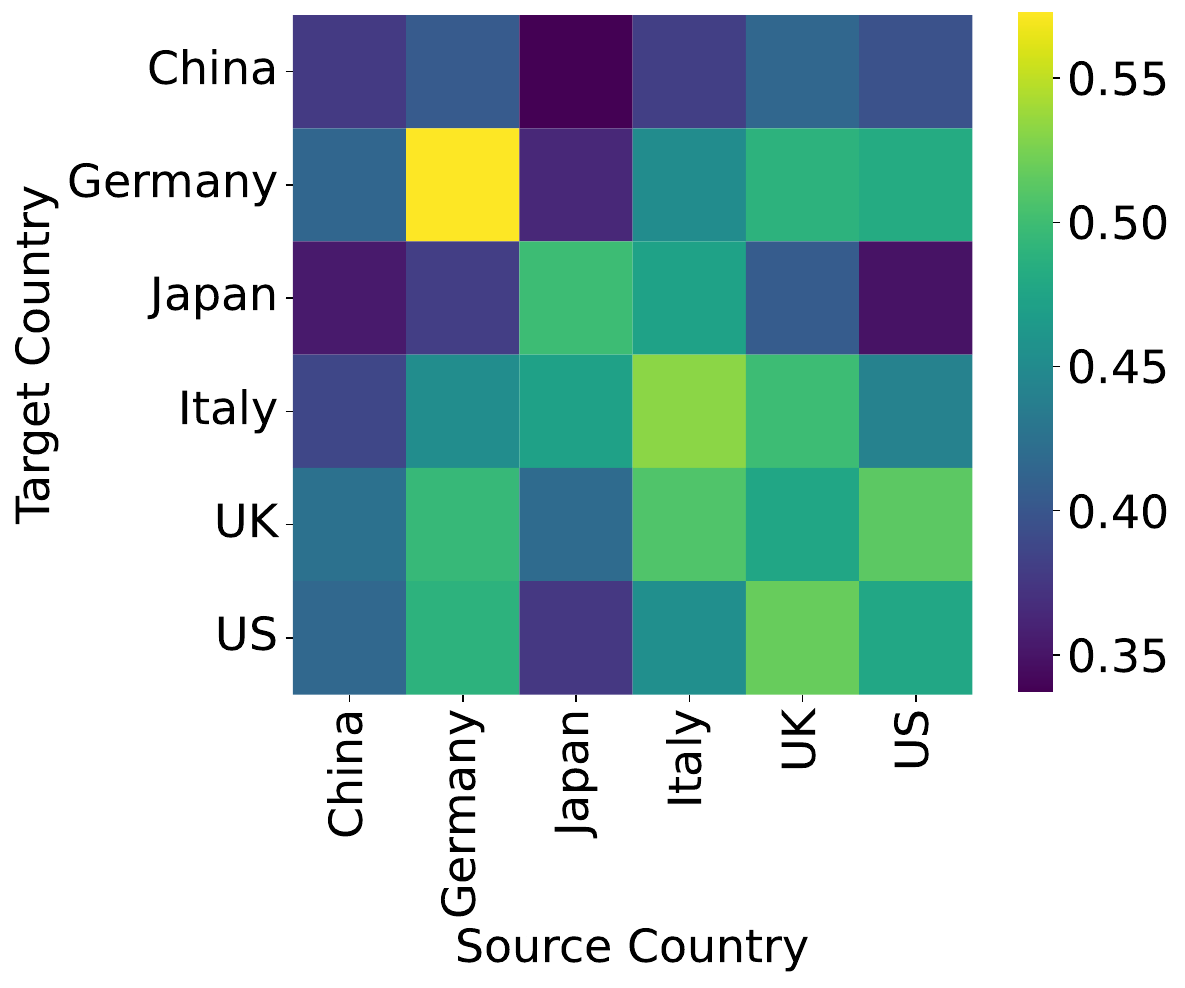}
        \caption{Collaborations (WoS)}
        \label{fig:wos_collab}
    \end{subfigure}\\
    \begin{subfigure}{.48\textwidth}
        \centering
        \includegraphics[width=0.95\textwidth]{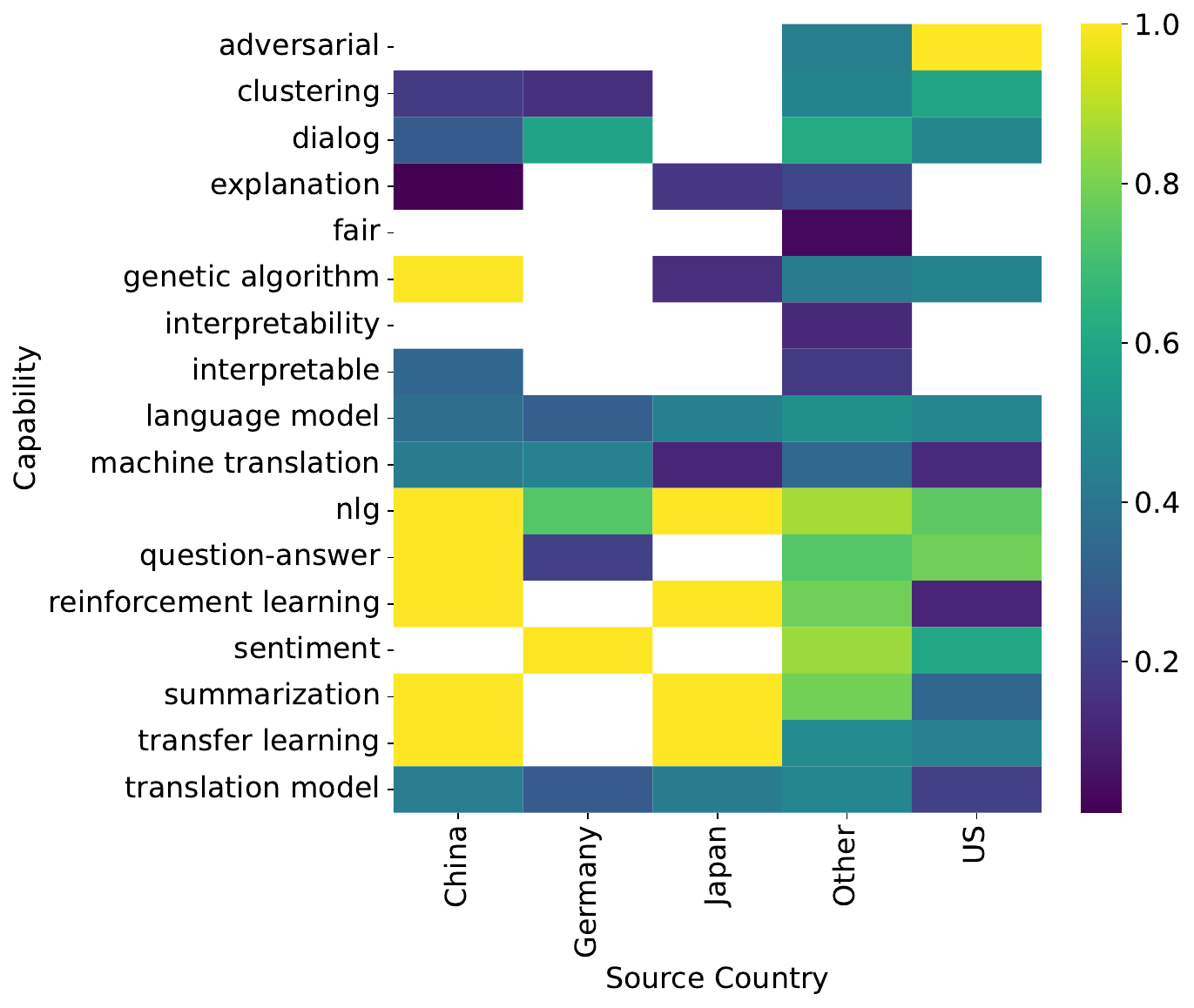}
        \caption{Technical Expertise (ACL)}
        \label{fig:acl_cap}
    \end{subfigure}
    \begin{subfigure}{.48\textwidth}
        \centering
        \includegraphics[width=0.95\textwidth]{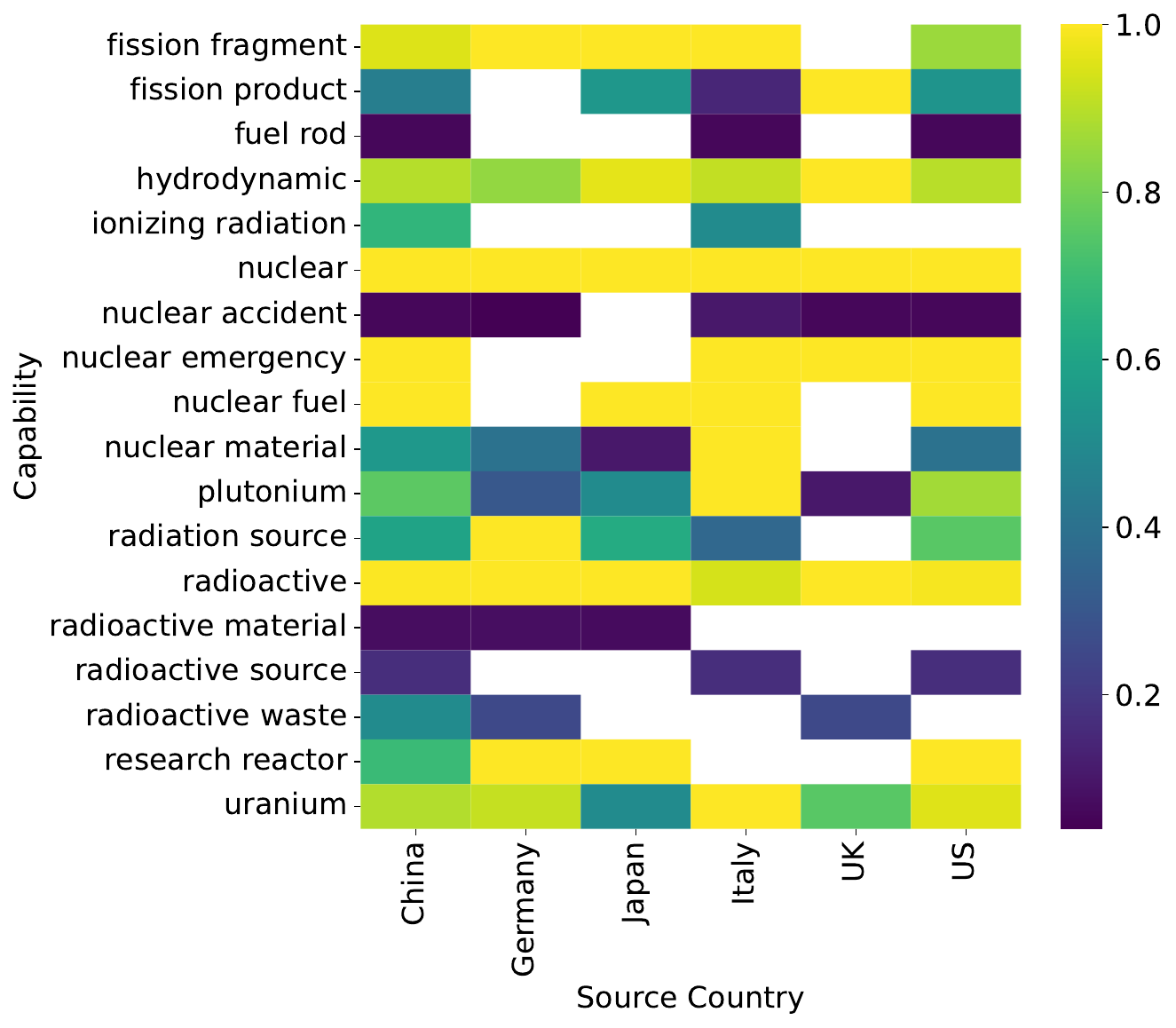}
        \caption{Technical Expertise (WoS)}
        \label{fig:wos_cap}
    \end{subfigure}
    \caption{Model performance (MRR) of collaboration edges between the five most frequent countries and capabilities. Left: best model performance for collaborations between countries of scientist-scientist and scientist-institution edges. Right: best model performance for scientist-capability edges. DGT-C and DGT-D are the best-performing models for ACL and WoS datasets, respectively. Blank cells indicate no edge between entities.}
\end{figure*}


We examined international and domestic collaborations in two datasets: ACL and WoS as shown in Figures \ref{fig:acl_collab} and \ref{fig:wos_collab}.
Interactions between the same countries have different performances across the two domains, \eg China to Japan has a higher MRR value of 0.52 compared to 0.35 MRR from WoS. 
Interactions involving China are the hardest for the WoS DGT-D model to predict despite being the most frequent country.
In WoS, domestic collaborations (\ie interactions within the same country) achieve greater performance over international collaborations (see Figure \ref{fig:wos_collab}).
In contrast, model performance from the ACL community varies in both domestic and international collaborations.

Next, we looked at model performances across capabilities in the AI and NN domains. Figures \ref{fig:acl_cap} and \ref{fig:wos_cap} show the performance (MRR) on edges between the top five countries and capabilities (subsampled due to space) for ACL and WoS. In Figure \ref{fig:acl_cap}, near perfect MRR scores for China and Japan highlight the model's ability to forecast in those expertise areas, \ie natural language generation (nlg), reinforcement learning, summarization, and transfer learning. The 'Other' country category has interactions with all capabilities, but performance varies significantly. One reason for this is the uncertainty in country labeling. Potentially dozens of different countries are being represented by this one category.

\begin{figure*}[htbp]
\begin{subfigure}{.48\textwidth}
  \centering
  \includegraphics[width=.98\linewidth]{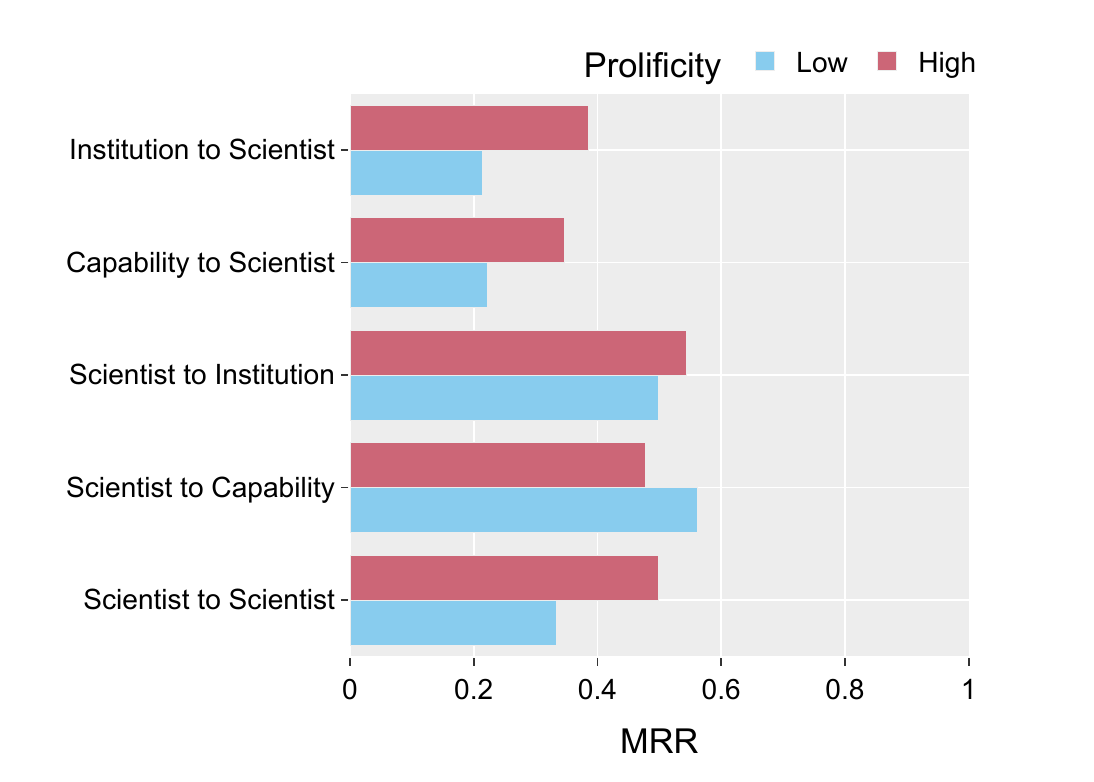}
  \caption{ACL}
  \label{fig:prolific_edge_types_ACL}
\end{subfigure}
\begin{subfigure}{.48\textwidth}
  \centering
  \includegraphics[width=.98\linewidth]{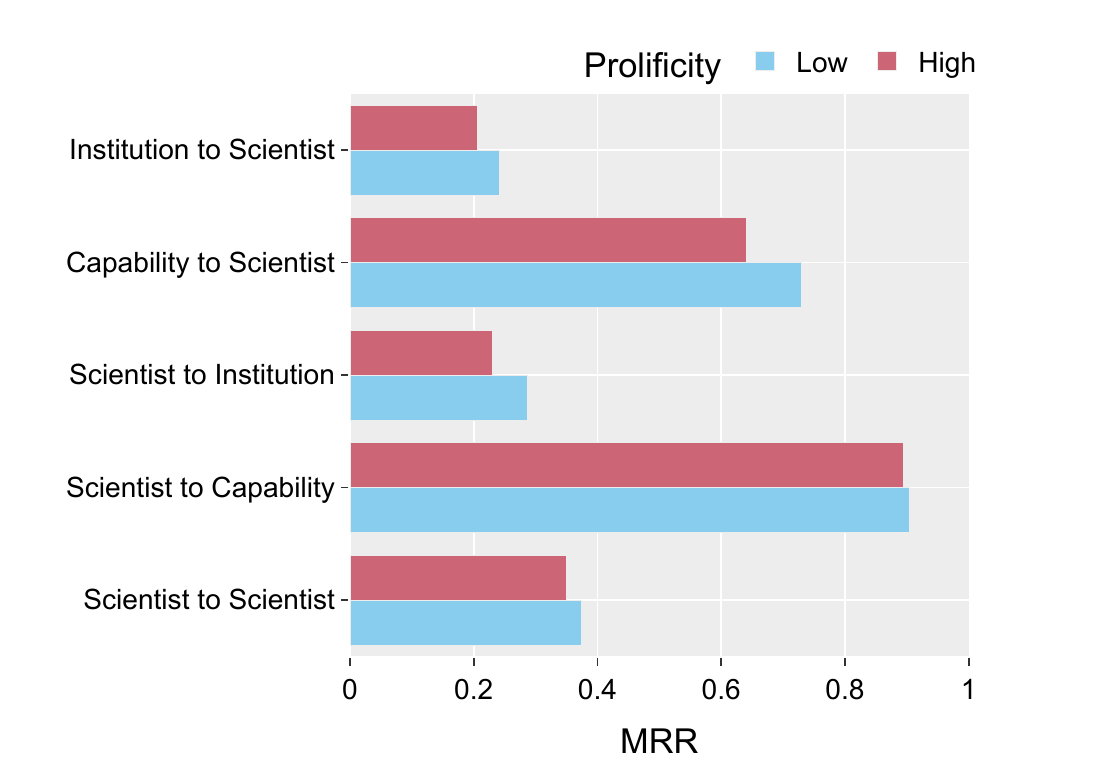}
  \caption{WoS}
  \label{fig:prolific_edge_types_WOS}
\end{subfigure}
\begin{subfigure}{.48\textwidth}
  \centering
  \includegraphics[width=.98\linewidth]{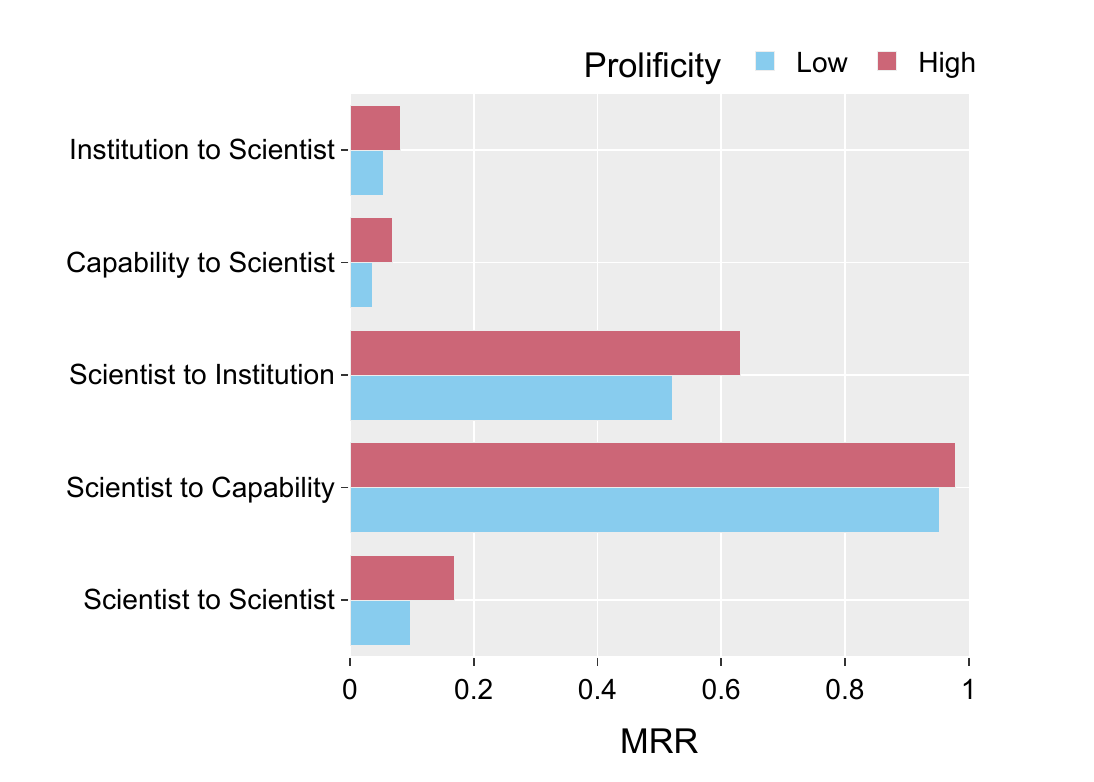}
  \caption{ML}
  \label{fig:prolific_edge_types_ML}
\end{subfigure}
\begin{subfigure}{.48\textwidth}
  \centering
  \includegraphics[width=.98\linewidth]{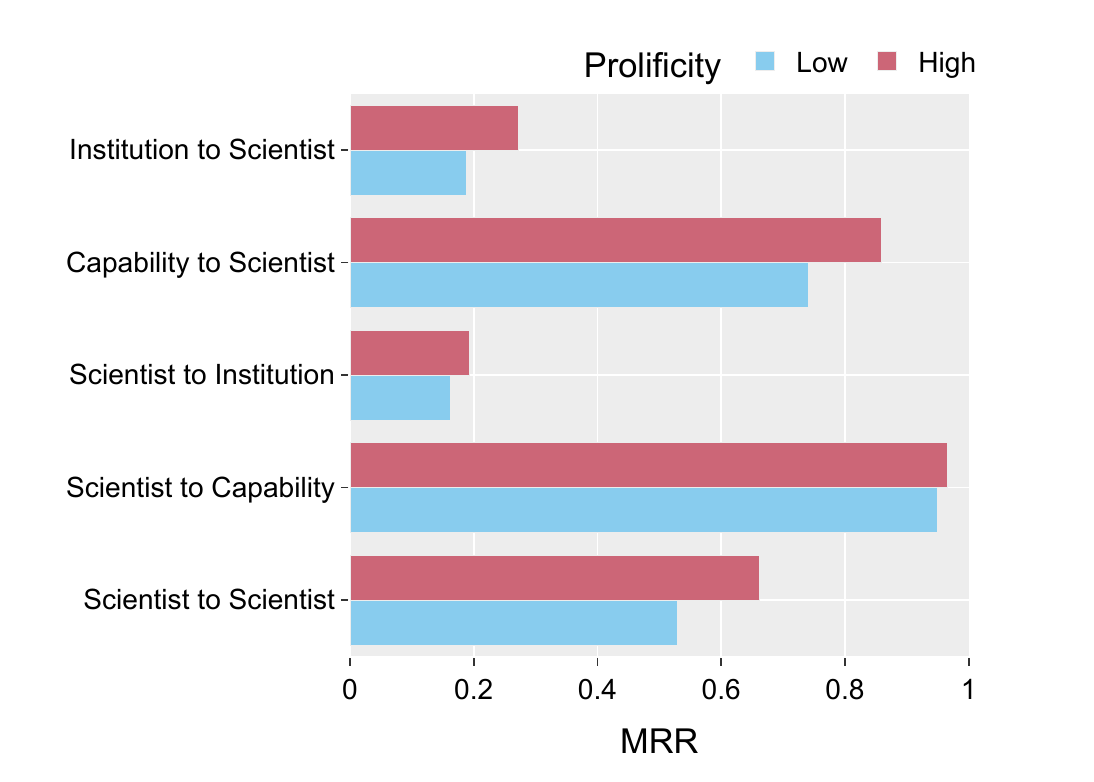}
  \caption{Scopus}
  \label{fig:prolific_edge_types_Scopus}
\end{subfigure}
\caption{Forecasting performance for highly prolific scientists. We classify the scientists into high and low prolific groups based on the median number of papers they published. We only report the performance of the best-performing model in the respective dataset to reduce visual clutter. DGT-C and DGT-D are the best-performing models for AI (ACL/ML) and NN (WoS/Scopus) datasets, respectively.
}
\label{fig:prolific_edge_types}
\end{figure*}

\subsubsection{Highly Prolific Scientists}
While previous work has shown the number of publications grows exponentially~\cite{scisci_review} in general, we showed that scientists have different rates of publishing across NN and AI domains (as discussed in Section~\ref{sec:data_char}).
In this section, our objective is to compare the performance of the models with the publication rate of scientists.
To this end, we group the scientists in each dataset into high and low prolific groups with respect to median number of publications. 
Figure~\ref{fig:prolific_edge_types} shows a comparison of the model performance between these two groups of scientists.

We noticed that the model performs better for high prolific scientists across all datasets than for low prolific scientists.
For example, there is a 50-75\% performance increase for high prolific scientists in the AI datasets (Figures~\ref{fig:prolific_edge_types_ACL} and~\ref{fig:prolific_edge_types_ML}).
This quantitative evidence suggests that models are capturing rich signals from the high prolific scientists in the collaboration, partnership, and capability evolution.
However, model performance in the WoS is an exception to this pattern.
As shown in Figure~\ref{fig:prolific_edge_types_WOS}, the performance across high and low prolific groups of scientists is more comparable, while the model records slightly better performance for low prolific scientists than the other group.
This is due to the higher rate of publications on scientists in the WoS dataset.
As shown in Figure~\ref{fig:activity}, scientists in the WoS datasets published 41 papers on average between 2015-2020, roughly one paper every two months.
This provides enough historical signals for the model to learn the collaboration, partnership, and capability evolution patterns.

\begin{figure*}[htbp]
\begin{subfigure}{.48\textwidth}
  \centering
  \includegraphics[width=.98\linewidth]{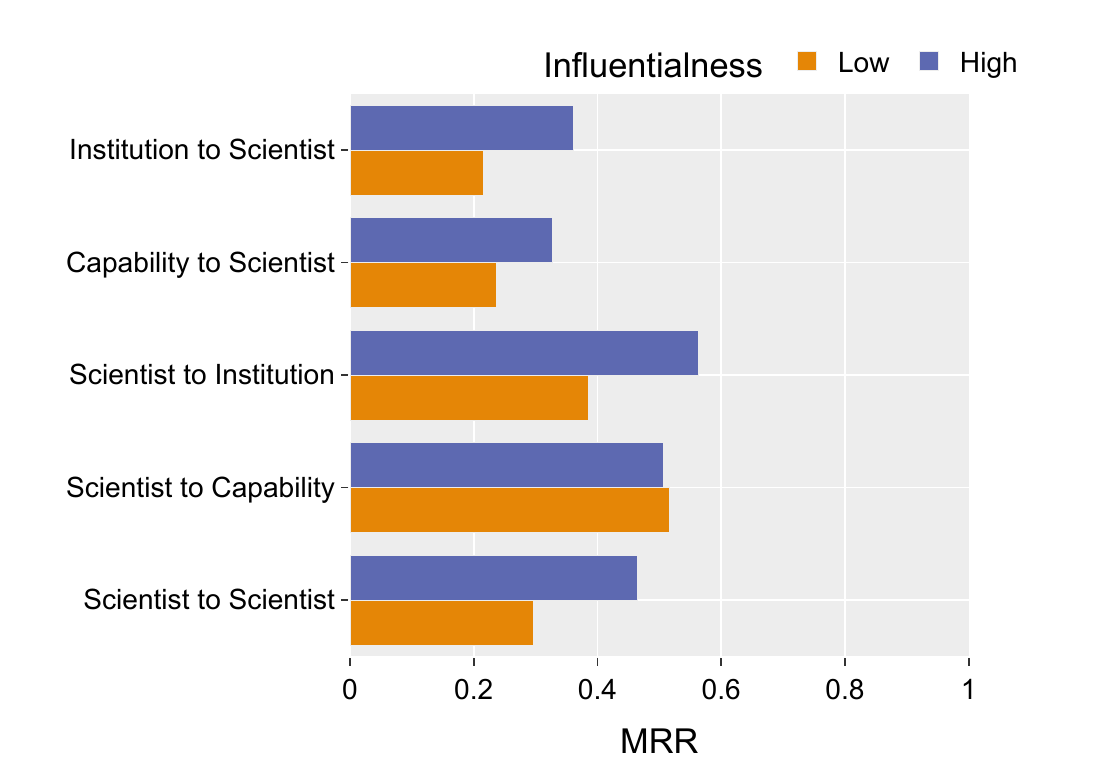}
  \caption{ACL}
  \label{fig:eigen_edge_types_ACL}
\end{subfigure}
\begin{subfigure}{.48\textwidth}
  \centering
  \includegraphics[width=.98\linewidth]{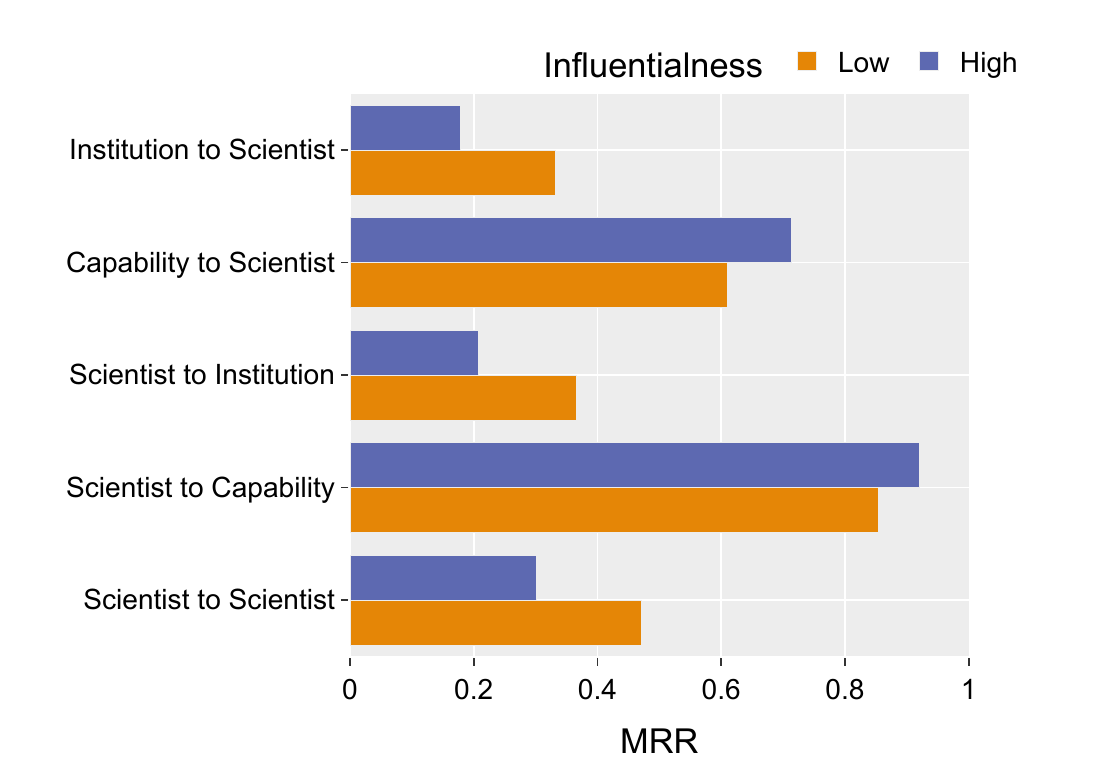}
  \caption{WoS}
  \label{fig:eigen_edge_types_WOS}
\end{subfigure}
\begin{subfigure}{.48\textwidth}
  \centering
  \includegraphics[width=.98\linewidth]{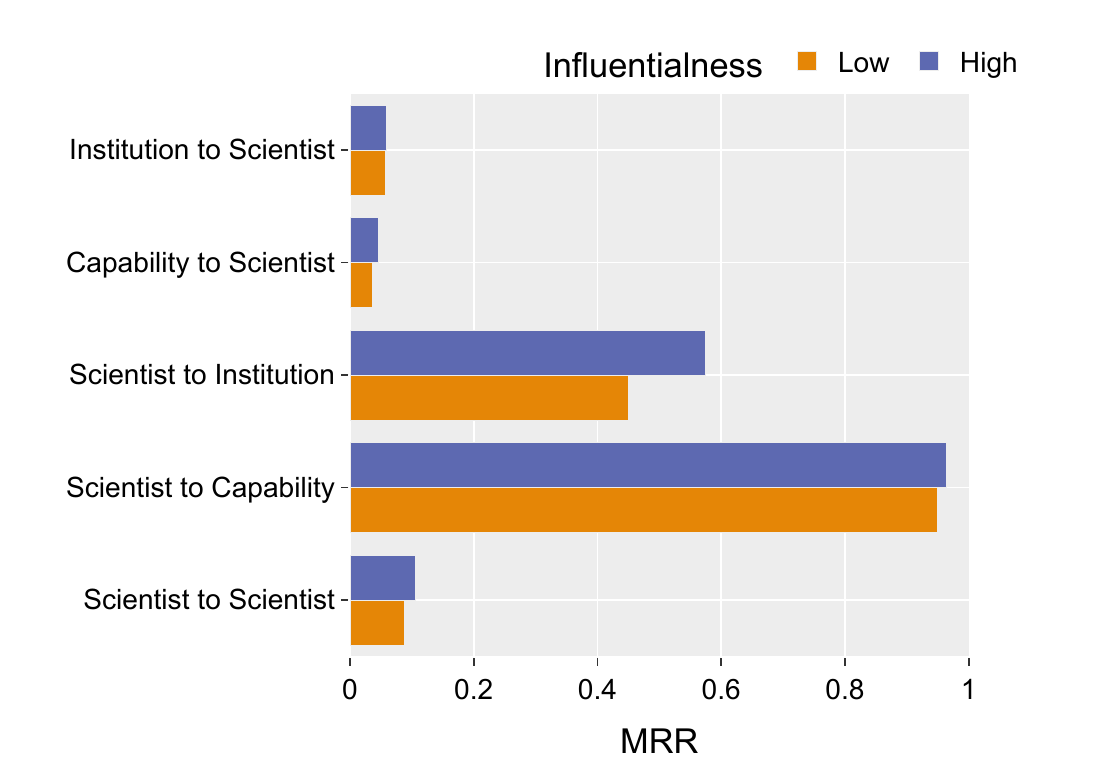}
  \caption{ML}
  \label{fig:eigen_edge_types_ML}
\end{subfigure}
\begin{subfigure}{.48\textwidth}
  \centering
  \includegraphics[width=.98\linewidth]{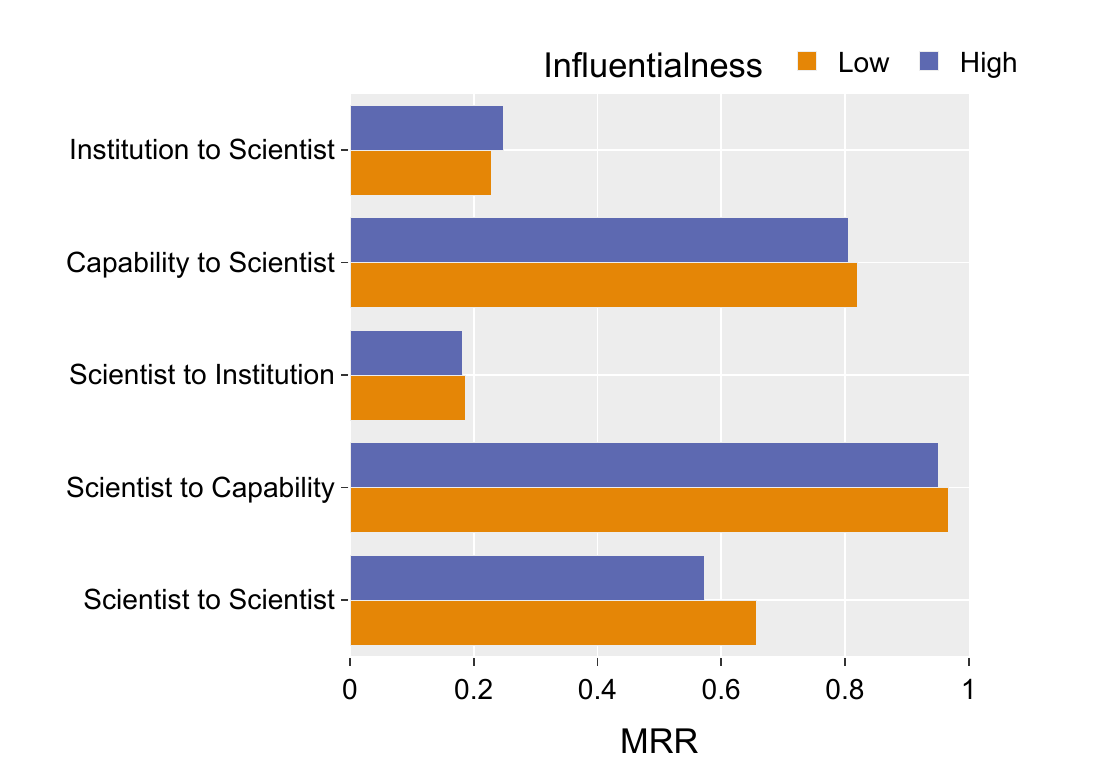}
  \caption{Scopus}
  \label{fig:eigen_edge_types_Scopus}
\end{subfigure}
\caption{Forecasting performance for highly influential scientists. We calculate the Eigenvector centrality for the scientist nodes appearing in the collaboration network and classify them to highly influential scientists based on the median centrality value resulted in the respective dataset. We only report the performance of the best-performing models, DGT-C and DGT-D in the AI (CL/ML) and NN (WoS/Scopus) datasets, respectively.
}
\label{fig:eigen_edge_types}
\end{figure*}

\subsubsection{Scientific Elites}

To this end, we computed the Eigenvector centrality to assess how influential a scientist is given the collaboration network.
We take the median Eigenvector centrality in each dataset to classify the highly influential scientists.
In this section, we analyze forecasting performance on predicting collaboration, partnership, and expertise edges across scientific elites.
As shown in Figures~\ref{fig:eigen_edge_types_ACL}-\ref{fig:eigen_edge_types_Scopus}, we compare highly influential scientists to other scientists.

We noticed that the models predict the collaboration patterns for highly influential scientists more accurately than other scientists in AI datasets  (Figures~\ref{fig:eigen_edge_types_ACL} and~\ref{fig:eigen_edge_types_ML}).
However, we noticed the opposite in the NN datasets, where the models perform well on predicting collaboration patterns for less influential scientists (Figures~\ref{fig:eigen_edge_types_WOS} and~\ref{fig:eigen_edge_types_Scopus}).
We believe this is due to the differences in the topological characteristics between AI and NN highly influential scientists.
For example, the models perform well capturing the densely clustered topological structure in the NN domain than the less clustered, hierarchical structure in the AI domain.

\begin{figure*}[!t]
\begin{subfigure}{.48\textwidth}
  \centering
  \includegraphics[width=.98\linewidth]{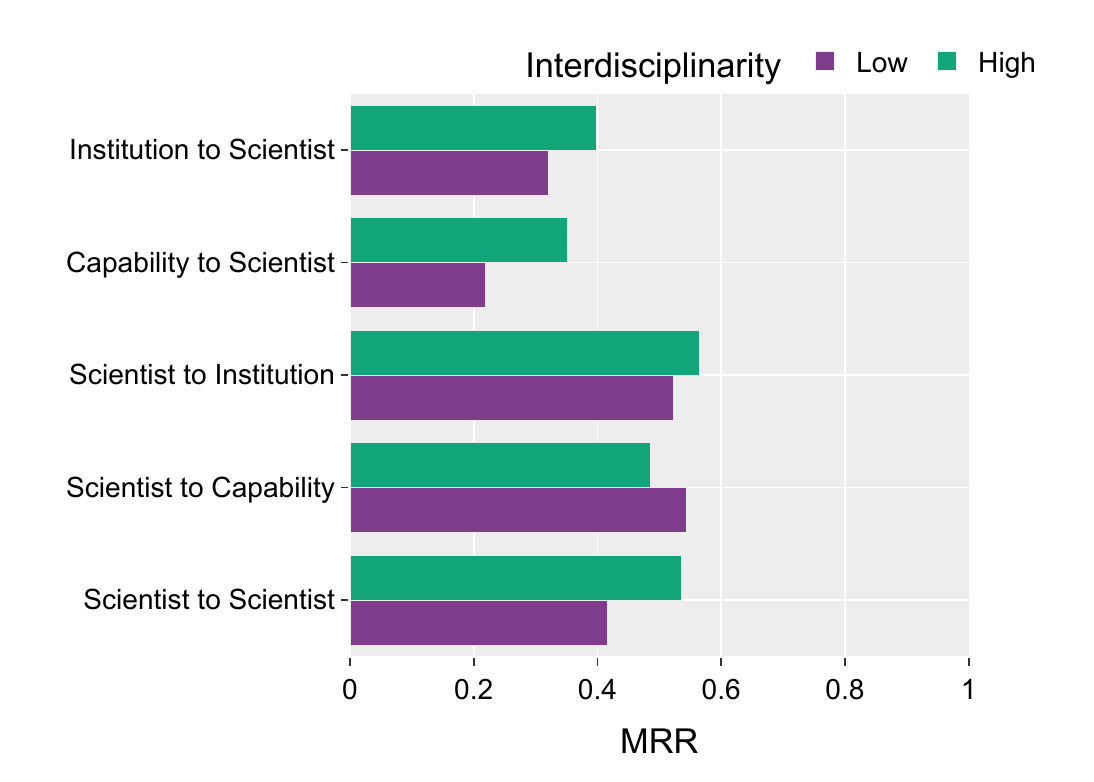}
  \caption{ACL}
  \label{fig:interd_edge_types_ACL}
\end{subfigure}
\begin{subfigure}{.48\textwidth}
  \centering
  \includegraphics[width=.98\linewidth]{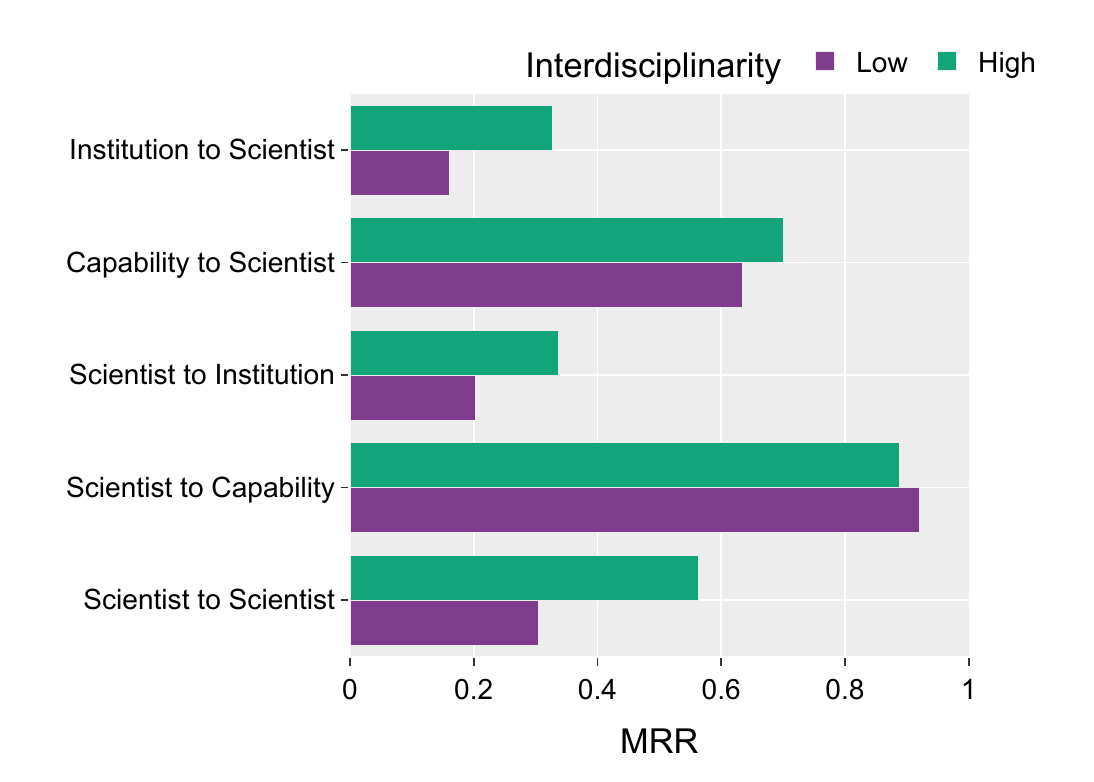}
  \caption{WoS}
  \label{fig:interd_edge_types_WOS}
\end{subfigure}
\begin{subfigure}{.48\textwidth}
  \centering
  \includegraphics[width=.98\linewidth]{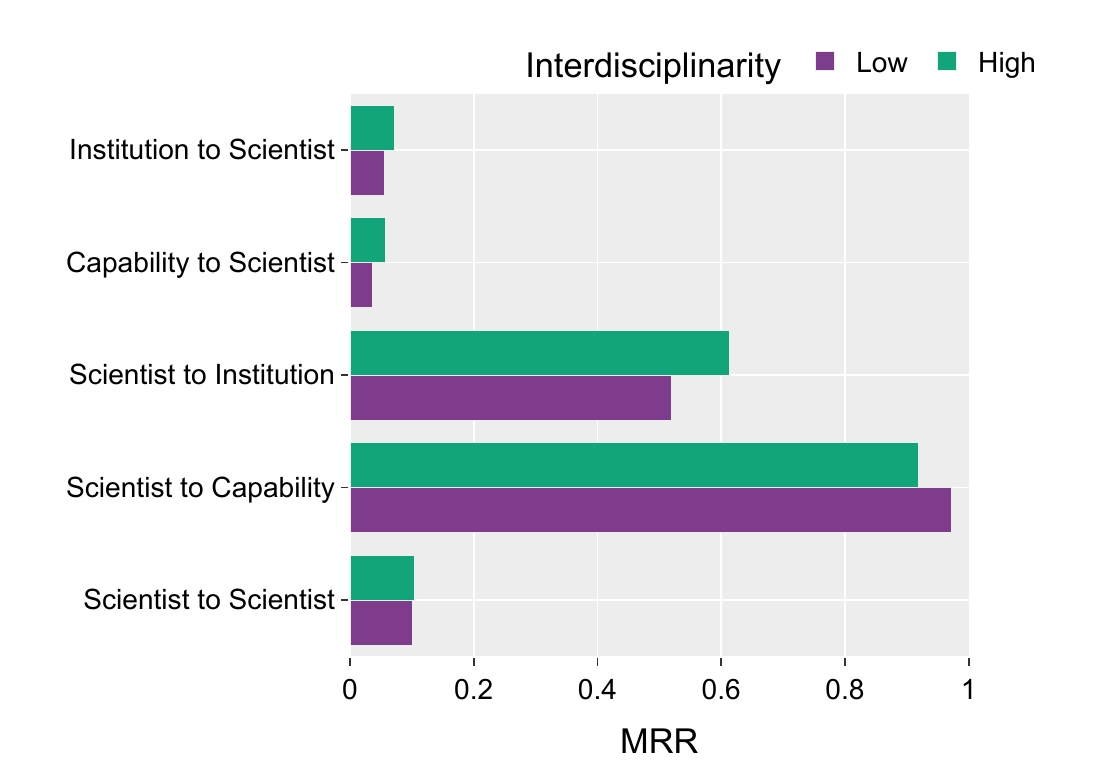}
  \caption{ML}
  \label{fig:interd_edge_types_ML}
\end{subfigure}
\begin{subfigure}{.48\textwidth}
  \centering
  \includegraphics[width=.98\linewidth]{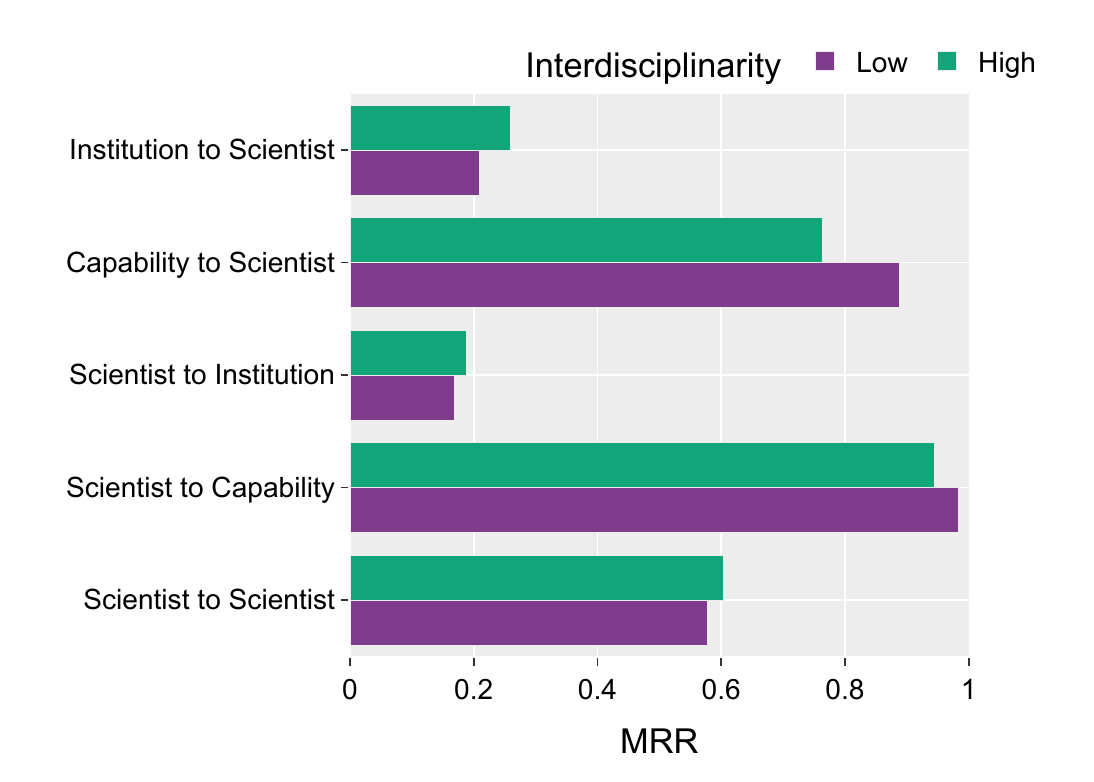}
  \caption{Scopus}
  \label{fig:interd_edge_types_Scopus}
\end{subfigure}
\caption{Forecasting performance for highly interdisciplinary scientists. For each scientist, we recorded a vector of frequency describing the number of times they published on a certain topic. We calculate the entropy of topic vector for all scientists and classify them to highly interdisciplinary scientists based on the median entropy value resulted in the respective dataset. We only report the performance of the best-performing model in the respective dataset to reduce visual clutter. DGT-C and DGT-D are the best-performing models for AI (ACL/ML) and NN (WoS/Scopus) datasets, respectively.
}
\label{fig:interd_edge_types}
\end{figure*}

\begin{figure*}[!t]
\begin{subfigure}{.24\textwidth}
  \centering
  \includegraphics[width=.98\linewidth]{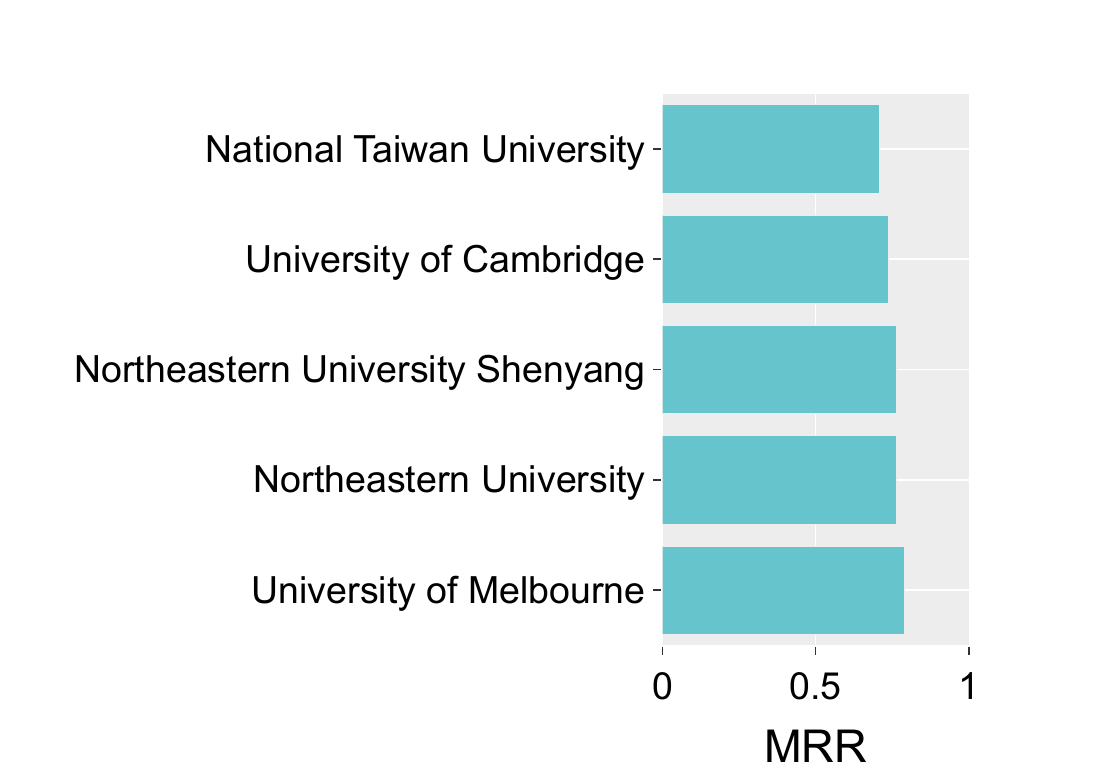}
  \caption{Academia (ACL)}
  \label{fig:intera_edge_types_ACL}
\end{subfigure}
\begin{subfigure}{.24\textwidth}
  \centering
  \includegraphics[width=.98\linewidth]{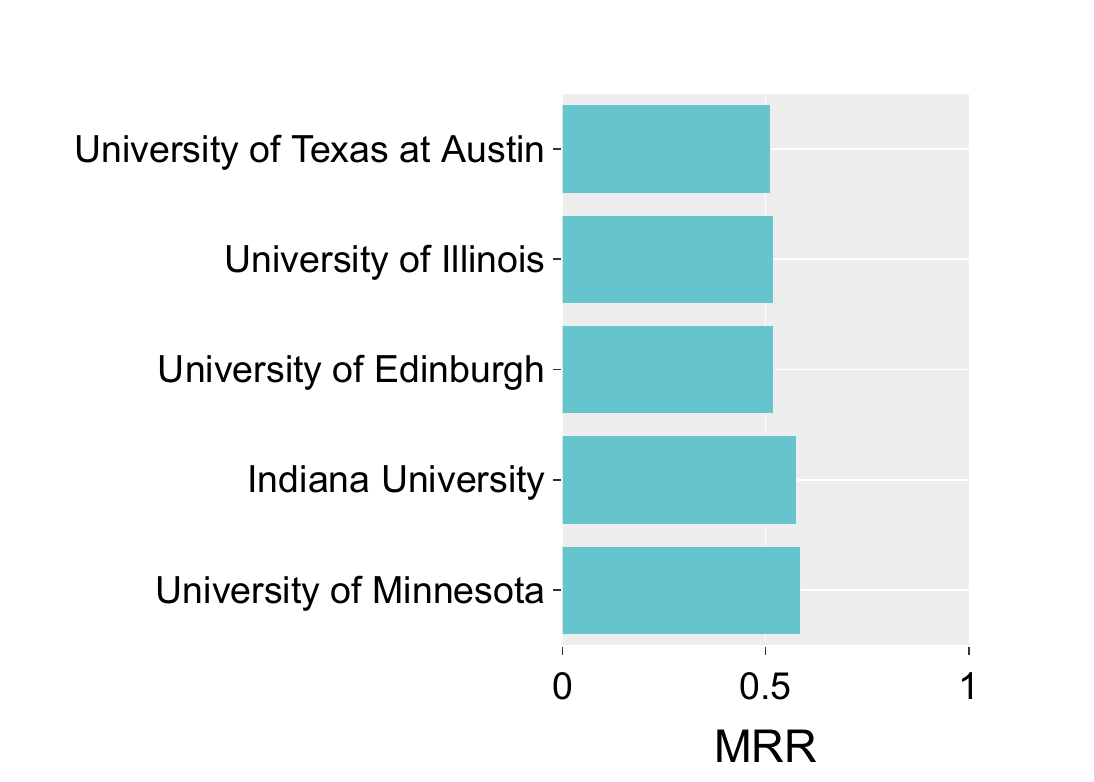}
  \caption{Academia (ML)}
  \label{fig:intera_edge_types_WOS}
\end{subfigure}
\begin{subfigure}{.24\textwidth}
  \centering
  \includegraphics[width=.98\linewidth]{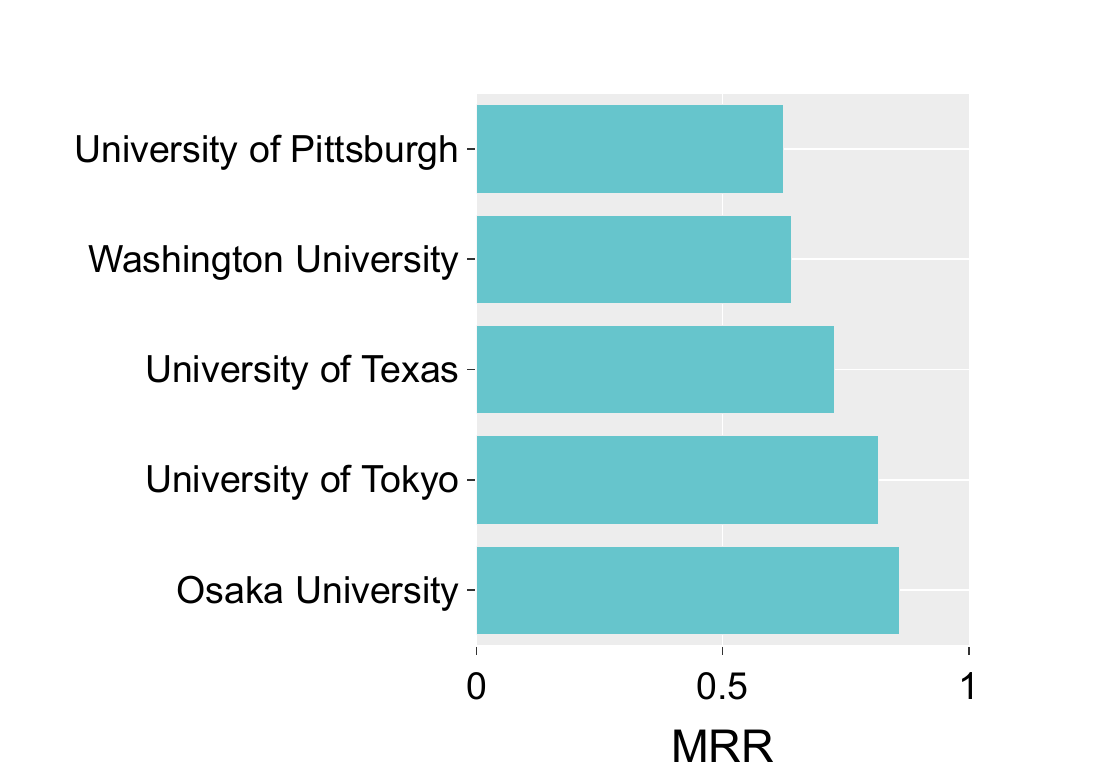}
  \caption{Academia (WoS)}
  \label{fig:intera_edge_types_ML}
\end{subfigure}
\begin{subfigure}{.24\textwidth}
  \centering
  \includegraphics[width=.98\linewidth]{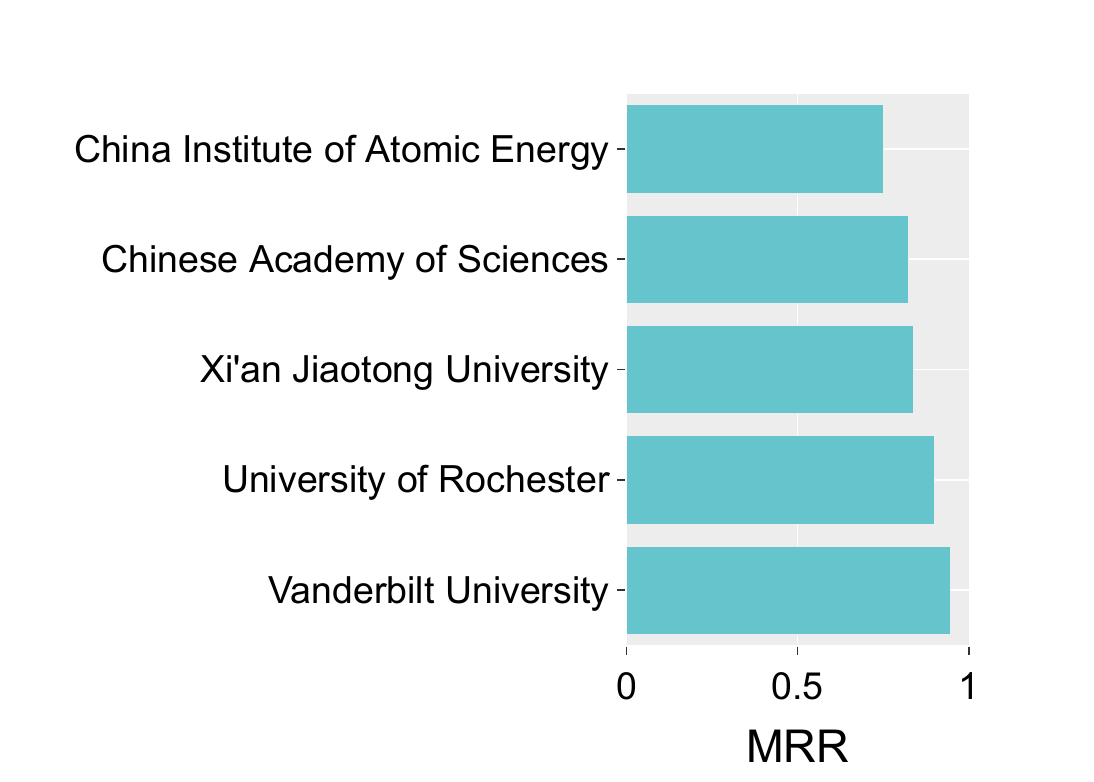}
  \caption{Academia (Scopus)}
  \label{fig:intera_edge_types_Scopus}
\end{subfigure}
\begin{subfigure}{.24\textwidth}
  \centering
  \includegraphics[width=.98\linewidth]{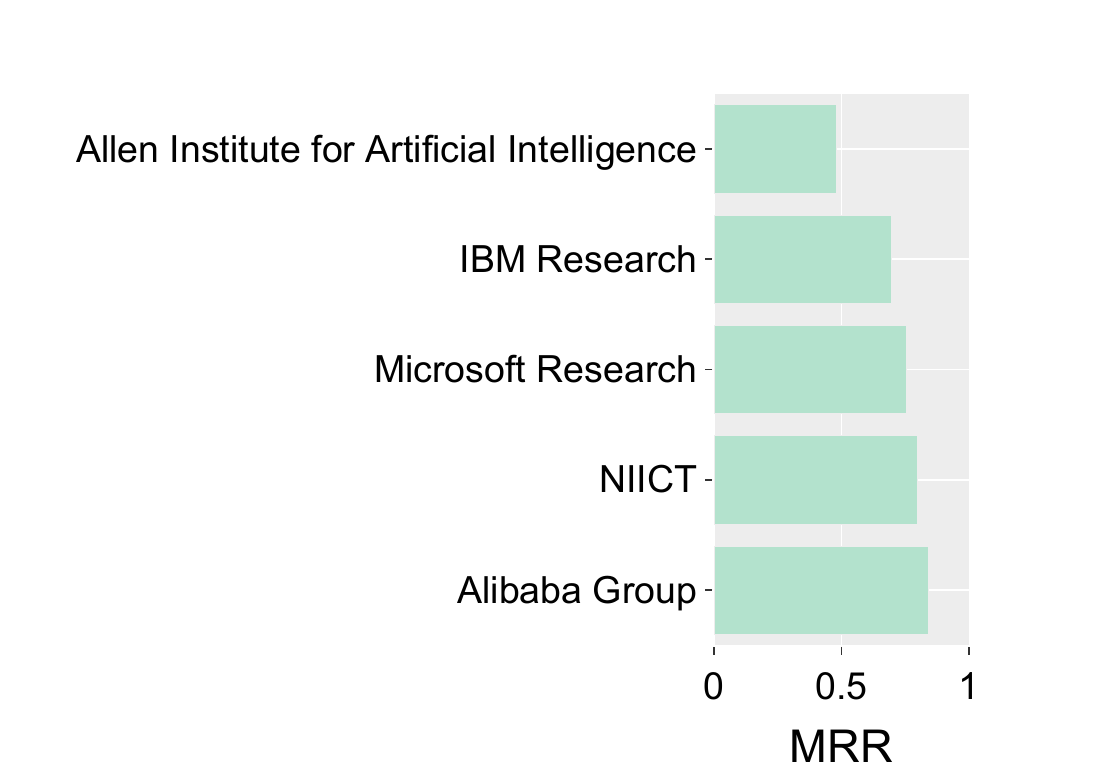}
  \caption{Non-Academia (ACL)}
  \label{fig:interi_edge_types_ACL}
\end{subfigure}
\begin{subfigure}{.24\textwidth}
  \centering
  \includegraphics[width=.98\linewidth]{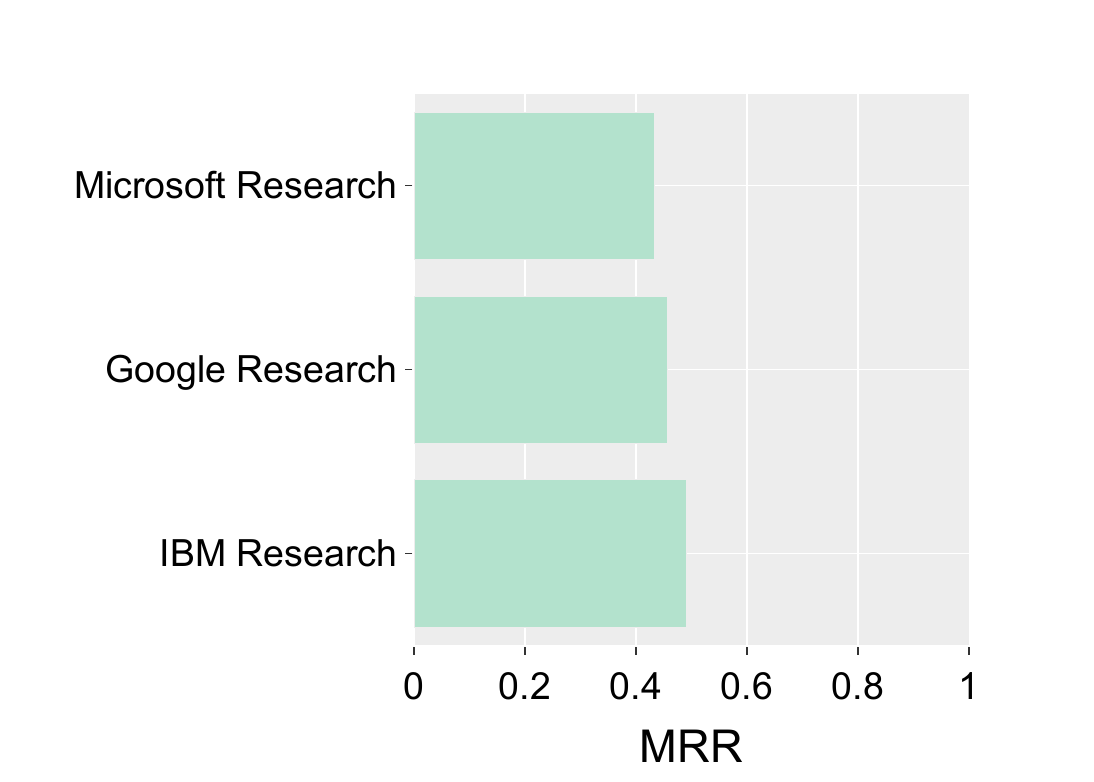}
  \caption{Non-Academia (ML)}
  \label{fig:interi_edge_types_WOS}
\end{subfigure}
\begin{subfigure}{.24\textwidth}
  \centering
  \includegraphics[width=.98\linewidth]{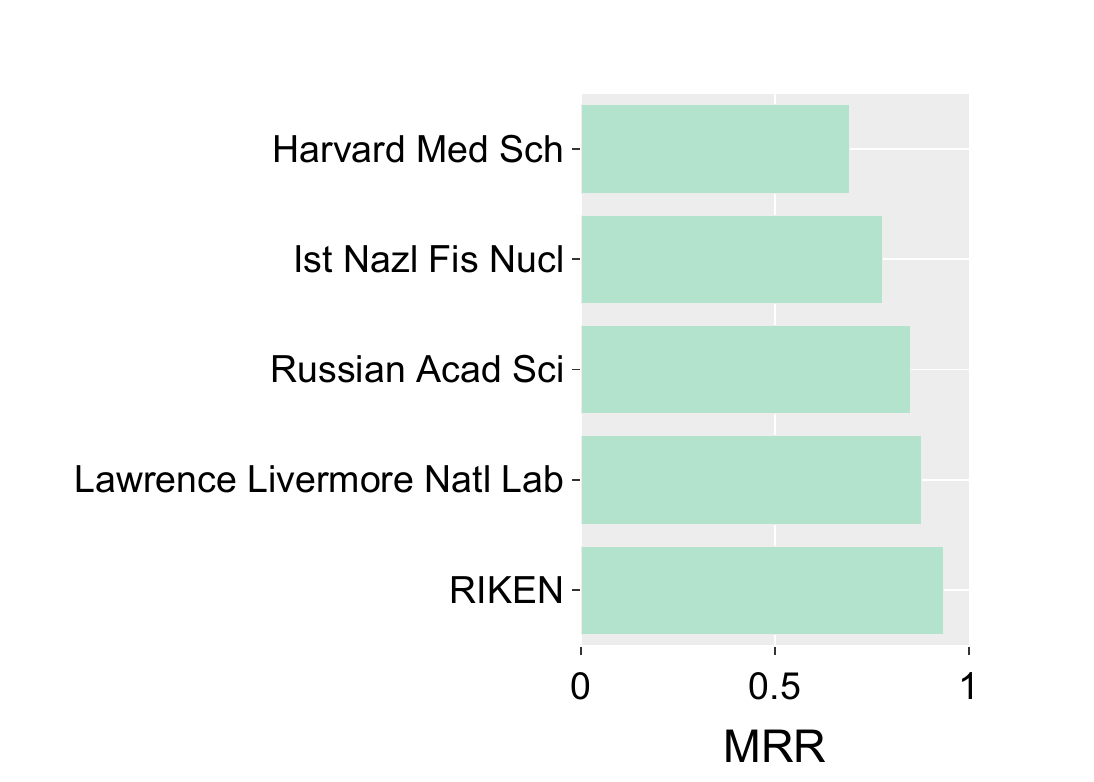}
  \caption{Non-Academia (WoS)}
  \label{fig:interi_edge_types_ML}
\end{subfigure}
\begin{subfigure}{.24\textwidth}
  \centering
  \includegraphics[width=.98\linewidth]{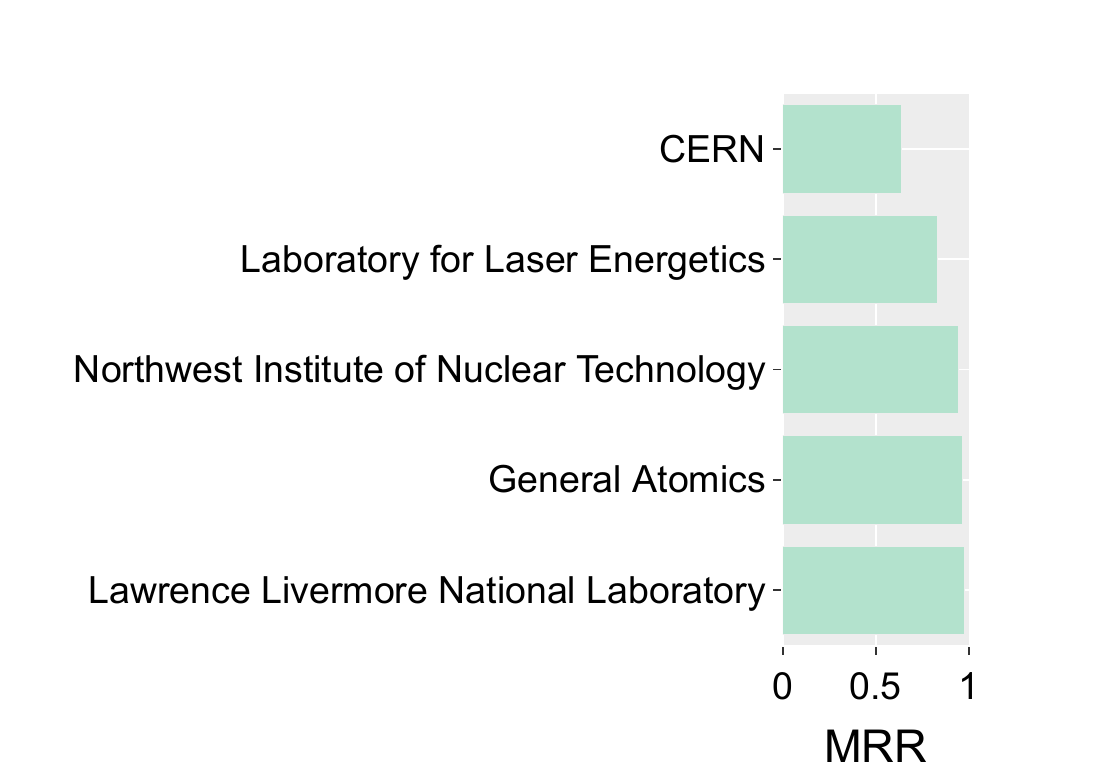}
  \caption{Non-Academia (Scopus)}
  \label{fig:interi_edge_types_Scopus}
\end{subfigure}
\caption{Top five academic and nonacademic institutions with best model performance. The top and bottom rows represent the academic and nonacademic (\eg industry and government institutions, respectively.
}
\label{fig:academia_industry}
\end{figure*}

\subsubsection{Interdisciplinarity}
Scientists work on multiple research topics over time due to media popularity, research interests, availability of funds, etc.
In this section, we assess the predictability of interdisciplinary scientists and institutions across AI and NN domains.
We calculated the interdisciplinarity of each author as the entropy of the author’s topic distribution. 
Scientists with higher entropy values publish on multiple topics, thus they are highly interdisciplinary compared to other scientists.
We take the median entropy value in each dataset to form two groups such as highly interdisciplinary and other scientists.
We compare the performance of models between the two groups for forecasting collaboration, partnership, and expertise edges (as shown in Figure~\ref{fig:interd_edge_types}).
We focus on the best-performing DGT-C and DGT-D models for AI and NN datasets, respectively.

We noticed that models forecast the collaboration, partnership, and expertise edges for highly interdisciplinary scientists more accurately than other scientists.
For example, the DGT-C model records 0.53 MRR for forecasting the collaboration edges across highly interdisciplinary scientists in ACL in contrast to 0.42 MRR for the rest of scientists.
We observe a similar pattern in the WoS dataset where the performance advantage (54\%) is more significant than the ACL dataset.
We believe this is due to the highly interdisciplinary scientists who work on NN rather than the AI domain.
For example, the average entropy value is higher (0.73) for NN scientists than AI scientists (0.51).

\subsubsection{Industry vs. Academia}

Scientific collaborations are not limited to academic institutions, but include  research organizations and universities and industry partners for joint research efforts.
Most often, industry scientists show the technological applications of scientific research that would foster the intellectual synergy between academia and industry~\cite{industry_research_2018}.
At the same time, industry and academia collaborations face some challenges in research dissemination such as intellectual properties issues, publishing versus patenting competitions, etc.
In this section, we analyze the model performance to understand the predictability of scientists partnering with industry and academia.
To this end, we manually classify the top 100 most frequently appearing institutions into academia and nonacademia.
Note that we classify both industry and government organizations as nonacademia.
Table~\ref{tab:institutions} shows the characteristics of academic and nonacademic institutions.
Then, we compare the model performance across scientists who partner with academic and nonacademic institutions in research collaboration.
Figure~\ref{fig:academia_industry} shows the best-performing institutions across all datasets.
We have three main observations from this analysis.

First, we note that the models achieve the high MRR values in the NN datasets.
For example, the DGT-D model predicts which scientists would partner with Lawrence Livermore National Laboratory in 0.97 MRR (Figure~\ref{fig:interi_edge_types_Scopus}).
The same model performs best on predicting partnerships for General Atomics, an American energy and defense corporation, and the Northwest Institute of Nuclear Technology (NINT), which provides support for the nuclear weapons test program in China.
It is important to note  the ability of the models to generalize across these institutions that are geographically distributed across the United States and China.
In addition, the DGT-D model predicts well for scientists who partner with academic institutions.
In the Scopus dataset, while the best-performing academic institutions are two U.S. institutions ($>0.9$ MRR), the next three best-performing academic institutions ($>0.75$ MRR) are located in China (Figure~\ref{fig:intera_edge_types_Scopus}).

Second, the DGT-C model predicts partnerships for nonacademic institutions more accurately than the academic institutions in the ACL dataset.
For example, the Alibaba Group (Chinese multinational technology company) and the National Institute of Information and Communications Technology (Japan's primary national research institute for information and communications) are the best-performing institutions with $>0.79$ MRR.
The same model achieves $< 0.79$ MRR values for academic institutions where two U.S. institutions are the best performing.

Finally, we noticed that model performance is different for the same institution across different datasets.
For example, the DGT-C model predicts the partnership edges for Microsoft Research with 0.75 MRR in the ACL dataset but achieves 0.43 MRR in the ML dataset.
We believe this is mostly due to emerging partnerships with institutions in the ML dataset.
There are 94\% first-time partnerships with Microsoft Research to collaborate on ML topics in the testing period in comparison to 54\% of such partnerships in the ACL dataset.
Reinforcement and adversarial learning are the most popular topics under such partnerships.

}

\end{document}